%% file: arxiv.tex
\documentclass[10pt,twocolumn,letterpaper]{article}

\usepackage[pagenumbers]{cvpr}

\input{preamble}

\input{misc/miscs}

\definecolor{cvprblue}{rgb}{0.21,0.49,0.74}
\usepackage[pagebackref,breaklinks,colorlinks,allcolors=cvprblue]{hyperref}

\def\confName{CVPR}
\def\confYear{2027}

\title{Is Endpoint Matching All You Need for Lane Topology Reasoning?}

\author{Kyungmin Kim \quad Sumin Lee \quad Sungoh Jeong\\
Hyunseok Choi \quad Geonyeong Park \quad Soonmin Hwang$^{*}$\\
Hanyang University\\
{\tt\small \{tnrud17, maroona, sungoh0036, hyun15946, dsa3515, soonminh\}@hanyang.ac.kr}
}

\begin{document}
\twocolumn[{%
\renewcommand\twocolumn[1][]{#1}%
\maketitle
\input{sec/teaser}
}]
{\let\thefootnote\relax\footnotetext{$^{*}$Corresponding author.}}
{\let\vfill\relax\input{sec/0_abstract}}
\input{sec/1_intro}

\input{sec/2_related_work}
\input{sec/3_dataset}

\input{sec/4_method}

\input{sec/5_experiments}

\input{sec/6_conclusion}

{
    \small
    \bibliographystyle{ieeenat_fullname}
    \bibliography{main}
}

\input{sec/X_suppl}

\end{document}

%% file: preamble.tex
\usepackage{booktabs}
\usepackage{multirow}
\usepackage{amssymb}
\usepackage{colortbl}
\usepackage{hhline}
\usepackage{adjustbox}
\usepackage{tabularx}
\usepackage{algorithm}
\usepackage{algpseudocode}
\AtEndPreamble{\crefname{algorithm}{Alg.}{Algs.}}
\newcolumntype{Y}{>{\centering\arraybackslash}X}
\newcolumntype{G}{>{\columncolor{olsbg}\centering\arraybackslash}X}
\definecolor{olsbg}{RGB}{235,235,235}
\newcolumntype{g}{>{\columncolor{olsbg}}c}
\newcommand{\ghead}[1]{\cellcolor{white}\makebox[0pt][l]{\hspace*{-\tabcolsep}\textcolor{olsbg}{\rule[-\dp\strutbox]{\dimexpr\linewidth+2\tabcolsep\relax}{\dimexpr 2.7ex+\dp\strutbox-1pt\relax}}}\makebox[\linewidth]{#1}}
\newcommand{\rtop}{\ensuremath{\text{RTOP}_{ll}}}
\newcommand{\rols}{ROLS\xspace}

\makeatletter
\renewcommand\paragraph{\@startsection{paragraph}{4}{\z@}{4pt \@plus 1pt \@minus 1pt}{-1em}{\normalfont\normalsize\bfseries}}
\makeatother

\usepackage{xspace}
\newcommand{\ours}{TopoReach\xspace}
\definecolor{upgreen}{RGB}{0,128,64}

%% file: misc/miscs.tex
\definecolor{kmcolor}{RGB}{166,94,0}
\definecolor{smcolor}{RGB}{192,0,122}
\definecolor{socolor}{RGB}{0,121,107}
\definecolor{hscolor}{RGB}{94,53,177}

%% file: sec/teaser.tex
\begin{center}
  \centering
  \resizebox{\textwidth}{!}{%
    \includegraphics{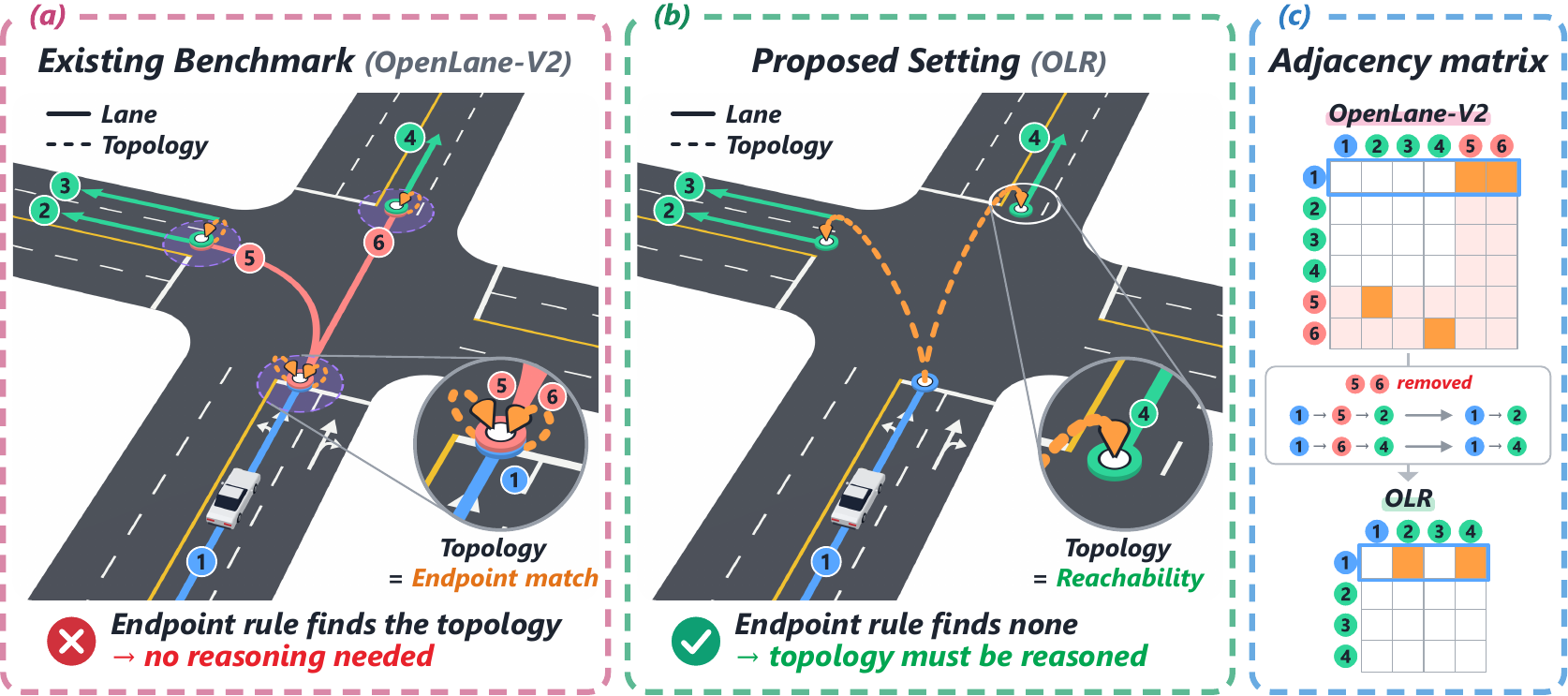}\hspace{6.8bp}%
    \includegraphics{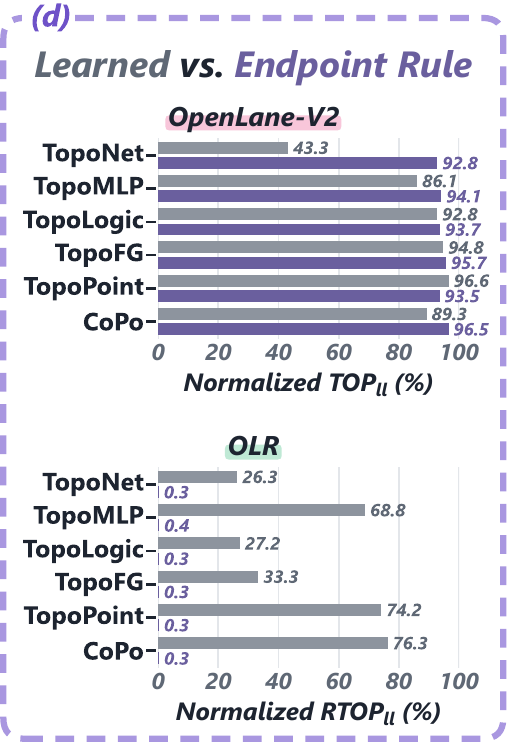}}
  \captionof{figure}{Lane topology at an intersection. (a) OpenLane-V2 annotates the centerlines inside the intersection (lanes 5 and 6), so connected lanes touch and an endpoint rule finds the connections. (b) OpenLane-V2 Reachability (OLR) removes them and links each entry lane directly to the exit lanes it can reach, (c) turning $1{\rightarrow}5{\rightarrow}2$ and $1{\rightarrow}6{\rightarrow}4$ into $1{\rightarrow}2$ and $1{\rightarrow}4$. (d) On connections at intersections, this rule reaches 92.8--96.5\% of the upper bound (the ground-truth connections on each method's detected lanes) on OpenLane-V2, on par with learned topology heads, but 0.3--0.4\% on OLR.}
  \label{fig:teaser}
\end{center}

%% file: sec/0_abstract.tex
\begin{abstract}
Lane topology reasoning infers the relations among lane centerlines and traffic elements detected in multi-view images.
Existing benchmarks, however, annotate centerlines through intersections, so connected lanes always share endpoints, reducing topology reasoning to endpoint matching.
To counter this shortcut, we replace the centerlines inside intersections with direct connections and refer to this setting as OpenLane-V2 Reachability (OLR).
This restores topology reasoning, as the topology score now centers on connections across intersections, which share no endpoints.
We also propose \ours, whose pair head decides each connection from the features and relative pose of the two lanes rather than from endpoint distance.
Its end-aware refinement corrects lane ends, topology-preserving assignment keeps connected lanes matched, and approach-guided association links lanes with traffic elements.
\ours achieves state-of-the-art performance on OLR.
Given the same ground-truth lanes, it still predicts connections with the highest accuracy, so its lead comes from the reachability decision rather than from better lane detection.
We will publicly release OLR and our code.
\end{abstract}
\vfill

%% file: sec/1_intro.tex
\section{Introduction}
\label{sec:intro}

To navigate an intersection, it is essential that an autonomous vehicle know not only where the lanes are but also where each lane leads~\cite{gao2020vectornet,liang2020lanegcn}.
Lane topology reasoning aims to provide this knowledge: it infers the relations among the lane centerlines and traffic elements detected in multi-view images, that is, where each lane leads and which traffic element controls each lane.
These relations are not visible in any single lane or signal; they follow from how the whole scene is laid out.
For example, at an intersection with a left-turn signal, the leftmost lane leads into the left turn and is controlled by that signal, as its position and the signal together imply.
Lane topology reasoning thus requires understanding the scene itself.

OpenLane-V2~\cite{openlanev2} has become the standard benchmark for this task, and a line of methods~\cite{toponet,topomlp,topologic,topofg,topopoint,copo,topohr,topocurve} has steadily raised its topology scores.
Its labels, however, reduce \emph{topology reasoning} to \emph{endpoint matching}: a connection is annotated when the end of one lane contacts the start of the other (\cref{fig:teaser}~(a)), so connected lanes always share an endpoint.
Indeed, once lanes are detected, a simple \emph{endpoint rule}, which links two lanes when the end of one lies within 3\,m of the start of the other, performs on par with the learned topology heads of existing models~\cite{toponet,topomlp,topologic,topofg,topopoint,copo} on the connections at intersections (\cref{fig:teaser}~(d)).
A learned topology head is thus \emph{dispensable} on this benchmark, replaceable by this rule at almost no cost, which means that topology reasoning collapses into lane detection.

To remedy this, we simply drop the centerlines inside intersections, which join each entry lane to its exit lanes at shared endpoints.
We replace each of them with a direct connection, recasting a lane connection from proximity to \emph{reachability} (\cref{fig:teaser}~(b) and (c)).
We refer to this setting as OpenLane-V2 Reachability (OLR).
It reveals how well topology heads actually reason, as they must now infer where each lane leads from the scene: the endpoint rule recovers almost none of the connections across intersections (\cref{fig:teaser}~(d)).
Existing methods can be retrained and compared on OLR without modification, since it keeps the images, splits, and metrics of OpenLane-V2.

We also propose \ours, a model designed around what reachability demands, with four components.
i) \emph{End-Aware Point Refinement} samples bird's-eye-view features along each lane and beyond its ends, where connections attach and detected lanes are least accurate.
ii) A \emph{Relative-Pose Pair Head} scores each candidate connection from the features and relative pose of its two lanes and is trained to choose among competing exits.
iii) \emph{Topology-Preserving Assignment} keeps the lanes that carry connections matched, since a connection is scored only between matched lanes.
iv) \emph{Approach-Guided TE Association} groups lanes into intersection approaches by the predicted connections and corrects lane--TE links per approach.

On OLR, \ours achieves state-of-the-art performance, surpassing the strongest baseline by 4.2 points in the overall score (\rols) and 5.9 points in lane--lane topology.
Given the same ground-truth lanes, so that methods differ only in how they connect them, \ours still predicts connections across intersections with the highest accuracy.
Its lead thus comes from the reachability decision rather than from better lane detection.
In the same test, the endpoint-distance term that several prior methods add to their scores lowers their accuracy.

\paragraph{Contributions.}
Our contributions are summarized with:
\begin{itemize}
  \item We demonstrate that the proximity-based connection labels of OpenLane-V2 reduce lane topology reasoning to endpoint matching. Thus, a simple endpoint rule can replace the learned topology heads of existing methods at negligible cost.
  \item We reformulate OpenLane-V2 as \textbf{OLR}, an evaluation setting that replaces the centerlines inside intersections with direct connections, so that a model must infer where each lane leads from the scene.
  \item We propose \textbf{\ours}, a model designed around reachability, whose four components address lane ends, connection decisions, lane assignment, and lane--TE association.
  It achieves \emph{state-of-the-art} performance on both \textit{subset\_A} and \textit{subset\_B}.
  \item We evaluate every method on the same ground-truth lanes and find that \ours still predicts connections with the highest accuracy.
\end{itemize}

%% file: sec/2_related_work.tex
\section{Related Work}

\subsection{Benchmarks for Lane Topology Reasoning}

Lane detection datasets~\cite{pan2018scnn,garnett2019lanenet,xu2020curvelane,guo2020genlanenet,openlane,yan2022once3dlanes} annotate the geometry of each lane but not how lanes connect or which traffic elements control them.
OpenLane-V2~\cite{openlanev2} adds both relations on top of Argoverse 2~\cite{argoverse2} and nuScenes~\cite{nuscenes}.
As in HD maps~\cite{lanelet2,chang2019argoverse}, it treats each path through an intersection as a lane of its own and joins lanes at shared endpoints.
Recent benchmarks enrich the relations between lanes and traffic elements.
MapDR~\cite{mapdr} ties traffic-sign rules to the lanes they govern, and KITScenes~\cite{kitscenes} ties lanes to signs and lights in Lanelet2 maps.
Neither connects an entry lane of an intersection directly to its exit lanes: MapDR annotates no lane-to-lane links, and KITScenes takes them from its maps.
\mbox{OLR} does so by replacing each centerline inside an intersection with a direct connection.
\subsection{Methods for Lane Topology Reasoning}

Online HD map construction~\cite{li2022hdmapnet,liu2023vectormapnet,liao2023maptr} detects map elements such as lane dividers and road boundaries as vectors, leaving how lanes connect to lane topology methods.
Existing methods differ mainly in the evidence they use to decide whether two lanes connect.
One line relies on learned lane features~\cite{can2021stsu,can2022tplr}.
TopoNet~\cite{toponet} passes messages between lanes and traffic elements over a scene graph, and TopoMLP~\cite{topomlp} scores lane pairs with MLP heads on strong detections.
Later work enriches what these heads read, with lane segments~\cite{lanesegnet}, instance masks~\cite{topomaskv2}, or B\'ezier deformable attention~\cite{topobda}.
Further evidence comes from 2D lane detections~\cite{topo2d} or a prior road-structure map~\cite{hgeo}.
Others strengthen the heads themselves with iterative topology refinement~\cite{cgnet}, train them with redundant supervision~\cite{ratopo}, or enhance them with a plug-in framework~\cite{topoenhance}.
Their topology scores, however, largely follow detection quality~\cite{topomlp}.

A second line builds endpoint geometry into the decision.
TopoLogic~\cite{topologic} adds the distance from one lane's end to another's start to the learned similarity.
Later methods use this distance as an attention bias~\cite{topoformer}, compute it between boundary points~\cite{topofg}, or encode it in pair embeddings~\cite{copo}.
TopoHR~\cite{topohr} initializes its topology relations with this distance, and UniTopo~\cite{unitopo} draws each connection as one curve through both lanes.
Others make the endpoints explicit, detecting them separately~\cite{topopoint} or anchoring curves at them~\cite{topocurve}.
All of these cues assume that connected lanes meet, as they do in OpenLane-V2.
On \mbox{OLR}, an intersection separates connected lanes, so our pair head scores each candidate connection from the features and relative pose of its two lanes.

%% file: sec/3_dataset.tex
\section{OpenLane-V2 Reachability (OLR)}
\label{sec:evaluation}

\subsection{OpenLane-V2 Fails to Capture Reachability}
\label{sec:olv2_limits}

Topology reasoning judges the relations between road elements, such as
whether one lane leads into another, while recovering the position and shape
of a lane belongs to detection~\cite{openlanev2}. In OpenLane-V2, two
lanes are connected when the end of one meets the start of the other.
This fits non-intersection roads, where each lane ends where the next one begins. At an intersection, however, a lane entering it
(an entry lane) ends on one side, and the lanes leaving it (exit lanes) start
on the other. A topology reasoning model must decide which of the many exit lanes an entry lane can reach. OpenLane-V2 joins an entry lane to its exits only through centerlines drawn
across the intersection, which undermines topology reasoning in three ways. A connection counts only if its centerline is drawn accurately, a simple endpoint rule already saturates the current connection score, and the labels teach a model only which lanes meet, not which exits a lane can reach across an intersection.

\begin{figure}[t]
  \centering
  \includegraphics[width=\linewidth]{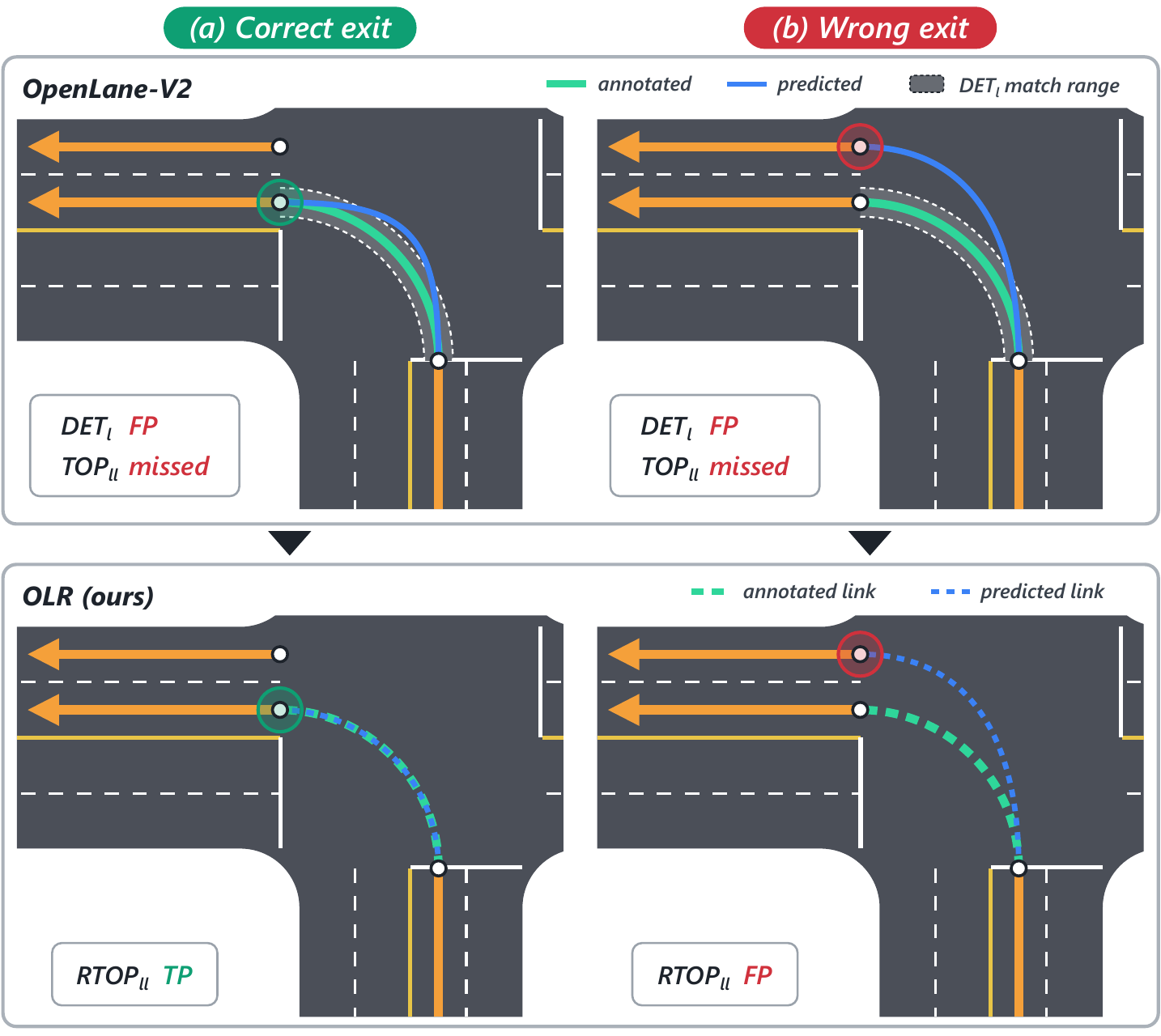}
  \caption{Scoring a connection across an intersection. (a)~A prediction reaches the correct exit, but its centerline leaves the DET$_l$ match range of the annotated one. (b)~A prediction reaches the wrong exit. OpenLane-V2 scores both the same: DET$_l$ counts each centerline as a false positive, so TOP$_{ll}$ misses the annotated link. OLR has no intersection centerlines and scores the link itself with \rtop{}, crediting (a) and penalizing (b).}
  \label{fig:centerline_example}
\end{figure}

\paragraph{Connections depend on detection.}
By OpenLane-V2's own account, centerlines are invisible trajectories derived from lane markings, which a crossroad typically lacks~\cite{openlanev2}. Their shape inside an intersection therefore has no visual cue. Even between the same entry and exit lanes, vehicles spread out over many paths~\cite{alhajyaseen2011,zhao2023}, so no single curve is the correct one. A connection across an intersection counts only if its centerline is drawn
accurately enough to be matched. Suppose a model predicts the correct exit but draws the centerline inaccurately. The lane detection metric DET$_l$ counts it as a false positive (\cref{fig:centerline_example}~(a), first row), just like a centerline connected to the wrong exit (\cref{fig:centerline_example}~(b), first row). The lane-lane connectivity metric TOP$_{ll}$ scores only matched lanes, so it ignores the link even though it is correct. The outcome is thus decided by drawing lanes rather than by deciding lane connection, yet the shape
being drawn is poorly defined.

\paragraph{TOP$_{ll}$ measures proximity, not reasoning.}
Even when the centerlines are matched, TOP$_{ll}$ can be passed without reasoning about where a lane leads. Matched predictions of connected lanes end near each other, so a learning-free rule that links predicted lanes whose ends lie close already reaches 92.8--96.5\% of the upper bound on intersection connections (\cref{fig:teaser}~(d)). The rule links two lanes when the end of one lies within 3\,m of the start of the other, the largest lane-matching threshold. The upper bound keeps the same predicted lanes but replaces their connections with the ground-truth ones. TOP$_{ll}$ thus rewards lanes whose ends lie close (proximity), not a decision about where a lane leads (reasoning).

\paragraph{OpenLane-V2 cannot teach where a lane leads.} Since connected lanes always meet, a model trained on OpenLane-V2 is asked only about this contact, never about which exits a vehicle in a lane can reach. Indeed, in the same \cref{fig:teaser}~(d), the connections that models learn rarely beat this endpoint rule, adding little beyond endpoint contact. Where an entry lane can go is labeled only by the intersection centerlines drawn from it, so a model learns this only by fitting the points of curves that no road marking guides. Nothing in OpenLane-V2 states directly that an entry lane can reach a given exit.

\begin{figure*}[t]
  \centering
  \includegraphics[width=\linewidth]{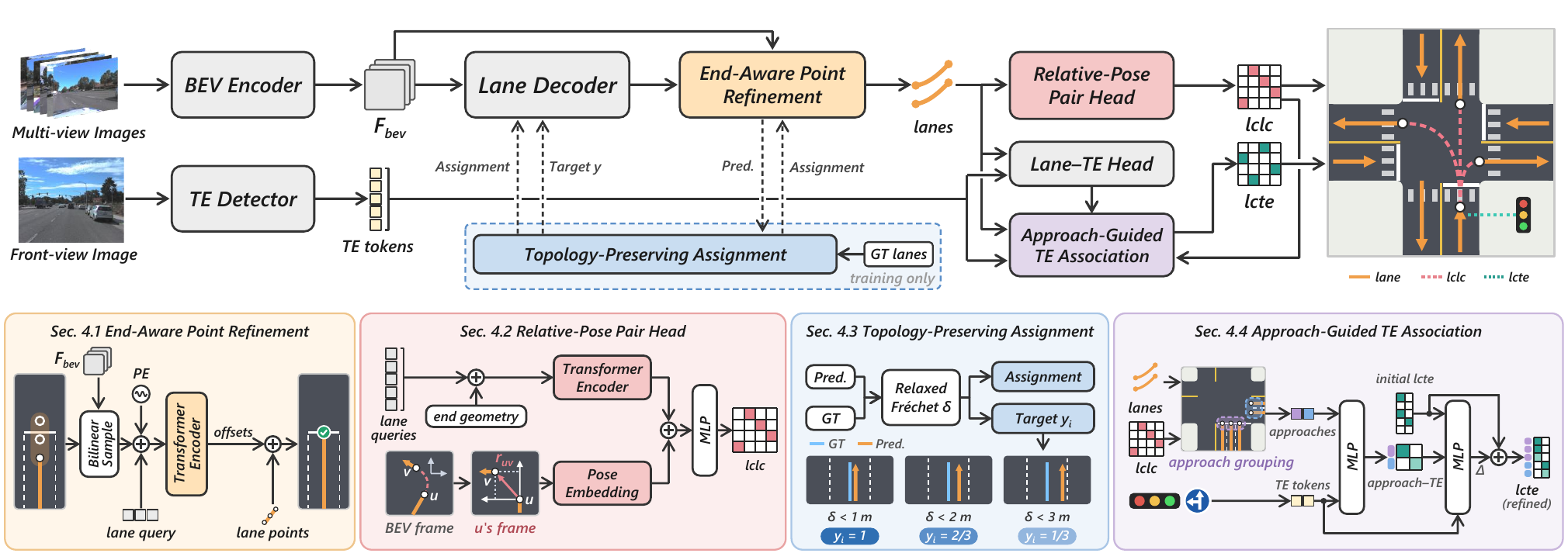}
  \caption{Overview of \ours. Top: lanes decoded from BEV features are refined at their ends, the pair head predicts lane--lane connections, and approach-guided TE association corrects the lane--TE scores of the detector from these connections; during training, topology-preserving assignment assigns the decoded and refined lanes to the ground truth and sets the classification targets $y$ of the decoder. Bottom: details of each component (\cref{sec:refine,sec:pair,sec:tpa,sec:approach}).}
  \label{fig:arch}
\end{figure*}

\subsection{From Proximity to Reachability}
\label{sec:olv2r}

All three problems in \cref{sec:olv2_limits} come from the centerlines inside intersections, so the fix lies in how a connection is defined, not in how it is scored. Therefore, we remove these centerlines, found from the intersection labels of
the source maps, and directly connect the lanes they joined (\cref{fig:centerline_example}, second row). We call this OpenLane-V2 Reachability (OLR).

Removing them keeps the reachability decision out of DET$_l$ and leaves only lanes that end where lane markings end. The paths through intersections are kept as unscored attributes of the connections. Lane--TE links to the removed centerlines move to the remaining lanes, a traffic light to the entry lanes facing it and a sign or road marking to the lanes on its side (see the supplementary material). Every connection across an intersection now states directly that an entry lane reaches an exit, so models learn reachability from the labels, and TOP$_{ll}$, denoted \rtop{} on OLR, measures it directly.

\paragraph{Designing for reachability.}
OLR asks which exits each entry lane can reach across an intersection. Since connected lanes now lie apart, this must be decided without relying on endpoint distance. A connection, moreover, counts only between lanes matched to the ground truth. A predicted lane is matched only if it stays close to the ground truth along its whole length, including its ends. Evaluation matches the most confident predictions first. The connections of a lane therefore count only if its most confident prediction is accurate up to its ends. Lane--TE links, in turn, rest on the same decision. Which lanes a traffic light or sign governs follows from which entry lanes come from the same side of the intersection and where each turns, both of which the connections reveal.

%% file: sec/4_method.tex
\section{TopoReach}

\label{sec:method}
\ours follows the basic design of lane topology reasoning models, in which a lane decoder detects lanes and a separate topology head predicts their connections.
A ResNet-50 backbone~\cite{he2016deep} extracts features from the multi-view images, and a BEV encoder~\cite{li2022bevformer} turns them into bird's-eye-view (BEV) features $\mathbf{F}_{\mathrm{bev}}$.
The lane decoder has a standard DETR-style structure~\cite{carion2020detr}: in each layer, learnable lane queries $\mathbf{q}$ pass through self-attention and deformable cross-attention~\cite{zhu2021deformable} to $\mathbf{F}_{\mathrm{bev}}$, and regress lanes $\ell$ as ordered 3D points.
A separate DETR-style detector finds TEs in the front-view image, kept at full resolution as in recent methods~\cite{topopoint,topocurve}.
On top of this structure, our model adds four components to meet what OLR demands (\cref{sec:evaluation,fig:arch}), each described in the following subsections.

\subsection{End-Aware Point Refinement}
\label{sec:refine}
The lane decoder often misses where a lane ends, as it reads $\mathbf{F}_{\mathrm{bev}}$ only around a single reference point per query.
Such misses matter greatly at intersections, where the end of an entry lane anchors a connection to every exit lane it can reach.
If that end is misplaced, the lane can lose all of these connections at once.

To pinpoint each end, End-Aware Point Refinement reads $\mathbf{F}_{\mathrm{bev}}$ again along the lane and beyond both of its ends.
For a lane $\ell$ decoded from query $\mathbf{q}$, the set $\mathcal{P}$ holds the 11 points of $\ell$ and two more points on its straight continuation beyond each end.
Each $\mathbf{p}\in\mathcal{P}$ becomes a token
\begin{equation}
  \mathbf{x}_{\mathbf{p}} = \mathrm{Bilinear}(\mathbf{F}_{\mathrm{bev}}, \mathbf{p}) + \mathrm{PE}(\mathbf{p}) + \mathbf{q},
  \label{eq:refine}
\end{equation}
where $\mathrm{Bilinear}$ samples the BEV feature at $\mathbf{p}$, $\mathrm{PE}$ is a learned positional encoding of the coordinates of $\mathbf{p}$ and its index along the lane, and $\mathbf{q}$, the same for all tokens of the lane, carries the decoder's representation of it.
A transformer encoder~\cite{vaswani2017attention} $f$ lets the tokens of the lane attend to one another, and a linear layer regresses the offsets $\Delta\ell$ of its 11 points from the outputs $\mathbf{o}_{\mathbf{p}}$ of $f$:
\begin{equation}
  \{\mathbf{o}_{\mathbf{p}}\}_{\mathbf{p}\in\mathcal{P}} = f(\{\mathbf{x}_{\mathbf{p}}\}_{\mathbf{p}\in\mathcal{P}}),\enskip
  \Delta\ell = \{\mathrm{Linear}(\mathbf{o}_{\mathbf{p}})\}_{\mathbf{p}\in\ell}.
  \label{eq:refine_out}
\end{equation}
In \cref{eq:refine_out}, only the points of $\ell$ receive offsets, and the four added points serve as context in $f$.
The refined lane $\ell' = \ell + \Delta\ell$ replaces $\ell$ in all later steps.
Both $\ell$ and $\ell'$ receive the lane loss, $\mathcal{L}_{\mathrm{ref}}$ on $\ell'$, in which the two end points have twice the loss weight of the other points.

\subsection{Relative-Pose Pair Head}
\label{sec:pair}
Existing heads add the distance between lane endpoints as a cue~\cite{topologic,topofg,topopoint,copo}.
Across an intersection, however, this distance cannot tell which exit lanes $v$ an entry lane $u$ reaches, as every exit lane starts on the far side (\cref{fig:teaser}~(d)).
Rather, the exits differ in the turn from $u$ to $v$, the heading change from the end of $u$ to the start of $v$: straight, left, right, or U-turn.
Road markings on $u$, such as turn arrows, and the layout of the surrounding lanes show which turns $u$ takes and which exit of each turn it enters.
The Relative-Pose Pair Head thus builds a token for each lane from its decoder query, which carries the features of the lane read from $\mathbf{F}_{\mathrm{bev}}$, and lets the tokens attend to all other lanes for the layout.
It then scores each pair from the tokens of the two lanes and their relative pose, which captures the turn.

On the ground plane of the ego-vehicle frame, each lane $u$ starts at $\mathbf{s}_u$ in direction $\mathbf{d}^{\mathrm{s}}_u$ and ends at $\mathbf{e}_u$ in direction $\mathbf{d}^{\mathrm{e}}_u$.
It enters the head as a token
\begin{equation}
  \mathbf{x}_u = \mathbf{q}_u + \phi\big(\mathbf{s}_u, \mathbf{e}_u, \mathbf{d}^{\mathrm{s}}_u, \mathbf{d}^{\mathrm{e}}_u\big),
  \label{eq:pair_token}
\end{equation}
where $\mathbf{q}_u$ is its query after the last decoder layer and $\phi$ is an MLP on the sinusoidal encodings~\cite{mildenhall2020nerf} of the points and on the directions.
A transformer encoder $f'$ lets the tokens $\mathbf{x}_u$ of all lanes attend to one another:
\begin{equation}
  \{\mathbf{o}_u\} = f'(\{\mathbf{x}_u\}).
  \label{eq:pair_enc}
\end{equation}
For a pair $(u, v)$, the start of $v$ in the frame at the end of $u$ and the end of $u$ in the frame at the start of $v$ are
\begin{equation}
  \mathbf{r}_{uv} = \mathbf{R}(\mathbf{d}^{\mathrm{e}}_u)^{\top}(\mathbf{s}_v-\mathbf{e}_u), \quad
  \mathbf{r}_{vu} = \mathbf{R}(\mathbf{d}^{\mathrm{s}}_v)^{\top}(\mathbf{e}_u-\mathbf{s}_v),
  \label{eq:relpose}
\end{equation}
where $\mathbf{R}(\mathbf{d})$ rotates the $x$-axis onto $\mathbf{d}$.
Together with the turn angle $\Delta_{uv}$ from $\mathbf{d}^{\mathrm{e}}_u$ to $\mathbf{d}^{\mathrm{s}}_v$, they form the relative pose, which describes the turn independently of where the pair lies in the scene.
Finally, the logit $z_{uv}$ for the connection from $u$ to $v$ is
\begin{equation}
  z_{uv} = \mathrm{MLP}\big(\mathbf{W}_q\mathbf{o}_u + \mathbf{W}_k\mathbf{o}_v + g(\mathbf{r}_{uv}, \mathbf{r}_{vu}, \Delta_{uv})\big),
  \label{eq:score}
\end{equation}
where $\mathbf{W}_q$ and $\mathbf{W}_k$ are linear projections.
The pose embedding $g$ is the sum of two MLPs, one taking the sinusoidal encoding of $\mathbf{r}_{uv}$ and the other that of $\mathbf{r}_{vu}$, each together with the cosine and sine of $\Delta_{uv}$.
Over all pairs of predictions assigned to ground-truth lanes, these logits are trained with a binary cross-entropy (BCE) loss $\mathcal{L}_{\mathrm{pair}}$.

\subsection{Topology-Preserving Assignment}
\label{sec:tpa}
DETR-style training makes the decoder represent each lane by a single confident prediction.
The training, however, picks this prediction by Hungarian matching~\cite{kuhn1955hungarian} with a point-wise L1 cost, which averages over the points and hardly changes when one end is misplaced.
The prediction that represents a lane can thus anchor its connections at a missing end (\cref{sec:refine}), while a duplicate with correct ends is suppressed.
Evaluation then rejects this prediction together with all of its connections (\cref{sec:evaluation}).
It accepts a prediction for its nearest ground-truth lane only if their relaxed Fr\'echet distance $\delta$~\cite{eiter1994discrete,openlanev2}, set by the largest gap between the two lanes, is below a threshold $\tau$ of 1, 2, or 3\,m.

Topology-Preserving Assignment picks the prediction that represents each lane by $\delta$ throughout training.
Hungarian matching uses $\delta$ in place of the L1 cost, so the pair head learns connections on queries close to their lanes under $\delta$.
For the classification targets, each query $i$, predicting a lane $\ell_i$ with confidence $\hat{y}_i$, goes to its nearest ground-truth lane $\bar{\ell}$ under $\delta$, as in evaluation.
Of the queries that go to $\bar{\ell}$, the closest, $i^{*}$, is its representative.
A representative's target $y_{i^{*}}$ is the share of the thresholds at which it would be accepted if ranked first in confidence, and every other query's target is 0:
\begin{equation}
  y_{i} = \begin{cases}\dfrac{1}{|\mathcal{T}|}\displaystyle\sum_{\tau\in\mathcal{T}} \mathbf{1}\big[\delta(\ell_i, \bar{\ell}) < \tau\big] & \text{if } i=i^{*},\\
  0 & \text{otherwise,}\end{cases}
  \label{eq:tpa}
\end{equation}
where $\mathcal{T}=\{1,2,3\}$\,m is the set of thresholds and $\mathbf{1}[\cdot]$ is 1 if its condition holds and 0 otherwise.
The confidences are trained on these targets with the focal loss~\cite{lin2017focal} for soft targets~\cite{li2020gfl},
\begin{equation}
  \begin{aligned}
  \mathcal{L}_{\mathrm{cls}} &= \frac{1}{N}\sum_{i} w_i\,\mathrm{BCE}(\hat{y}_i, y_i),\\
  w_i &= \begin{cases}\alpha\,|y_i-\hat{y}_i|^{\gamma} & \text{if } y_i>0,\\ (1-\alpha)\,\hat{y}_i^{\gamma} & \text{otherwise,}\end{cases}
  \end{aligned}
  \label{eq:cls}
\end{equation}
where $N$ is the number of ground-truth lanes and $\alpha$, $\gamma$ are the parameters of the focal loss.
The prediction that represents each lane is thus the one that follows it best under $\delta$, ends included, and carries its connections, with no added parameters.

\input{tables/main}

\subsection{Approach-Guided TE Association}
\label{sec:approach}
The detector's lane--TE head, like those of existing methods~\cite{toponet,topomlp,topologic}, scores each lane--TE pair from the lane and TE features alone.
Scored pair by pair, however, the entry lanes on one side of the intersection can disagree on the same traffic light.
Which TEs govern a lane depends on that side, its \emph{approach}.
For example, a traffic light governs every entry lane that faces it (\cref{sec:olv2r}).

The predicted connections reveal the approaches, since a connection across the intersection marks an entry lane and its turn.
Approach-Guided TE Association groups the entry lanes into approaches from these connections and lets each approach inform the lane--TE scores of its lanes.
For an entry lane $u$, the turn profile holds, per turn (\cref{sec:pair}), the highest probability $\sigma(z_{uv})$ among its connections to exits $v$ of that turn.
Entry lanes whose ends lie close together and point the same way form an approach $A$.
For a traffic element $t$, an MLP scores whether $t$ governs $A$:
\begin{equation}
  a_{At} = \mathrm{MLP}\big(\phi_A(A) + \phi_T(t)\big),
  \label{eq:app_aux}
\end{equation}
where the MLP $\phi_A$ embeds the geometry of $A$ with the turn profiles of its lanes, and the MLP $\phi_T$ the box and class of $t$.
A BCE loss $\mathcal{L}_{\mathrm{aux}}$ trains $a_{At}$ to predict whether any lane of $A$ is linked to $t$.

Each lane $u$ of $A$ then corrects the detector's lane--TE logit $z^{\mathrm{lt}}_{ut}$:
\begin{equation}
  \hat{z}^{\mathrm{lt}}_{ut} = z^{\mathrm{lt}}_{ut} + h\big(\phi_C(\mathbf{c}_{ut}) + \phi_T(t)\big),
  \label{eq:app_corr}
\end{equation}
where $h$ and $\phi_C$ are MLPs, the latter embedding the cues $\mathbf{c}_{ut}$ from $A$ and from $u$.
From $A$, it takes $a_{At}$ and the detector's highest and mean scores of $t$ over the lanes of $A$.
From $u$, it takes the fit of the turn profile of $u$ to the class of $t$, the lateral position and role of $u$ in $A$, and $z^{\mathrm{lt}}_{ut}$ itself.
The last layer of $h$ starts at zero, so \cref{eq:app_corr} begins with $\hat{z}^{\mathrm{lt}}_{ut} = z^{\mathrm{lt}}_{ut}$.
Trained with $\mathcal{L}^{\mathrm{app}}_{\mathrm{lt}}$, the detector's lane--TE loss on the label of each lane, $\hat{z}^{\mathrm{lt}}$ lets the approach inform the score of a lane without imposing a link on it.
The approaches are built without gradient and need no labels beyond lanes, connections, and lane--TE links.

\subsection{Training}
\label{sec:train}
\ours is trained with the total loss
\begin{equation}
  \mathcal{L} = \mathcal{L}_{\mathrm{det}} + \mathcal{L}_{\mathrm{ref}} + \mathcal{L}_{\mathrm{pair}} + \mathcal{L}^{\mathrm{app}}_{\mathrm{lt}} + \mathcal{L}_{\mathrm{aux}},
  \label{eq:loss}
\end{equation}
where $\mathcal{L}_{\mathrm{det}}$ holds the lane, TE, and lane--TE losses of the detector, trained with hybrid matching~\cite{jia2023hdetr} as in TopoMLP~\cite{topomlp}.
The refinement adds $\mathcal{L}_{\mathrm{ref}}$, the lane loss of $\ell'$ (\cref{sec:refine}), and the pair head adds $\mathcal{L}_{\mathrm{pair}}$, the BCE on the pair logits summed over the encoder layers (\cref{sec:pair}).
The TE association adds $\mathcal{L}^{\mathrm{app}}_{\mathrm{lt}}$, the lane--TE loss on $\hat{z}^{\mathrm{lt}}$ of \cref{eq:app_corr}, and $\mathcal{L}_{\mathrm{aux}}$, the BCE on $a_{At}$ of \cref{eq:app_aux}.
The pair head and the TE association work on the detected lanes, so epochs 1--12 train only $\mathcal{L}_{\mathrm{det}}$ and $\mathcal{L}_{\mathrm{ref}}$.
Epochs 13--24 add the other three terms and switch the assignment and lane classification of the detector to those of \cref{sec:tpa}, with the gradient from the pair head into the detector scaled down (see the supplementary material).

%% file: tables/main.tex
\begin{table*}[t]
  \centering
  \small
  \setlength{\tabcolsep}{4pt}
  \begin{tabularx}{\textwidth}{l|l|Y|YY|GY|Y|YY|GY}
      \toprule
      \multirow{2}{*}[-0.6ex]{Method} & \multirow{2}{*}[-0.6ex]{Venue} & \multicolumn{5}{c|}{\textit{subset\_A}} & \multicolumn{5}{c}{\textit{subset\_B}} \\
      \hhline{~~----------}
        &  & \rule{0pt}{2.7ex}\rols & DET$_l$ & DET$_t$ & \ghead{\rtop{}} & TOP$_{lt}$ & \rols & DET$_l$ & DET$_t$ & \ghead{\rtop{}} & TOP$_{lt}$ \\
      \midrule
      TopoNet~\cite{toponet} & arXiv'23 & 39.7 & 31.1 & 44.9 & 11.3 & 24.4 & 38.0 & 26.3 & 52.1 & 10.1 & 17.5 \\
      TopoMLP~\cite{topomlp} & ICLR'24 & 44.2 & 29.8 & 51.2 & 19.7 & 26.4 & 43.9 & 28.9 & 58.9 & 18.5 & 20.1 \\
      TopoLogic~\cite{topologic} & NeurIPS'24 & 40.4 & 31.1 & 44.0 & 13.0 & 25.5 & 40.1 & 28.6 & 54.3 & 11.8 & 18.5 \\
      TopoFG~\cite{topofg} & AAAI'26 & 42.4 & \underline{32.1} & 44.4 & 16.6 & 27.4 & 40.8 & 27.0 & 46.8 & 18.0 & 22.1 \\
      TopoPoint~\cite{topopoint} & NeurIPS'25 & 46.9 & 29.9 & \underline{53.3} & 24.1 & 30.8 & 46.3 & 27.4 & \underline{60.5} & 22.7 & 24.5 \\
      CoPo~\cite{copo} & KDD'26 & \underline{47.2} & 31.9 & 47.4 & \underline{26.8} & \underline{33.5} & \underline{47.5} & \underline{31.3} & 59.8 & \underline{24.3} & \underline{24.7} \\
      \midrule
      \ours (Ours) & -- & \textbf{51.4} & \textbf{33.7} & \textbf{53.6} & \textbf{32.7} & \textbf{37.3} & \textbf{51.4} & \textbf{33.1} & \textbf{62.8} & \textbf{32.0} & \textbf{28.2} \\
      \bottomrule
  \end{tabularx}
  \caption{Main results on OLR. Best in bold, second best underlined. The shaded column is \rtop{}, the metric that OLR redefines.}
  \label{tab:main}
\end{table*}

%% file: sec/5_experiments.tex
\section{Experiments}
\label{sec:exp}

\subsection{Experimental Setup}
\label{sec:setup}
\input{figs/qualitative/v63/figure_olv2r}

\paragraph{Datasets and metrics.}
We use \textit{subset\_A} and \textit{subset\_B} of OpenLane-V2~\cite{openlanev2}, built on Argoverse~2~\cite{argoverse2} and nuScenes~\cite{nuscenes}, respectively, with the labels of OLR, which keeps their images and splits (\cref{sec:olv2r}).
DET$_l$ and DET$_t$ are the average precisions of lane and TE detection, the former averaged over the lane-matching thresholds of 1, 2, and 3\,m~\cite{openlanev2}.
\rtop{} and TOP$_{lt}$ are those of the lane--lane connections and lane--TE links between matched elements.
The overall score is
\begin{equation}
  \medmuskip=2mu \thickmuskip=3mu \mathrm{ROLS} = \frac{\mathrm{DET}_{l} + \mathrm{DET}_{t} + \sqrt{\rtop{}} + \sqrt{\mathrm{TOP}_{lt}}}{4}.
  \label{eq:ols}
\end{equation}
\rols is not comparable with OLS on OpenLane-V2, since \rtop{} scores different lane pairs from TOP$_{ll}$.
All metrics are computed on the validation sets with the official evaluation code.

\paragraph{Baselines and implementation.}
We compare with TopoNet~\cite{toponet}, TopoMLP~\cite{topomlp}, TopoLogic~\cite{topologic}, TopoFG~\cite{topofg}, TopoPoint~\cite{topopoint}, and CoPo~\cite{copo}, each retrained on OLR from its released code with its official settings.
All methods use a ResNet-50 backbone and are trained for 24 epochs.
Retraining in this way is not possible for TopoHR~\cite{topohr}, TopoCurve~\cite{topocurve}, and TopoBDA~\cite{topobda}, whose code is not publicly available, or for TopoFormer~\cite{topoformer}, whose training relies on additional labels and a pre-trained lane detector that are not released.
    As in TopoPoint~\cite{topopoint}, \ours downscales the multi-view images by half for the BEV features and keeps the front-view image at full resolution for TE detection.
Further details are given in the supplementary material.

\subsection{Main Results and Analysis}
\label{sec:main}

\paragraph{Quantitative results.}
\cref{tab:main} compares \ours with prior methods on both subsets of OLR.
\ours outperforms all prior methods in every metric on both subsets and
reaches 51.4 \rols on each, 4.2 and 3.9 points above the strongest prior
method, CoPo (47.2 and 47.5).
On OpenLane-V2 \textit{subset\_A}, \ours also achieves state-of-the-art performance (see the supplementary material).

On \textit{subset\_A}, \ours leads by 1.6 points in DET$_l$ (33.7 vs.\ 32.1
for TopoFG) and 0.3 in DET$_t$. Its lead grows to 3.8 points in TOP$_{lt}$
(37.3 vs.\ 33.5 for CoPo) and 5.9 in \rtop{} (32.7 vs.\ 26.8). The gain thus comes mainly from the reachability decision rather than from lane detection. The same holds on \textit{subset\_B}, where the \rtop{} margin grows to 7.7 points (32.0 vs.\ 24.3).

The ranking of prior methods also flips from OpenLane-V2 to OLR. TopoLogic and TopoFG,
which add endpoint distance to their topology heads, lead TopoMLP in
TOP$_{ll}$ on OpenLane-V2 by 2.3 and 6.6 points (see the supplementary
material) but trail it in \rtop{} by
6.7 and 3.1 points on \textit{subset\_A}. The 13.1-point lead of TopoLogic
over TopoNet likewise drops to 1.7. As \cref{sec:olv2_limits} predicts, endpoint distance helps on OpenLane-V2 because connected lanes share an endpoint. On OLR, connected lanes are far apart across an intersection, so it helps little.

As intended in \cref{sec:olv2_limits}, detecting lanes well does not guarantee connecting them well on OLR. TopoFG has the highest DET$_l$ among prior methods on \textit{subset\_A} (32.1) but only the fourth-highest \rtop{} (16.6), whereas TopoPoint, second-lowest in DET$_l$ (29.9), is second in \rtop{} (24.1).

\paragraph{Qualitative results.}
\label{sec:qual}
\cref{fig:qual} compares the predictions of all methods on two intersection frames of OLR. In the first row, \ours recovers nearly all connections, whereas prior methods miss many (orange) and some link lanes to unreachable exits (red). The gap widens in the second row, a more complex intersection with more lanes and connections, where prior methods make more errors while \ours still keeps nearly all connections correct. With similar centerlines across methods, the difference lies mainly in the connection decision.

\subsection{Reachability Given Ground-Truth Lanes}
\label{sec:gtdet}
\begin{figure}[t]
  \centering
  \includegraphics[width=\linewidth]{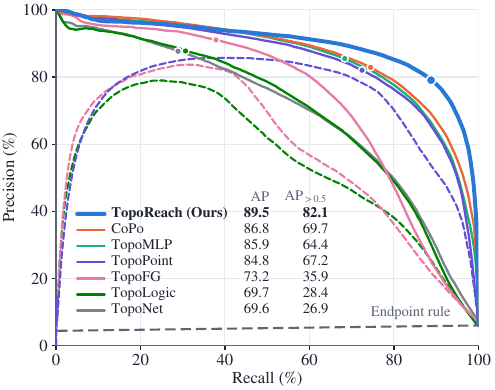}
  \caption{Precision--Recall curve of connections across an intersection with ground-truth lanes (OLR \textit{subset\_A} val). Solid: learned scores; dashed: TopoLogic, TopoFG, and TopoPoint as released, whose distance term makes any two lanes sharing an endpoint look connected. AP: area under each solid curve; AP$_{>0.5}$: its part left of the dot.}
  \label{fig:gtdet}
\end{figure}

\rtop{} counts a connection only between matched lanes, so \cref{fig:gtdet} isolates the reachability decision by giving every method the ground-truth lanes of OLR.
Each ground-truth lane takes the query assigned to it by Hungarian matching, with its coordinates replaced by the ground truth and its score set to 1.
All other queries are dropped, and the method's own head predicts the connections.
The positives are the ground-truth connections across an intersection and the negatives all other lane pairs, except those joined by a touching connection, as at a split or a merge.

The endpoint rule of \cref{sec:olv2_limits} recovers 0.3\% of these connections, since an entry lane and the exits it reaches do not touch.
TopoLogic, TopoFG, and TopoPoint add a fixed function of the end-to-start distance to the learned score~\cite{topologic,topofg,topopoint}.
On OpenLane-V2, this term alone raises the TOP$_{ll}$ of a trained TopoNet from 10.9 to 22.3~\cite{topologic}.
Here, it lowers their AP by 12--16 points (dashed curves in \cref{fig:gtdet}), and still by 2--5 when touching connections are counted as positives.
Without it, TopoLogic ranks the connections as TopoNet does (AP 69.7 and 69.6).
CoPo instead encodes the distance in its pair embedding~\cite{copo}, so its head learns how much to rely on it.

\ours has the highest AP (89.5, against 86.8 for CoPo) and AP$_{>0.5}$ (82.1, against 69.7).
The second gap, over the scores that \rtop{} counts, is larger because \ours scores 88.8\% of the connections above 0.5, against 74.5\% for CoPo.
Its precision is also the highest from a recall of 43\% on, 77.9\% against 67.0\% for CoPo at a recall of 90\%.
Every method receives the same lanes, so this lead comes from the reachability decision, not from where the lanes are placed.

\subsection{Ablation Study}
\label{sec:ablation}
\input{tables/ablation}

To measure the contribution of each component of \ours, we add them one at a time on \textit{subset\_A} (\cref{tab:ablation}). The baseline shares our detector but uses the lane--lane head of TopoNet~\cite{toponet}.

\paragraph{End-Aware Point Refinement.}
Adding End-Aware Point Refinement with its doubled end-point loss weight (\cref{sec:refine}) raises DET$_l$ from 29.9 to 33.2, the largest detection gain in the table. \rtop{} and TOP$_{lt}$ also rise by 1.2 and 1.6 points even though the topology heads are unchanged. As \cref{sec:refine} anticipates, more accurate endpoints let more lanes be matched, so more connections are counted.

\paragraph{Relative-Pose Pair Head.}
Adding the pair head (\cref{sec:pair}) raises \rtop{} by 14.1 points, from 13.9 to 28.0. This roughly doubles the score and is the largest gain of any component, while DET$_l$ and DET$_t$ stay within 0.4 points. Since lane detection barely changes, the gain comes from the connection decision itself. The pair head alone accounts for over two thirds of the total \rtop{} gain from the baseline to the full model, making it the core of \ours. It scores a pair from the relative pose of the two lanes which describes the turn.

\paragraph{Topology-Preserving Assignment.}
Topology-Preserving Assignment changes only how predictions are matched and scored in training, adding no parameters (\cref{sec:tpa}), yet raises \rtop{} by a further 4.6 points (28.0 to 32.6) and DET$_l$ by 0.9 (32.9 to 33.8). Evaluation matches predictions in order of confidence, and Topology-Preserving Assignment trains the prediction closest to each lane under the evaluation distance to be the most confident. Connections are thus learned on the predictions that evaluation matches, so more correct connections are counted.

\paragraph{Approach-Guided TE Association.}
Letting the lane--TE score of each lane draw on its approach (\cref{sec:approach}) raises TOP$_{lt}$ by 4.2 points (33.1 to 37.3), with the other metrics within 0.5 points. Entry lanes on one side of an intersection share a traffic light, and a lane’s turn profile shows which signs apply, so scoring with both links traffic elements more consistently. Since approaches and turn profiles come from the predicted connections, this gain rests on predicting reachability well.

%% file: figs/qualitative/v63/figure_olv2r.tex
\begin{figure*}[!t]
\centering
\resizebox{\linewidth}{!}{%
\setlength{\unitlength}{1bp}%
\begin{picture}(3673.50,1750)
\put(411.231456,1693){\makebox[0bp][c]{\fontsize{60}{64}\selectfont Multi-view images}}
\put(1008.79023,1693){\makebox[0bp][c]{\fontsize{60}{64}\selectfont GT}}
\put(1364.704866,1693){\makebox[0bp][c]{\fontsize{52}{56}\selectfont\bfseries TopoReach}}
\put(1720.619502,1693){\makebox[0bp][c]{\fontsize{52}{56}\selectfont CoPo~\cite{copo}}}
\put(2076.534138,1693){\makebox[0bp][c]{\fontsize{52}{56}\selectfont TopoFG~\cite{topofg}}}
\put(2432.448774,1693){\makebox[0bp][c]{\fontsize{52}{56}\selectfont TopoPoint~\cite{topopoint}}}
\put(2788.36341,1693){\makebox[0bp][c]{\fontsize{52}{56}\selectfont TopoLogic~\cite{topologic}}}
\put(3144.278046,1693){\makebox[0bp][c]{\fontsize{52}{56}\selectfont TopoMLP~\cite{topomlp}}}
\put(3500.192682,1693){\makebox[0bp][c]{\fontsize{52}{56}\selectfont TopoNet~\cite{toponet}}}
\put(835.482912,885.16){\includegraphics[trim=0bp 30.164672000bp 17.600000000bp 51.848204800bp,clip,width=346.614636bp]{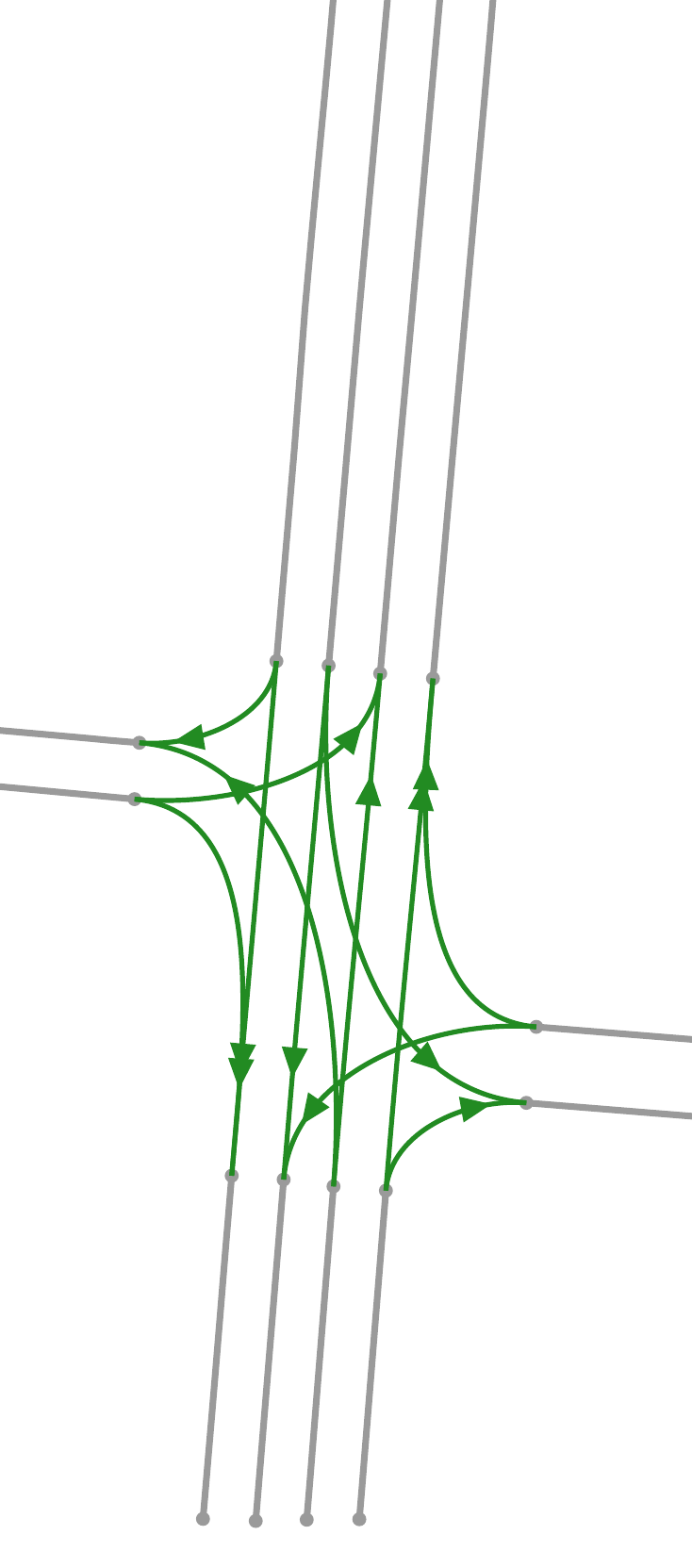}}
\put(1191.397548,885.16){\includegraphics[trim=0bp 30.164672000bp 17.600000000bp 51.848204800bp,clip,width=346.614636bp]{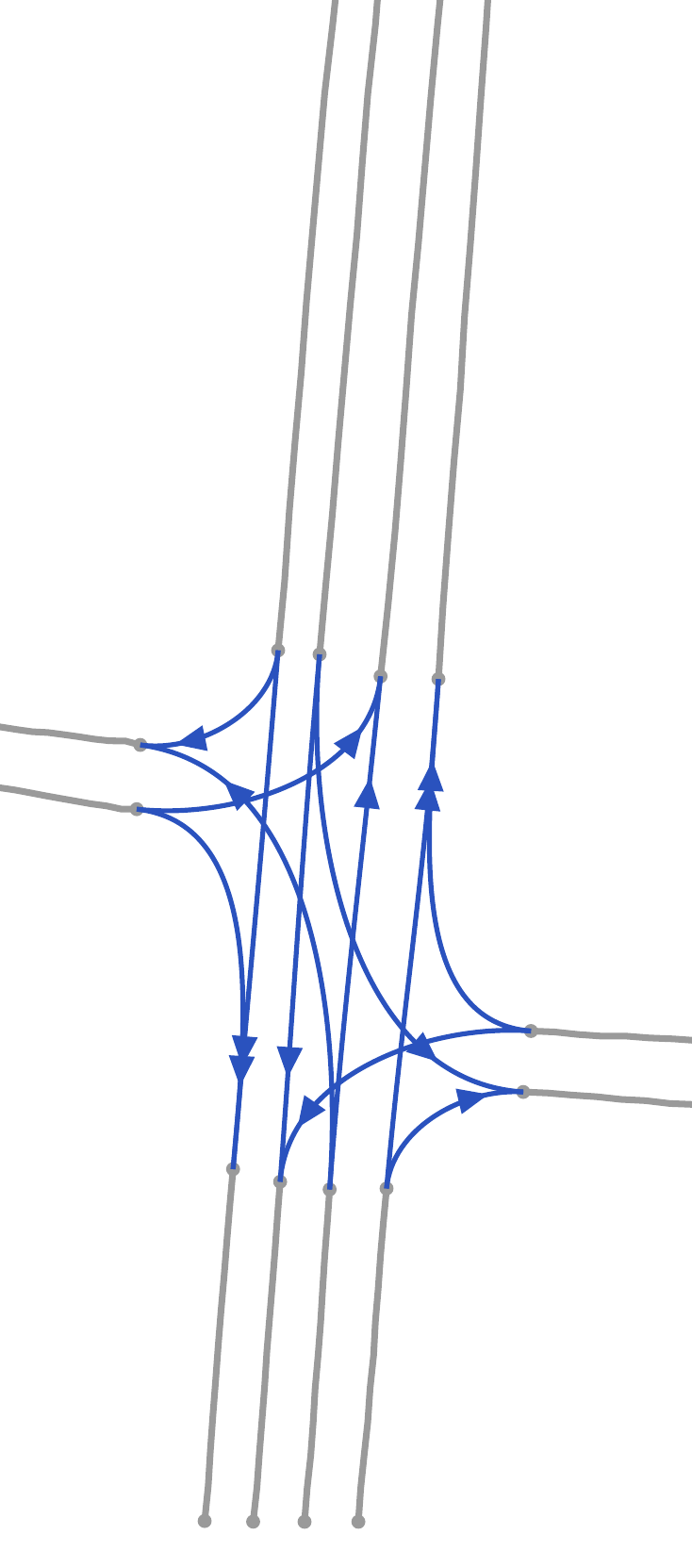}}
\put(1547.312184,885.16){\includegraphics[trim=0bp 30.164672000bp 17.600000000bp 51.848204800bp,clip,width=346.614636bp]{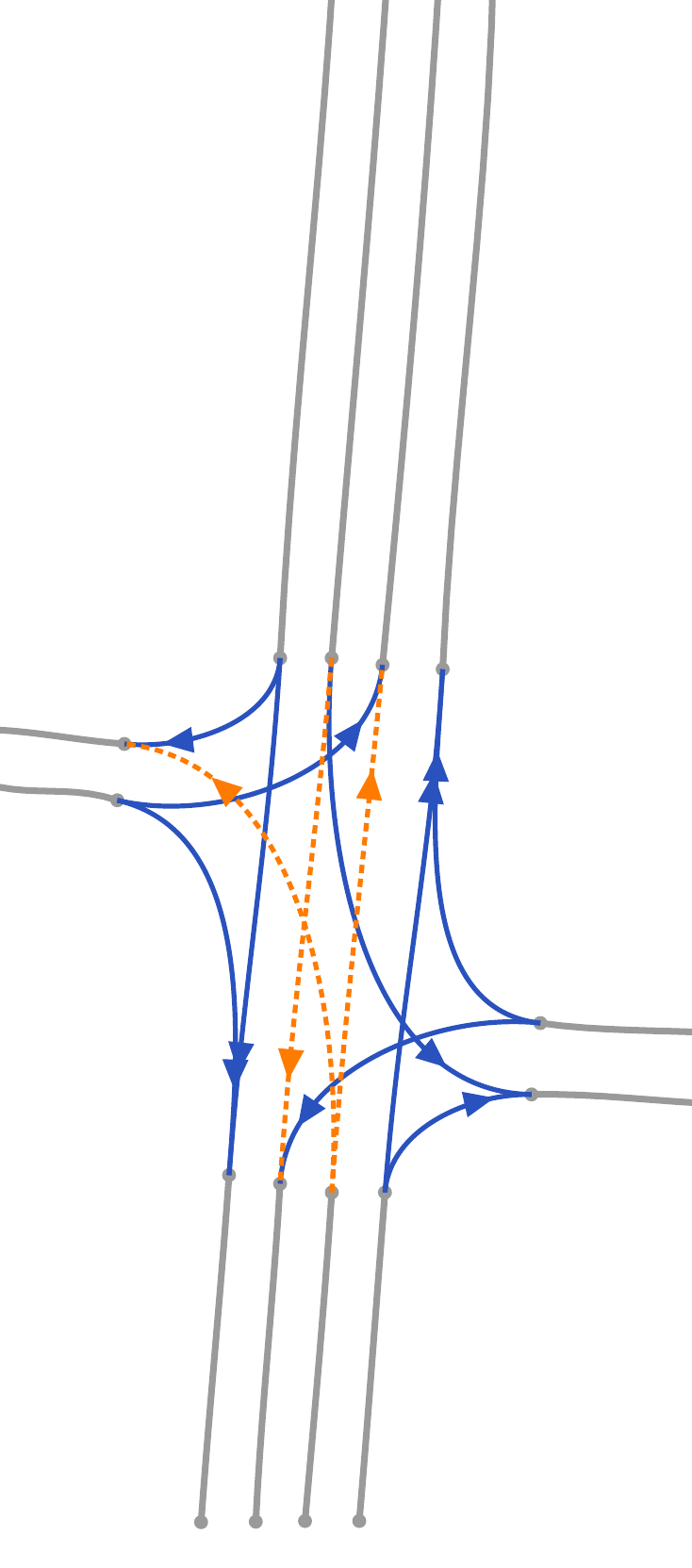}}
\put(1903.22682,885.16){\includegraphics[trim=0bp 30.164672000bp 17.600000000bp 51.848204800bp,clip,width=346.614636bp]{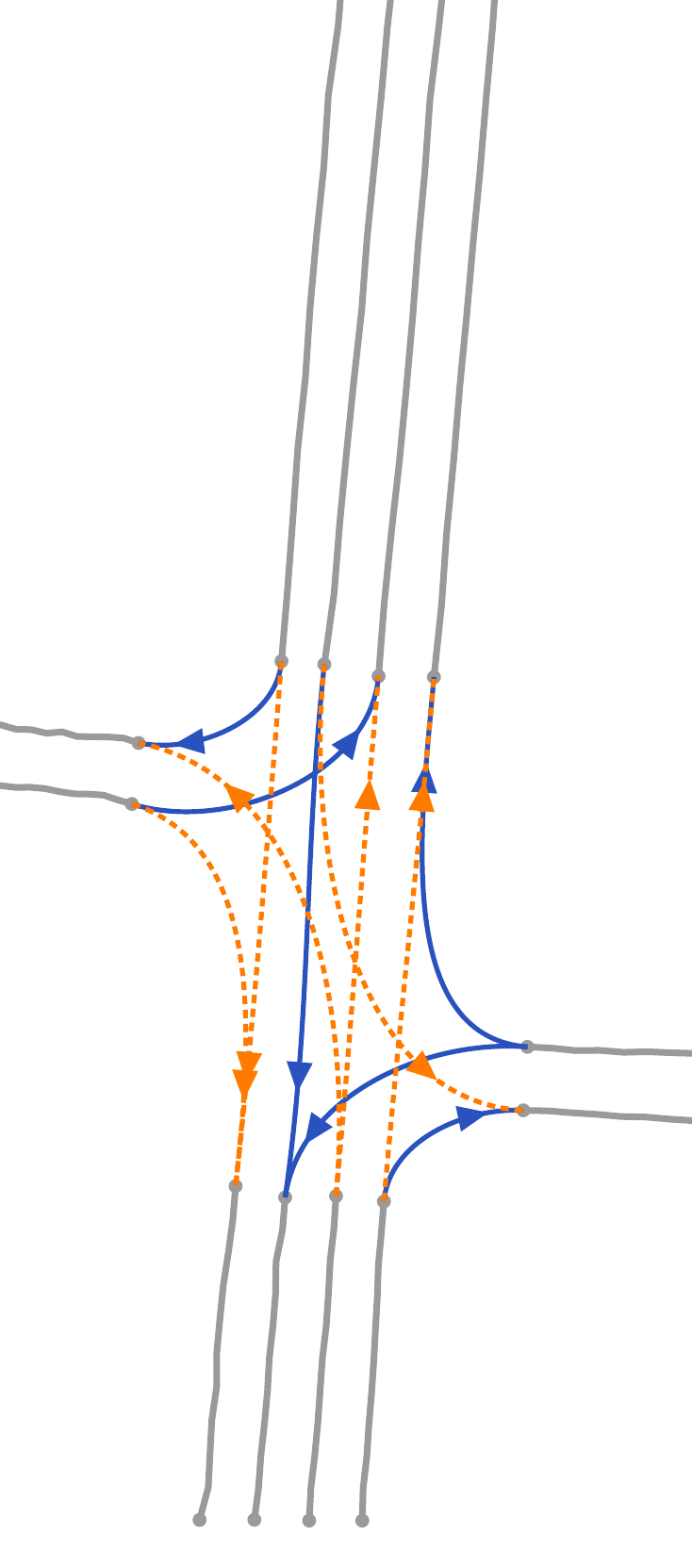}}
\put(2259.141456,885.16){\includegraphics[trim=0bp 30.164672000bp 17.600000000bp 51.848204800bp,clip,width=346.614636bp]{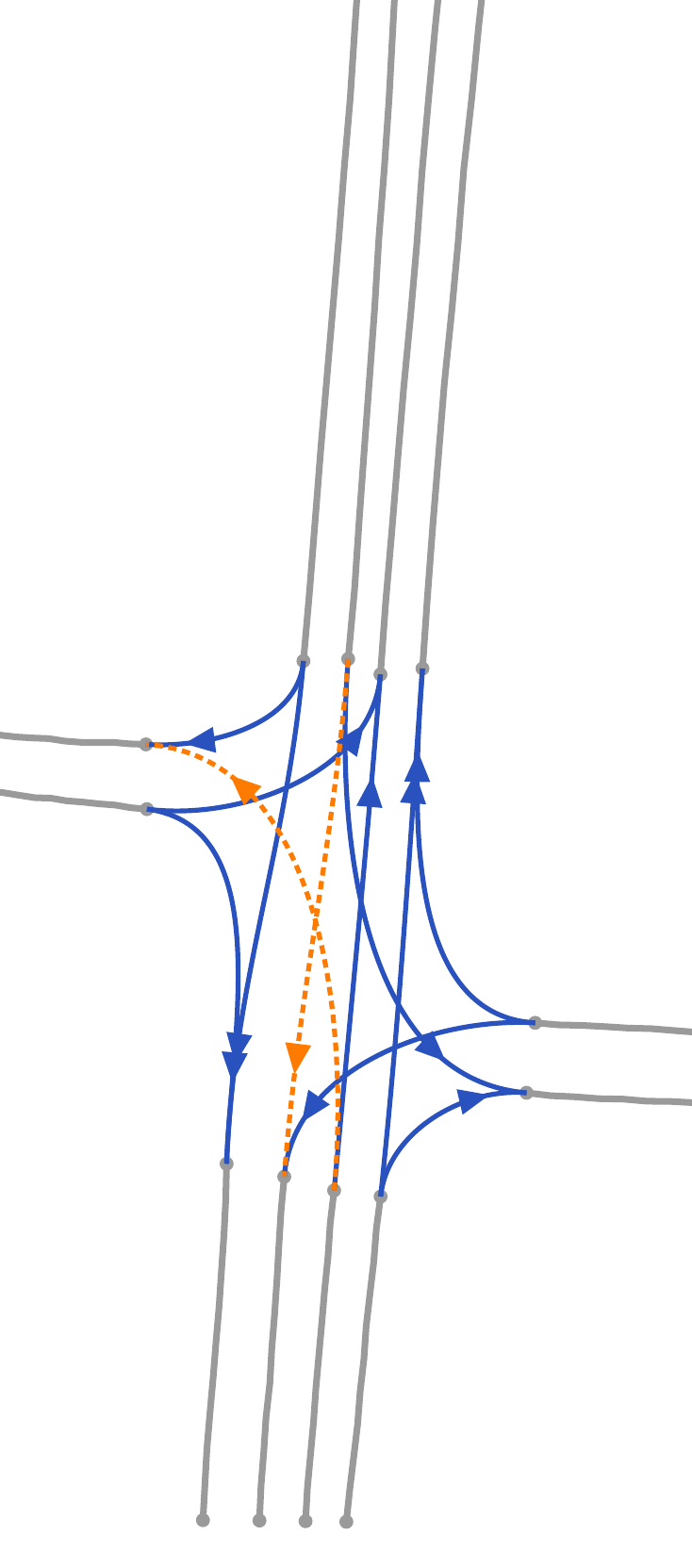}}
\put(2615.056092,885.16){\includegraphics[trim=0bp 30.164672000bp 17.600000000bp 51.848204800bp,clip,width=346.614636bp]{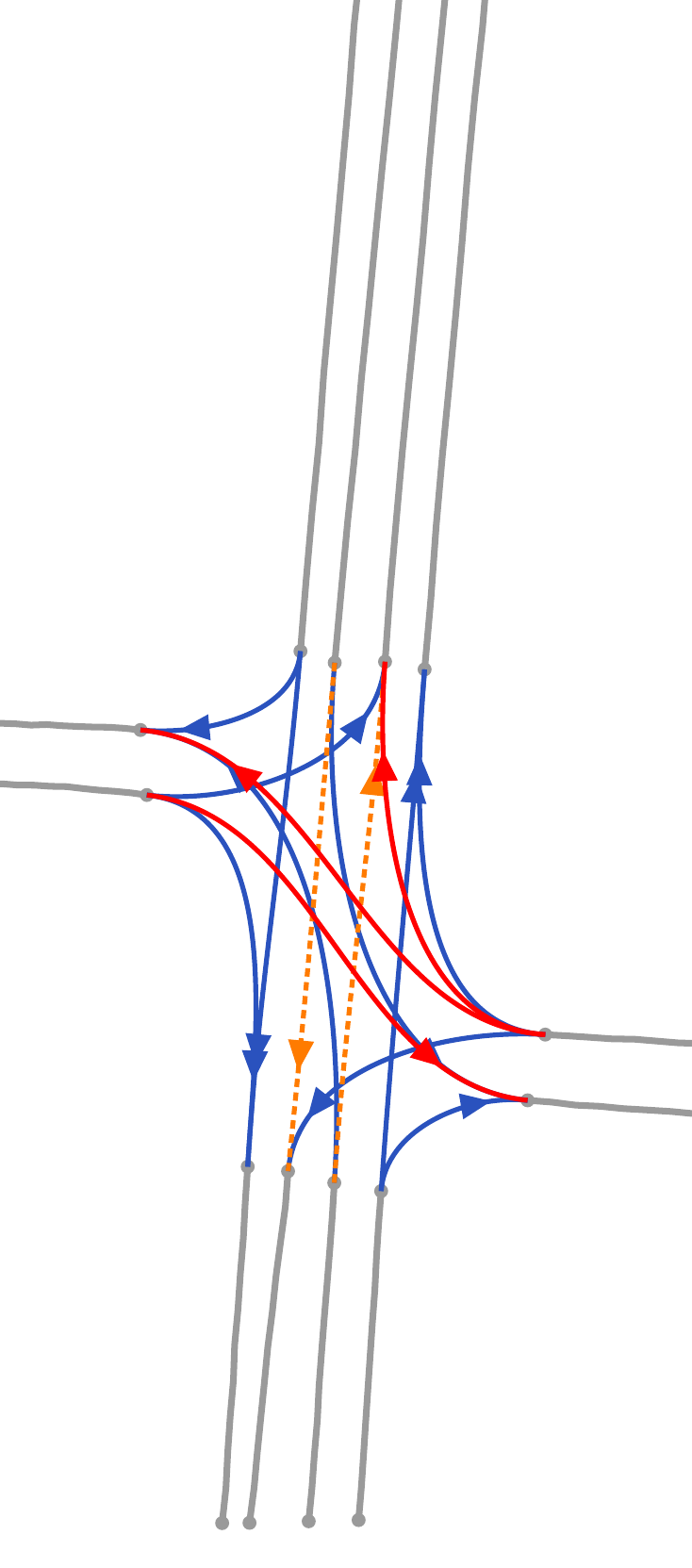}}
\put(2970.970728,885.16){\includegraphics[trim=0bp 30.164672000bp 17.600000000bp 51.848204800bp,clip,width=346.614636bp]{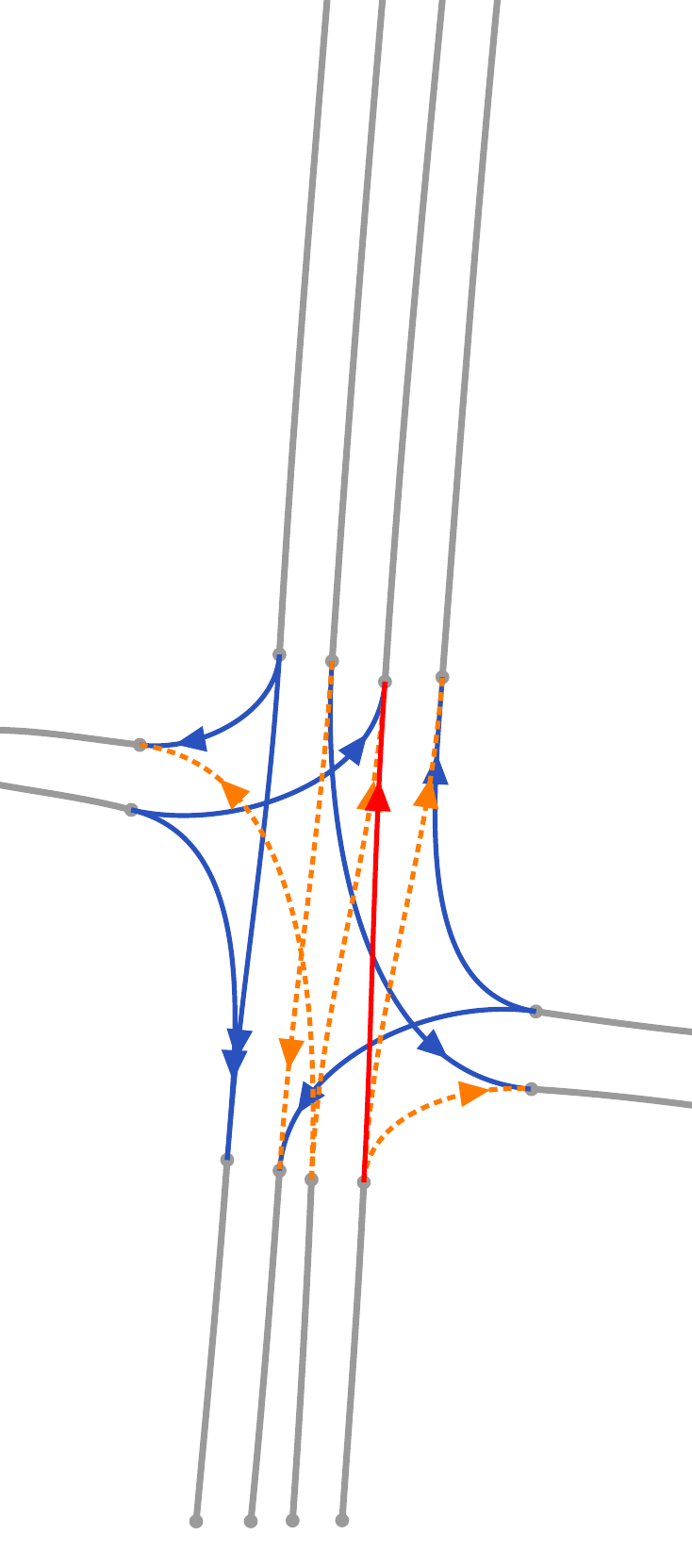}}
\put(3326.885364,885.16){\includegraphics[trim=0bp 30.164672000bp 17.600000000bp 51.848204800bp,clip,width=346.614636bp]{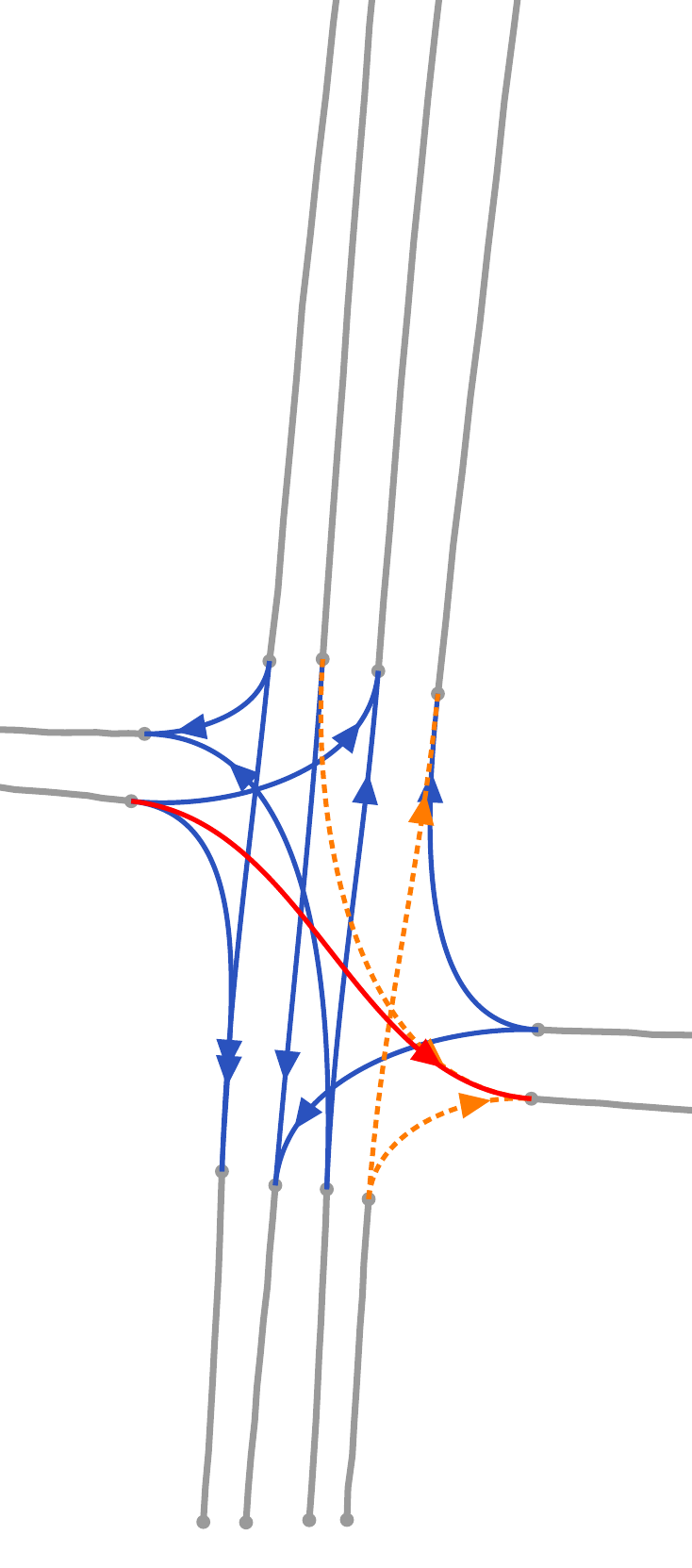}}
\put(0,1383.64){\includegraphics[trim=163.84bp 0bp 163.84bp 0bp,clip,width=270.434304bp,height=243.66bp,keepaspectratio]{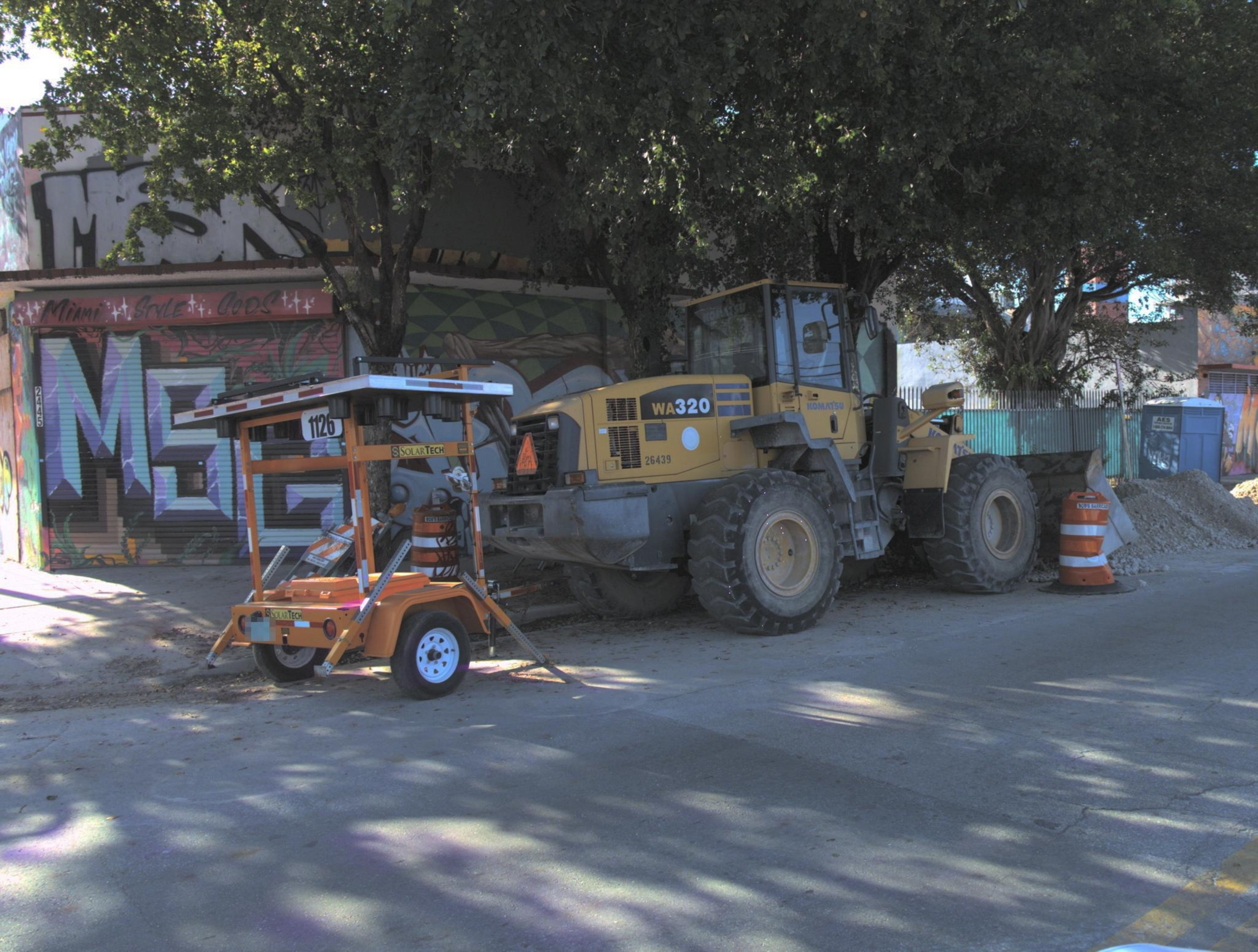}}
\put(0,1134.4){\includegraphics[trim=163.84bp 0bp 163.84bp 0bp,clip,width=270.434304bp,height=243.66bp,keepaspectratio]{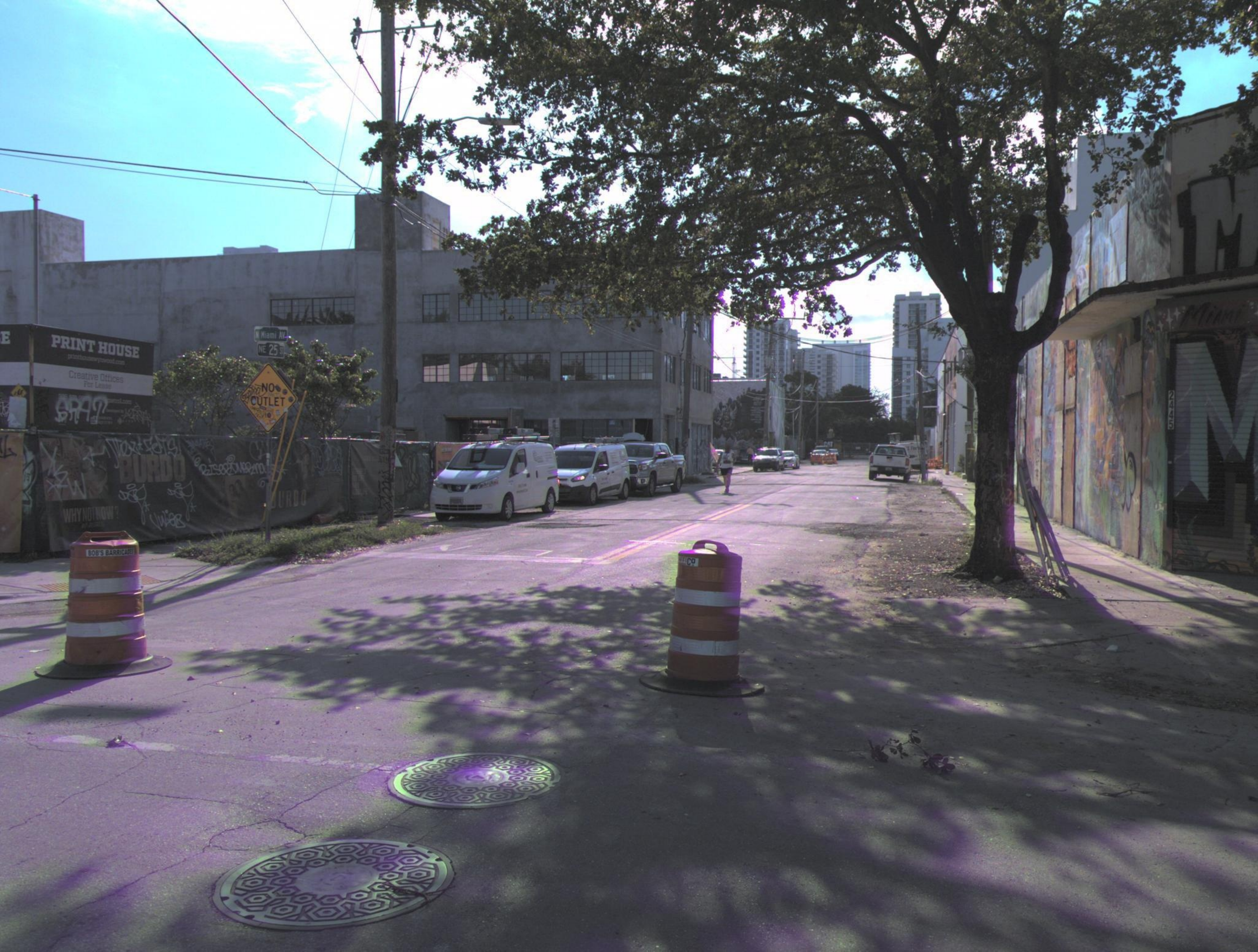}}
\put(0,885.16){\includegraphics[trim=163.84bp 0bp 163.84bp 0bp,clip,width=270.434304bp,height=243.66bp,keepaspectratio]{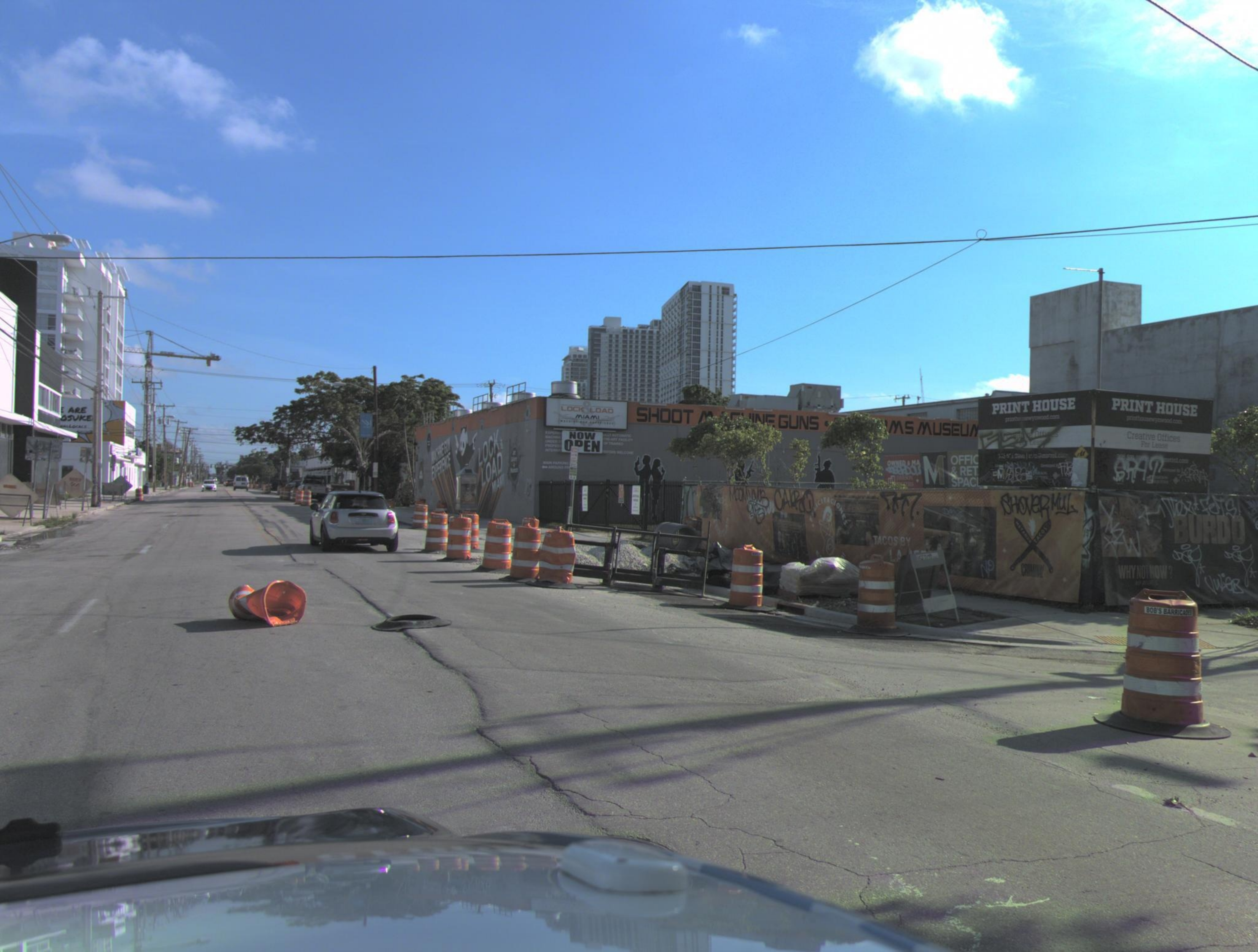}}
\put(552.028608,1383.64){\includegraphics[trim=163.84bp 0bp 163.84bp 0bp,clip,width=270.434304bp,height=243.66bp,keepaspectratio]{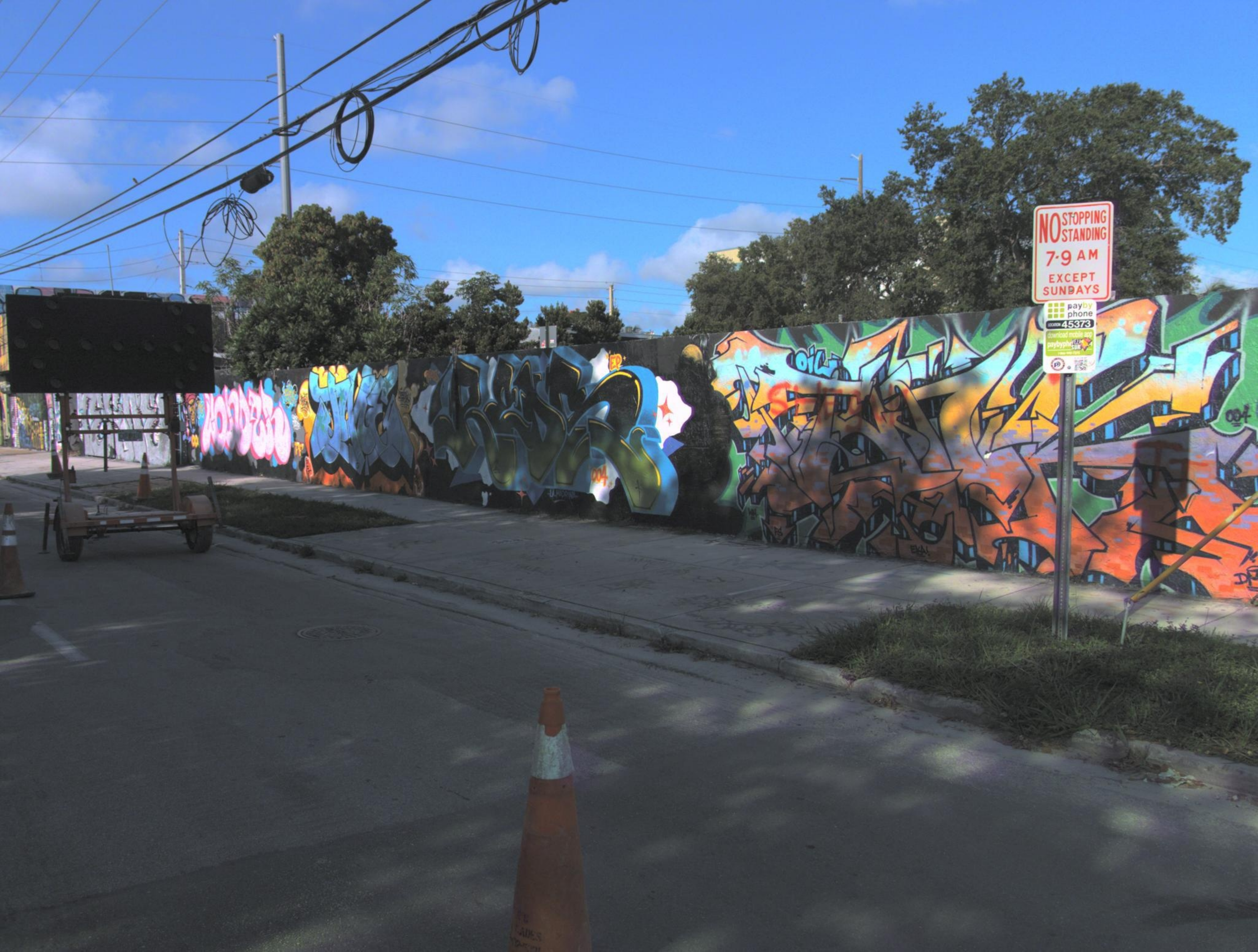}}
\put(552.028608,1134.4){\includegraphics[trim=163.84bp 0bp 163.84bp 0bp,clip,width=270.434304bp,height=243.66bp,keepaspectratio]{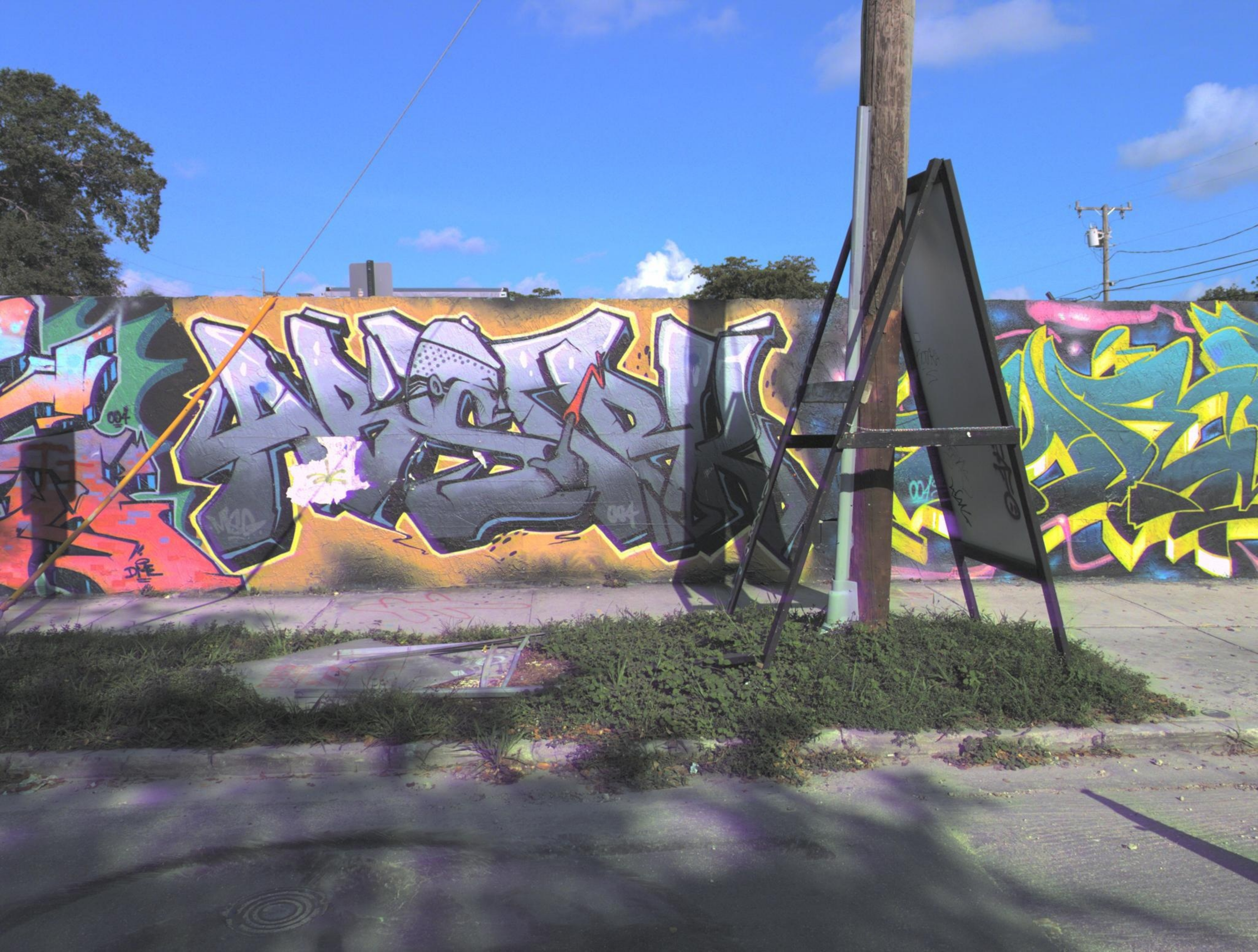}}
\put(552.028608,885.16){\includegraphics[trim=163.84bp 0bp 163.84bp 0bp,clip,width=270.434304bp,height=243.66bp,keepaspectratio]{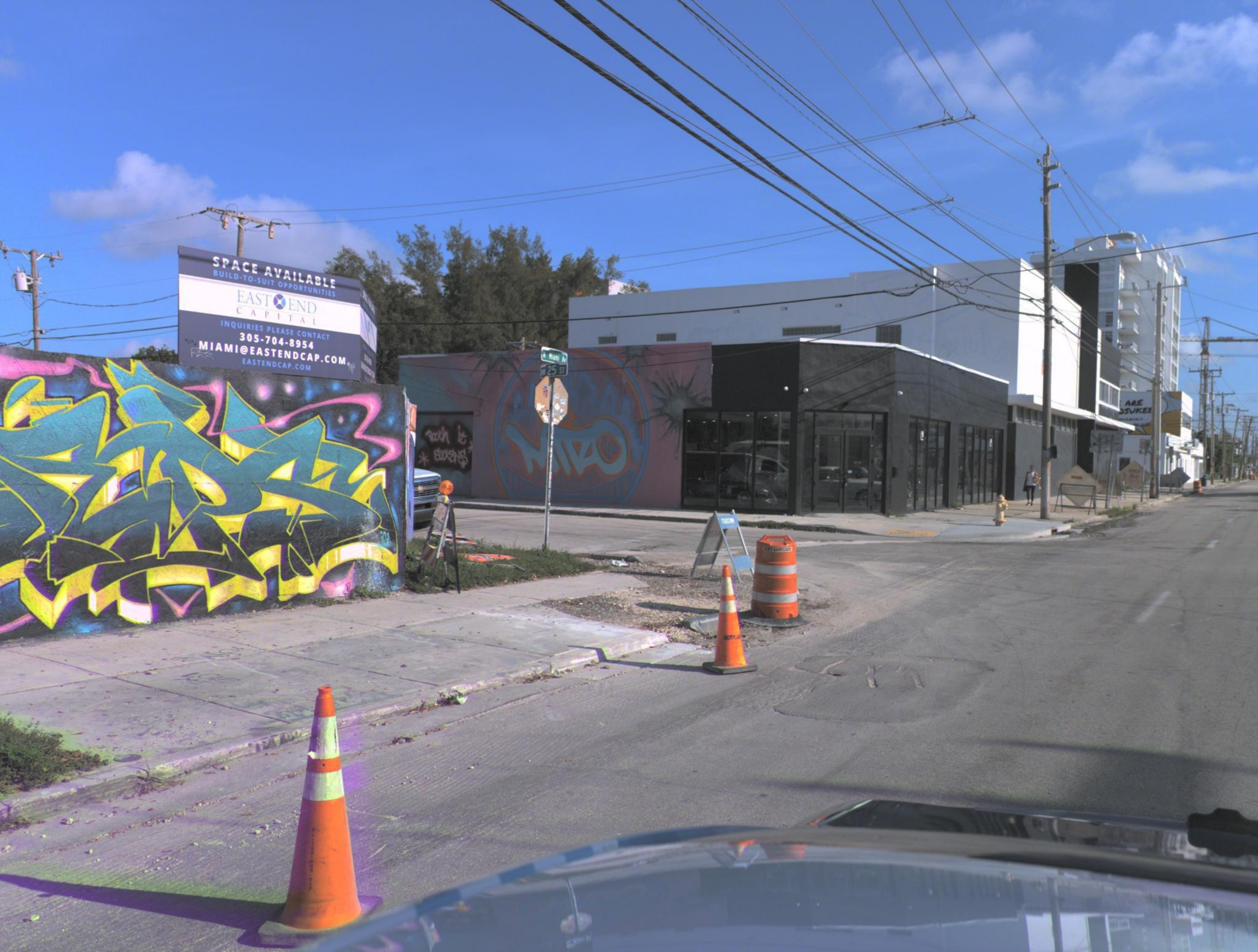}}
\put(276.014304,885.16){\includegraphics[trim=401.856486bp 0bp 401.856486bp 0bp,clip,width=270.434304bp,height=742.14bp,keepaspectratio]{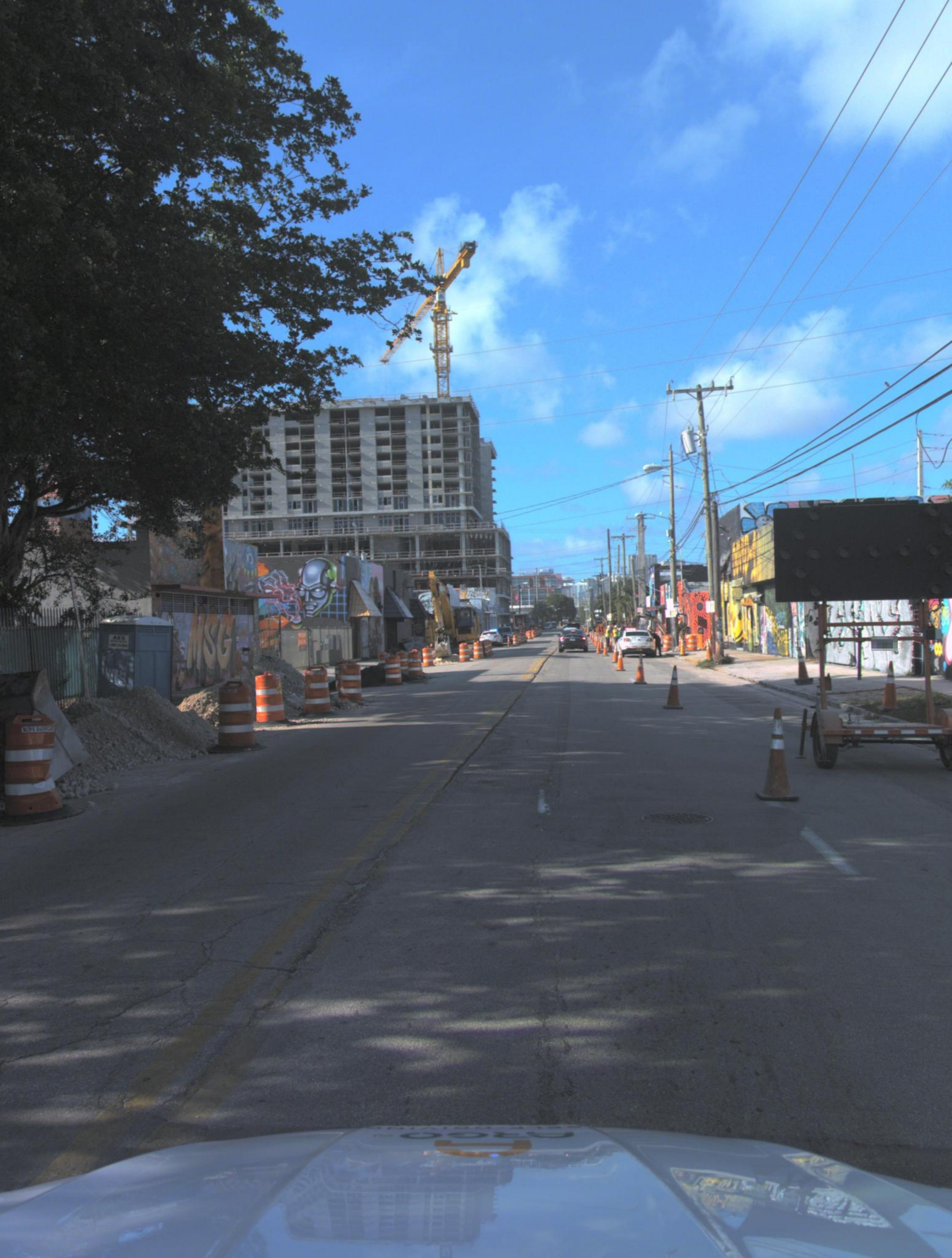}}
\put(284.014304,1619.3){\makebox(0,0)[tl]{\setlength{\fboxsep}{5bp}\colorbox{black!70}{\color{white}\fontsize{32}{32}\selectfont\sffamily\bfseries\raisebox{-0.5\height}[16bp][16bp]{F}}}}
\put(8,1120.82){\makebox(0,0)[tl]{\setlength{\fboxsep}{5bp}\colorbox{black!70}{\color{white}\fontsize{32}{32}\selectfont\sffamily\bfseries\raisebox{-0.5\height}[16bp][16bp]{RL}}}}
\put(8,1370.06){\makebox(0,0)[tl]{\setlength{\fboxsep}{5bp}\colorbox{black!70}{\color{white}\fontsize{32}{32}\selectfont\sffamily\bfseries\raisebox{-0.5\height}[16bp][16bp]{SL}}}}
\put(8,1619.3){\makebox(0,0)[tl]{\setlength{\fboxsep}{5bp}\colorbox{black!70}{\color{white}\fontsize{32}{32}\selectfont\sffamily\bfseries\raisebox{-0.5\height}[16bp][16bp]{FL}}}}
\put(560.028608,1120.82){\makebox(0,0)[tl]{\setlength{\fboxsep}{5bp}\colorbox{black!70}{\color{white}\fontsize{32}{32}\selectfont\sffamily\bfseries\raisebox{-0.5\height}[16bp][16bp]{RR}}}}
\put(560.028608,1370.06){\makebox(0,0)[tl]{\setlength{\fboxsep}{5bp}\colorbox{black!70}{\color{white}\fontsize{32}{32}\selectfont\sffamily\bfseries\raisebox{-0.5\height}[16bp][16bp]{SR}}}}
\put(560.028608,1619.3){\makebox(0,0)[tl]{\setlength{\fboxsep}{5bp}\colorbox{black!70}{\color{white}\fontsize{32}{32}\selectfont\sffamily\bfseries\raisebox{-0.5\height}[16bp][16bp]{FR}}}}
\put(835.482912,130){\includegraphics[trim=0bp 51.848204800bp 17.600000000bp 30.164672000bp,clip,width=346.614636bp]{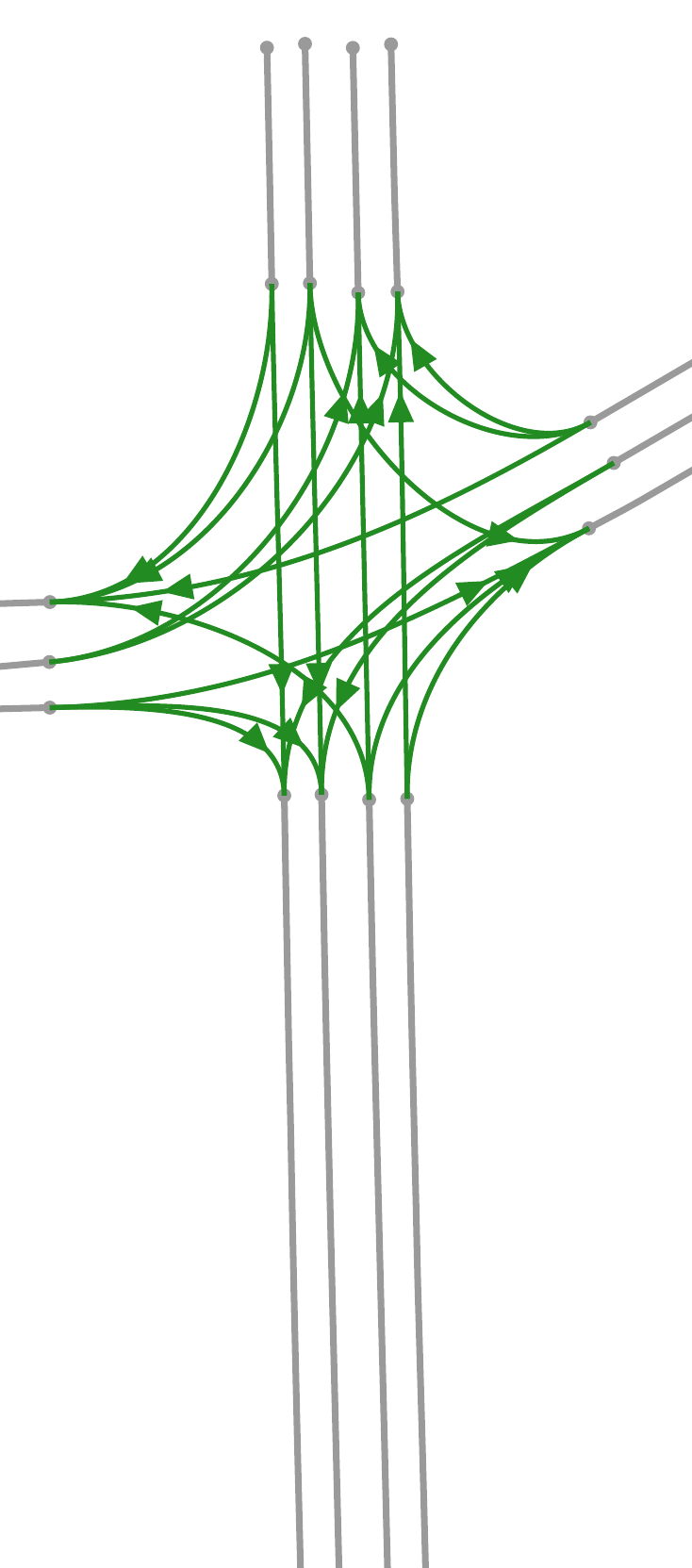}}
\put(1191.397548,130){\includegraphics[trim=0bp 51.848204800bp 17.600000000bp 30.164672000bp,clip,width=346.614636bp]{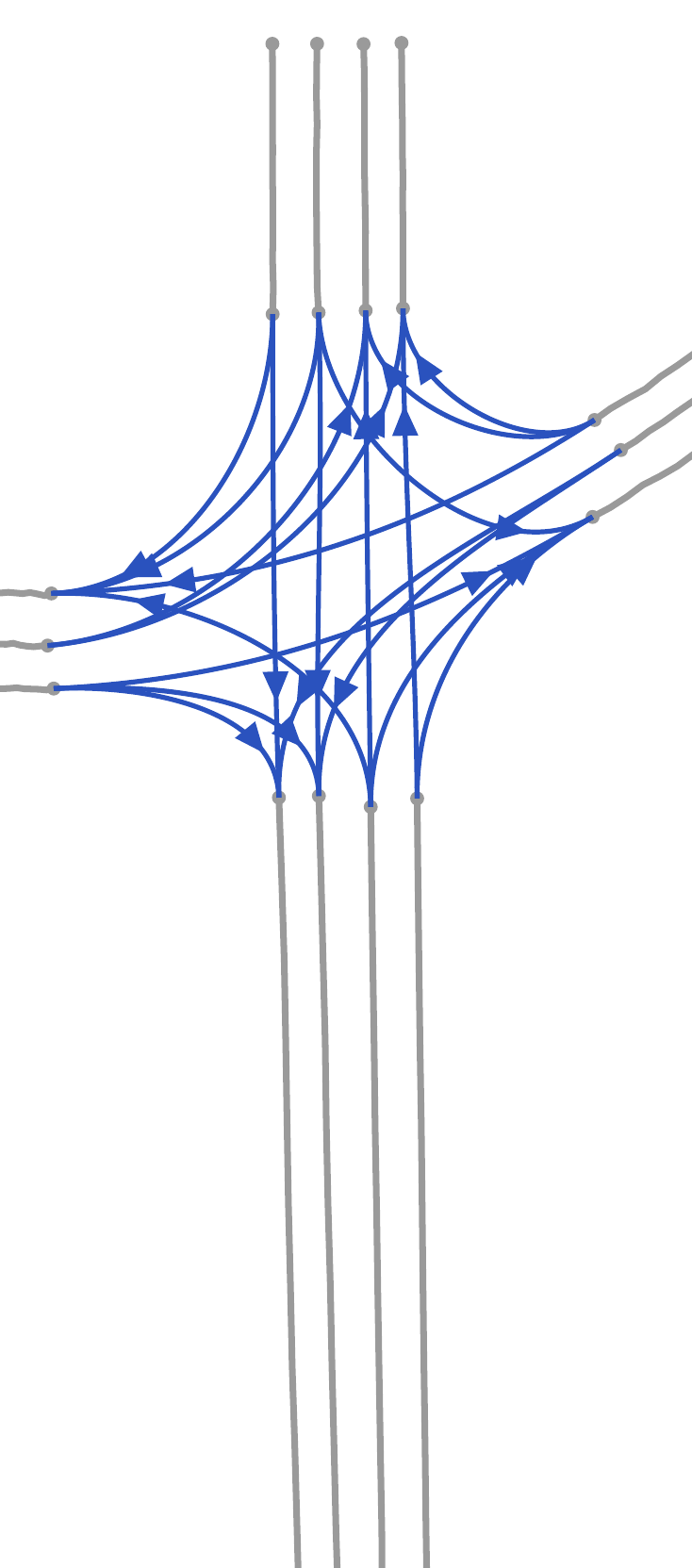}}
\put(1547.312184,130){\includegraphics[trim=0bp 51.848204800bp 17.600000000bp 30.164672000bp,clip,width=346.614636bp]{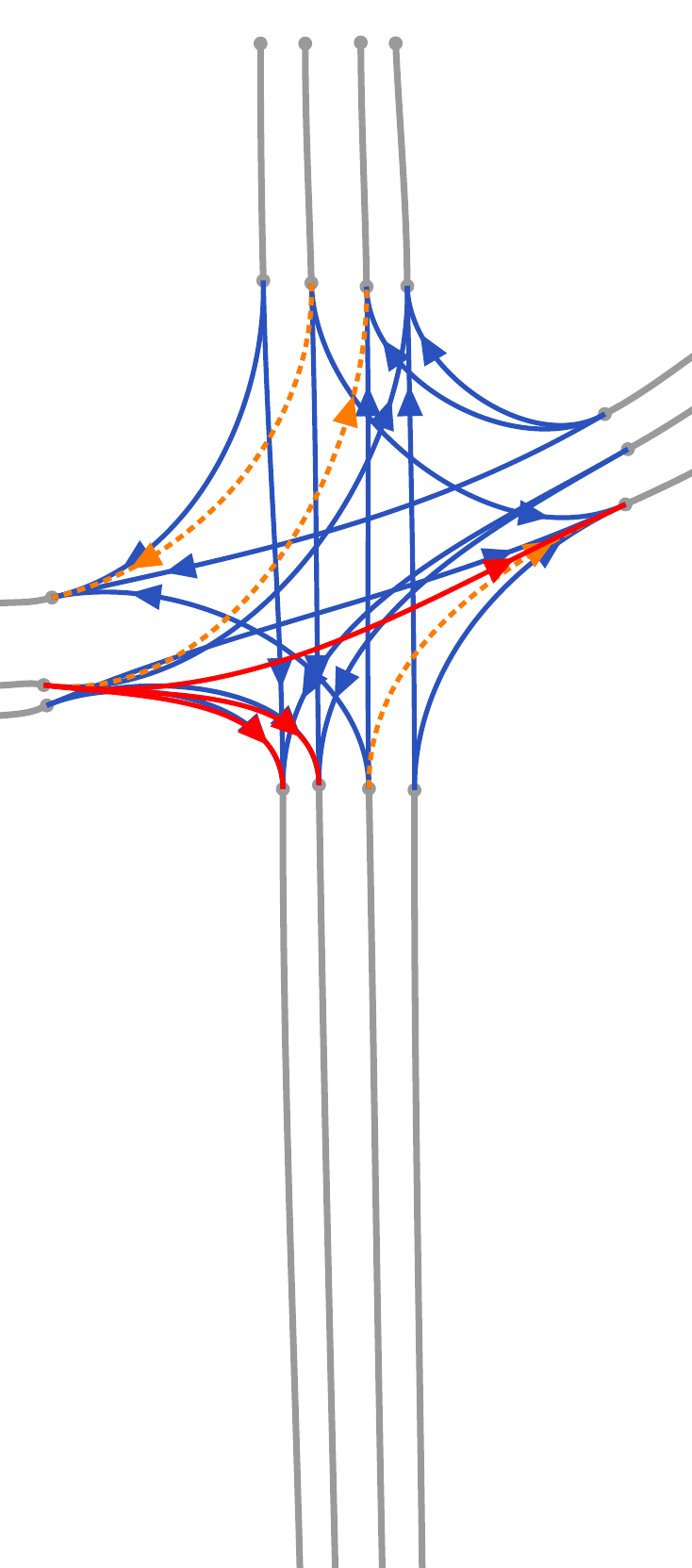}}
\put(1903.22682,130){\includegraphics[trim=0bp 51.848204800bp 17.600000000bp 30.164672000bp,clip,width=346.614636bp]{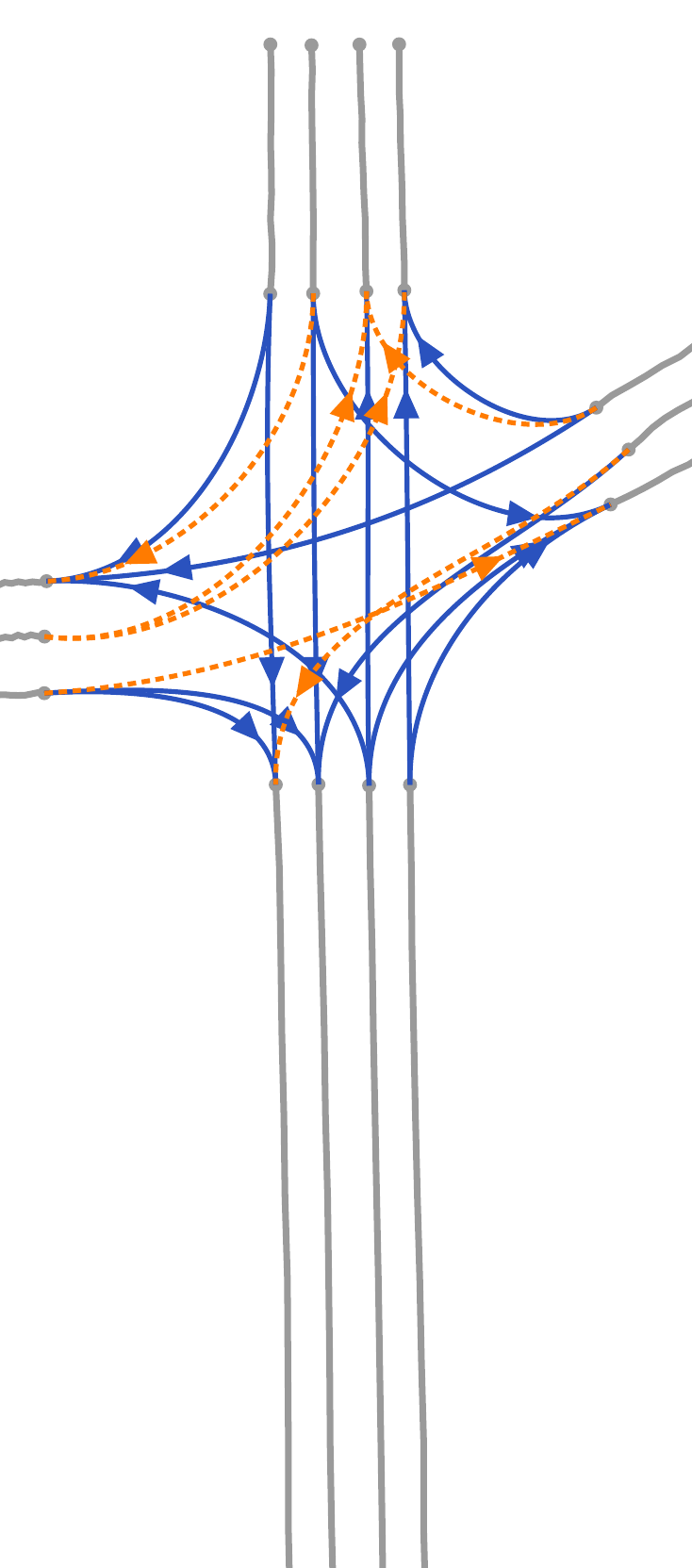}}
\put(2259.141456,130){\includegraphics[trim=0bp 51.848204800bp 17.600000000bp 30.164672000bp,clip,width=346.614636bp]{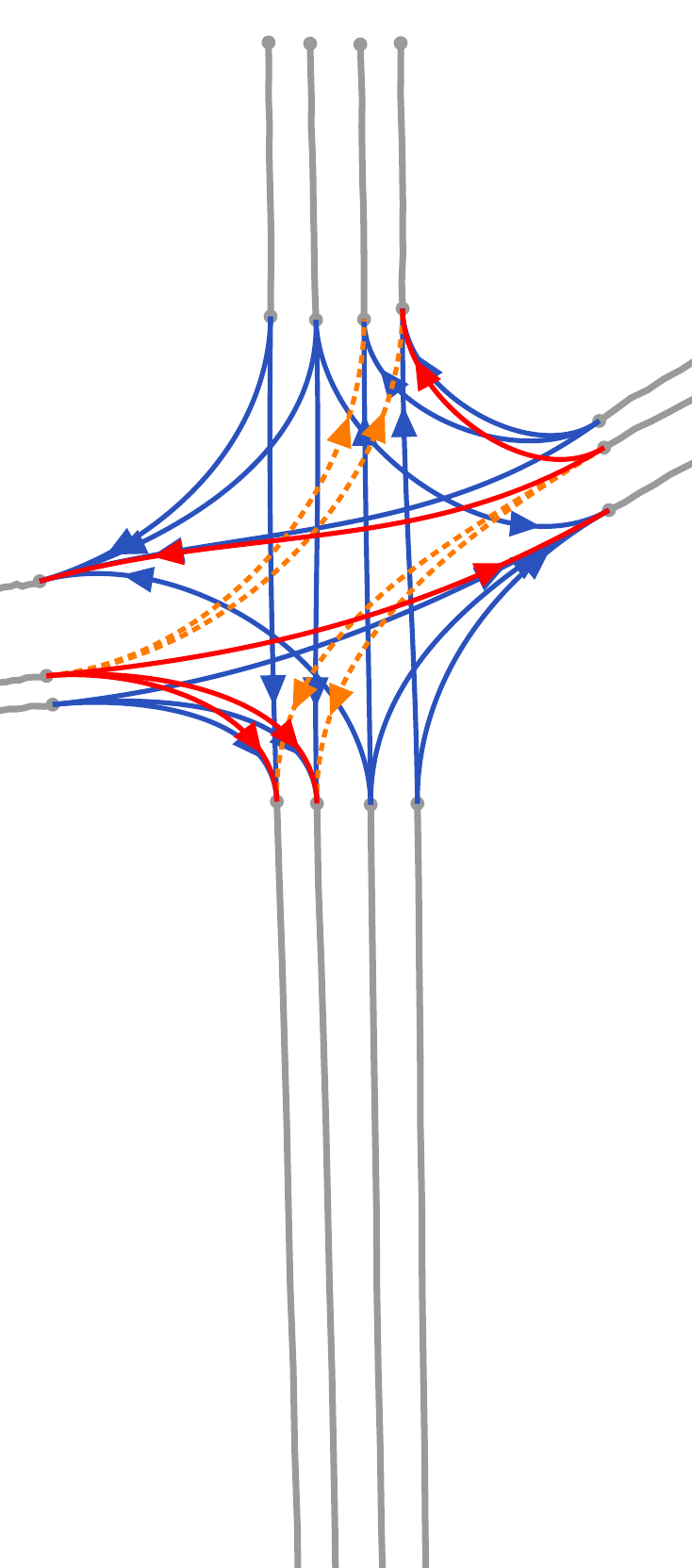}}
\put(2615.056092,130){\includegraphics[trim=0bp 51.848204800bp 17.600000000bp 30.164672000bp,clip,width=346.614636bp]{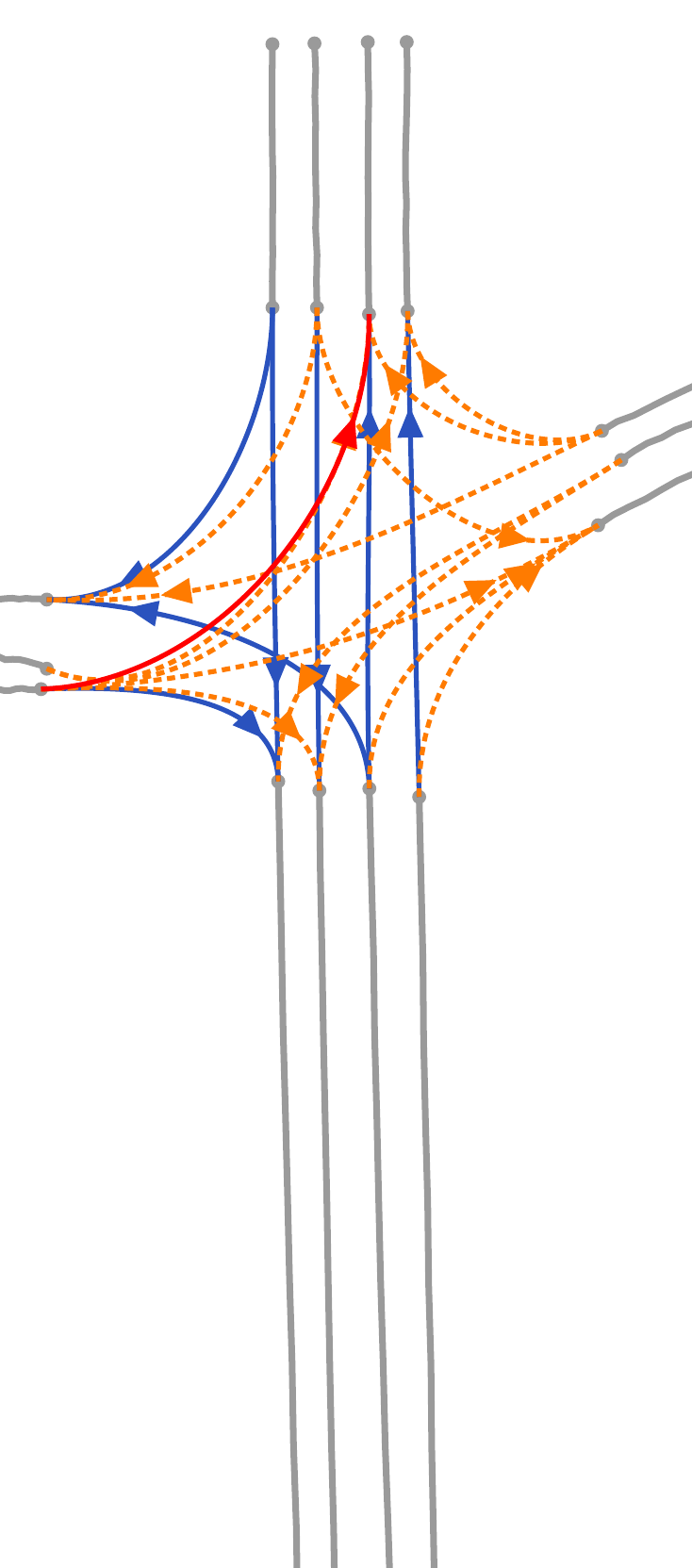}}
\put(2970.970728,130){\includegraphics[trim=0bp 51.848204800bp 17.600000000bp 30.164672000bp,clip,width=346.614636bp]{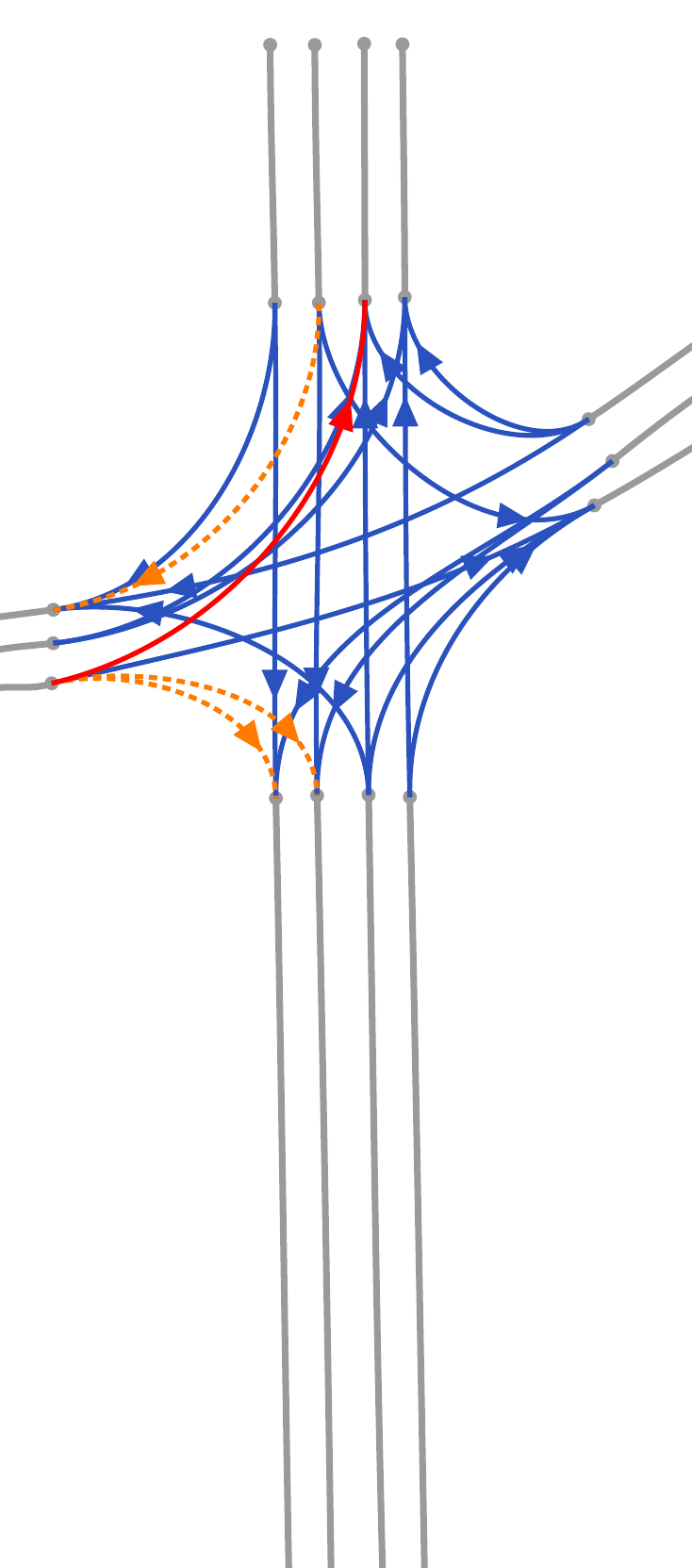}}
\put(3326.885364,130){\includegraphics[trim=0bp 51.848204800bp 17.600000000bp 30.164672000bp,clip,width=346.614636bp]{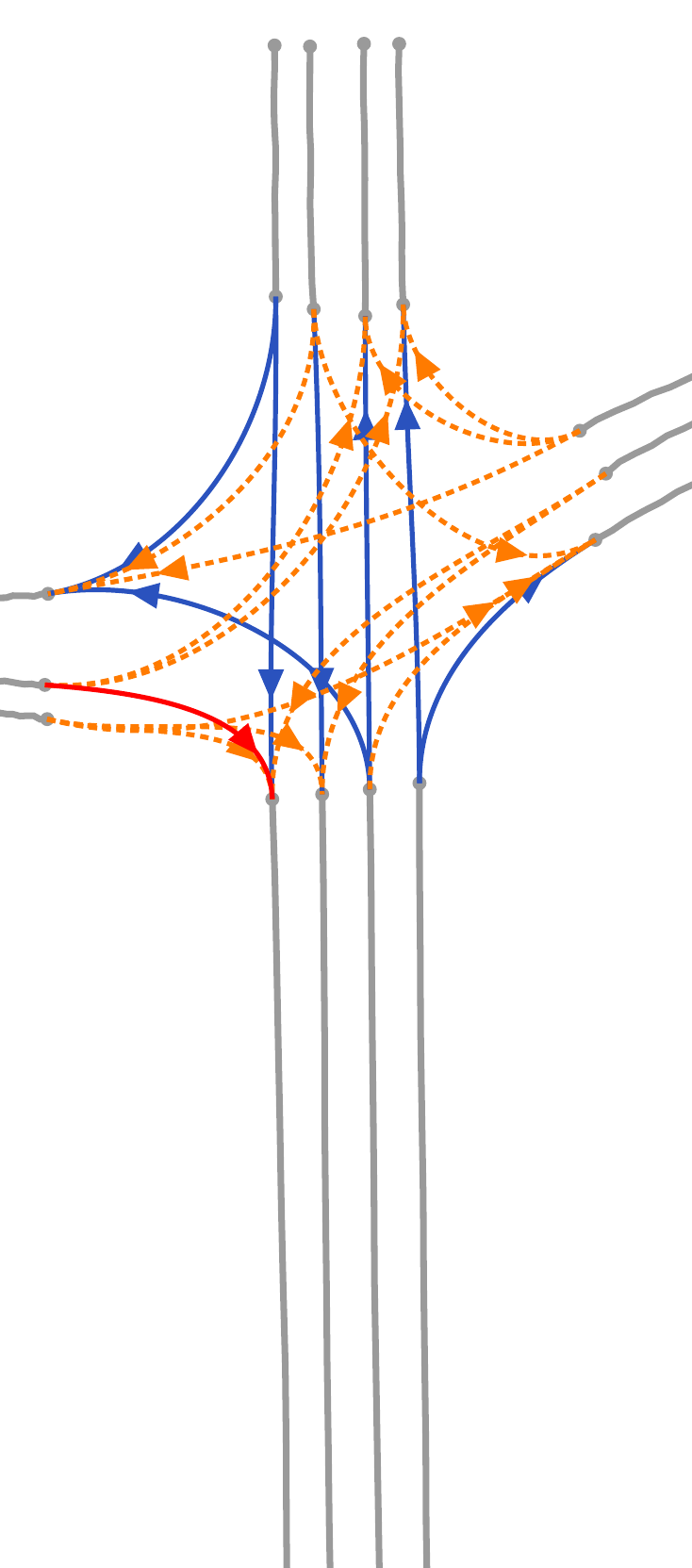}}
\put(0,628.48){\includegraphics[trim=163.84bp 0bp 163.84bp 0bp,clip,width=270.434304bp,height=243.66bp,keepaspectratio]{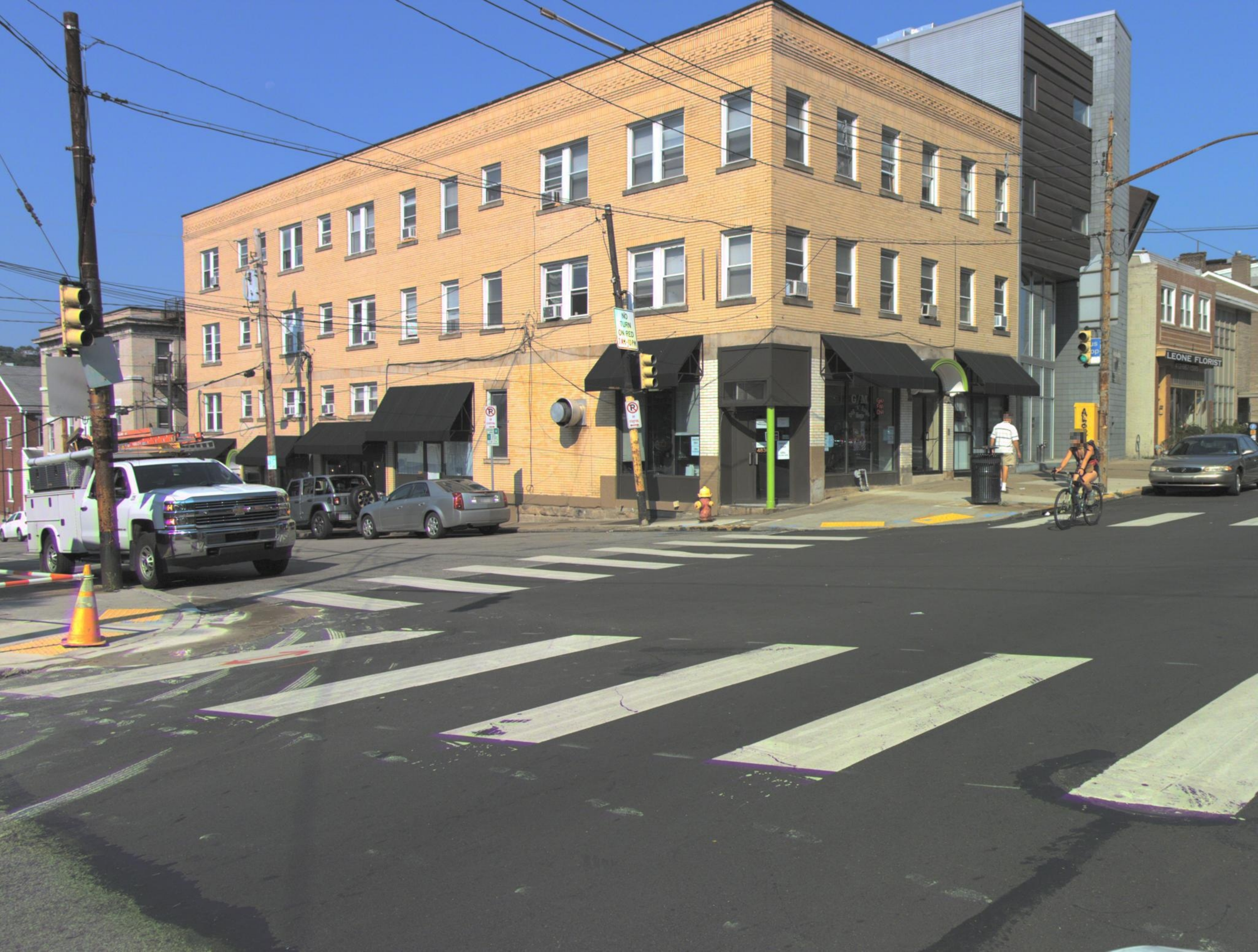}}
\put(0,379.24){\includegraphics[trim=163.84bp 0bp 163.84bp 0bp,clip,width=270.434304bp,height=243.66bp,keepaspectratio]{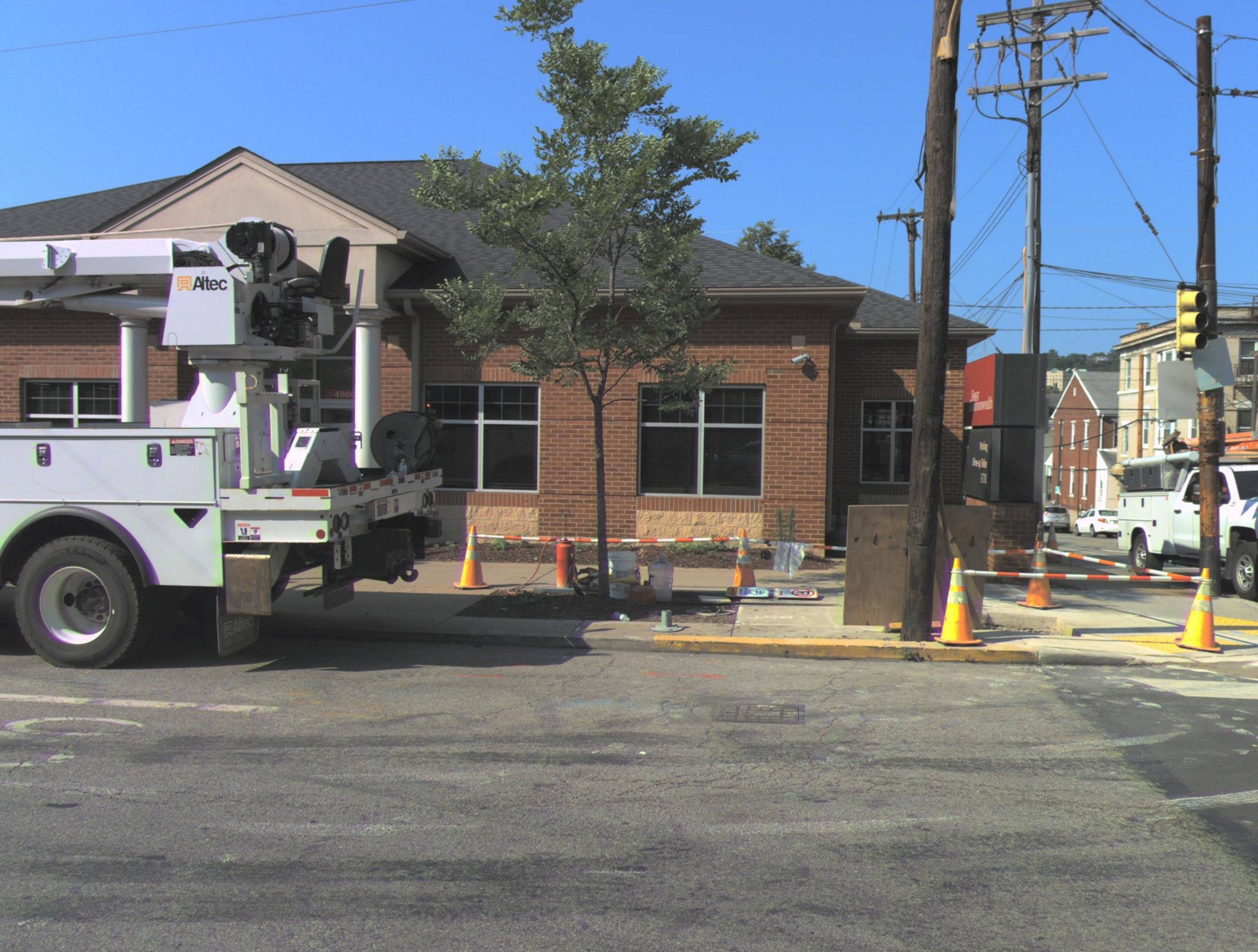}}
\put(0,130){\includegraphics[trim=163.84bp 0bp 163.84bp 0bp,clip,width=270.434304bp,height=243.66bp,keepaspectratio]{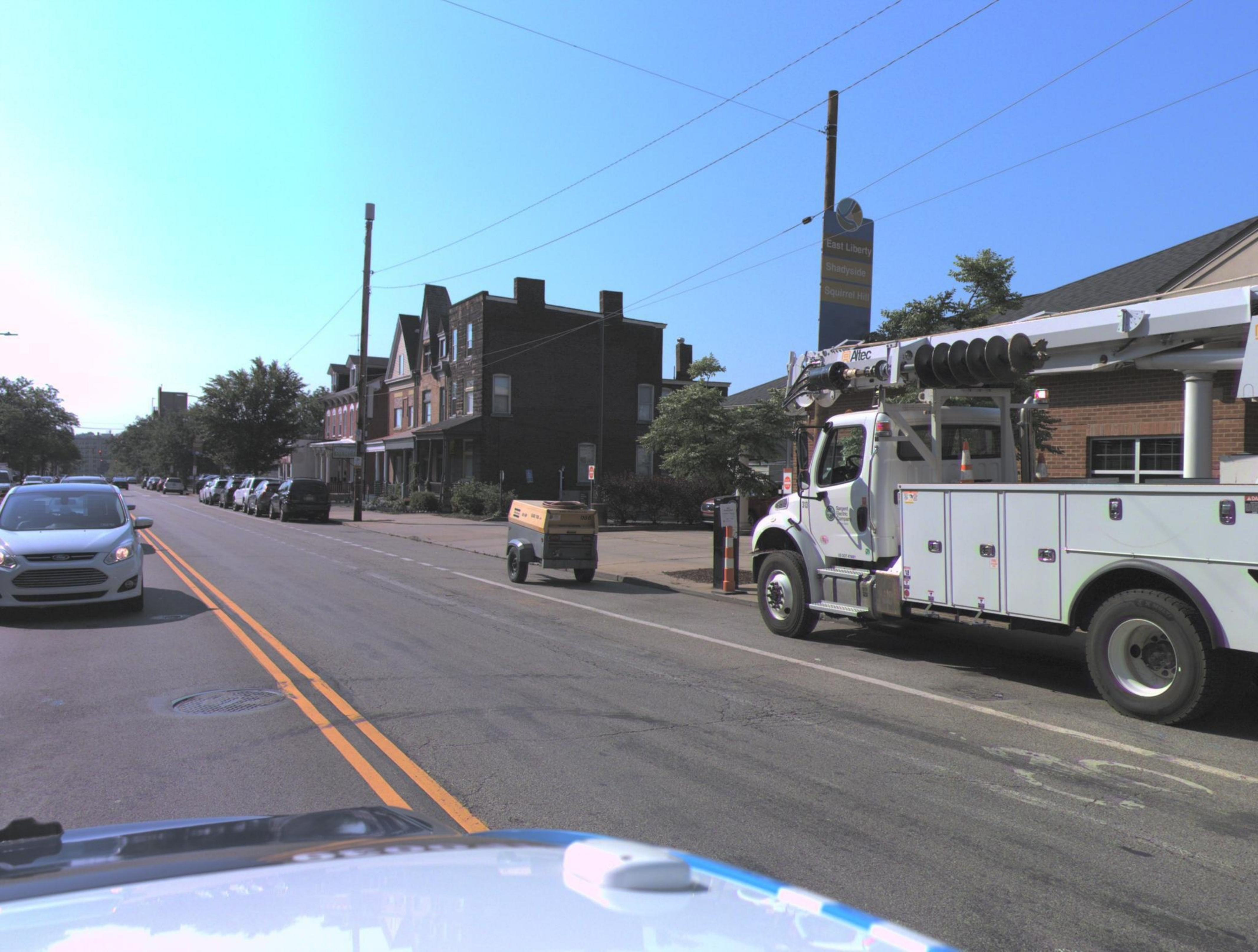}}
\put(552.028608,628.48){\includegraphics[trim=163.84bp 0bp 163.84bp 0bp,clip,width=270.434304bp,height=243.66bp,keepaspectratio]{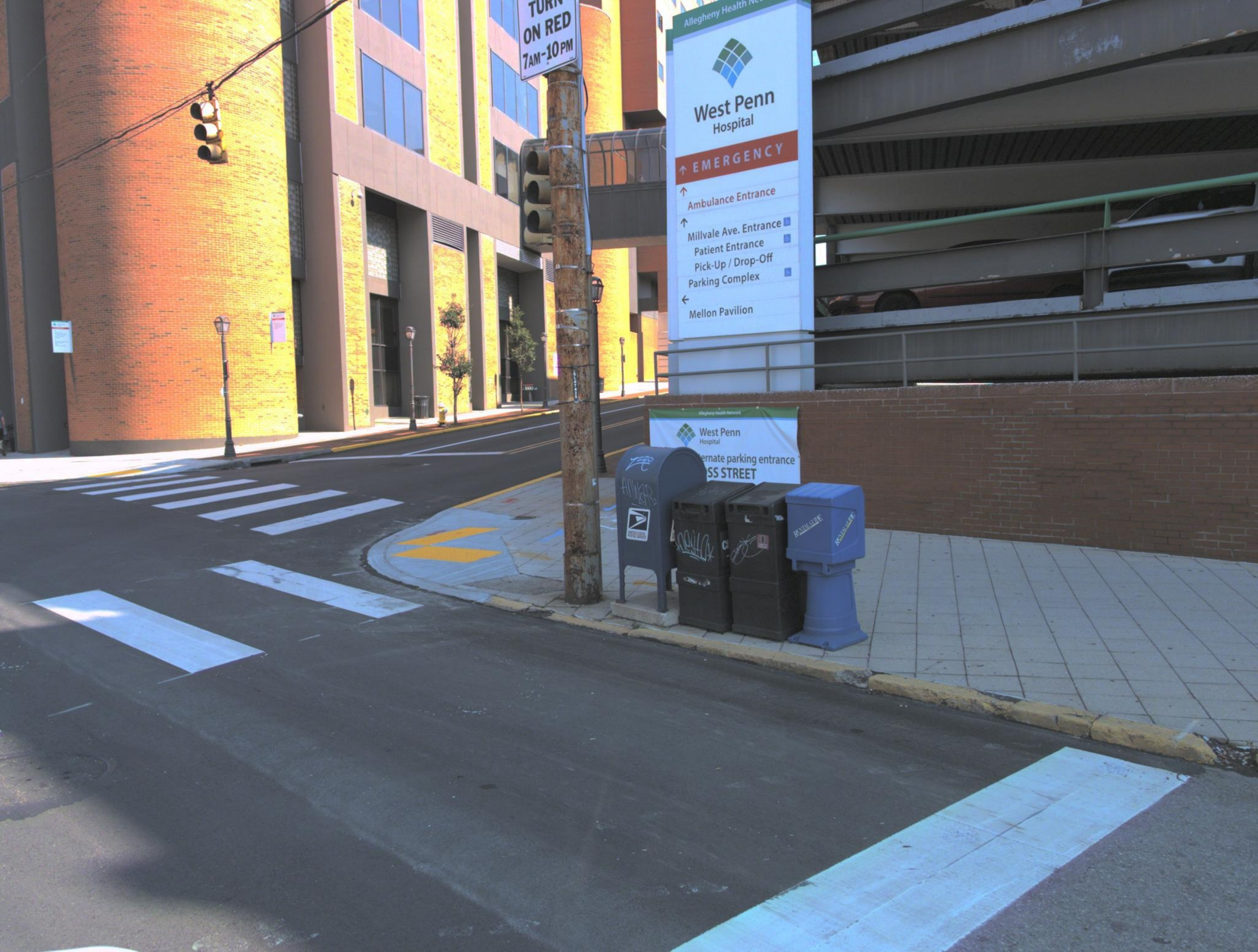}}
\put(552.028608,379.24){\includegraphics[trim=163.84bp 0bp 163.84bp 0bp,clip,width=270.434304bp,height=243.66bp,keepaspectratio]{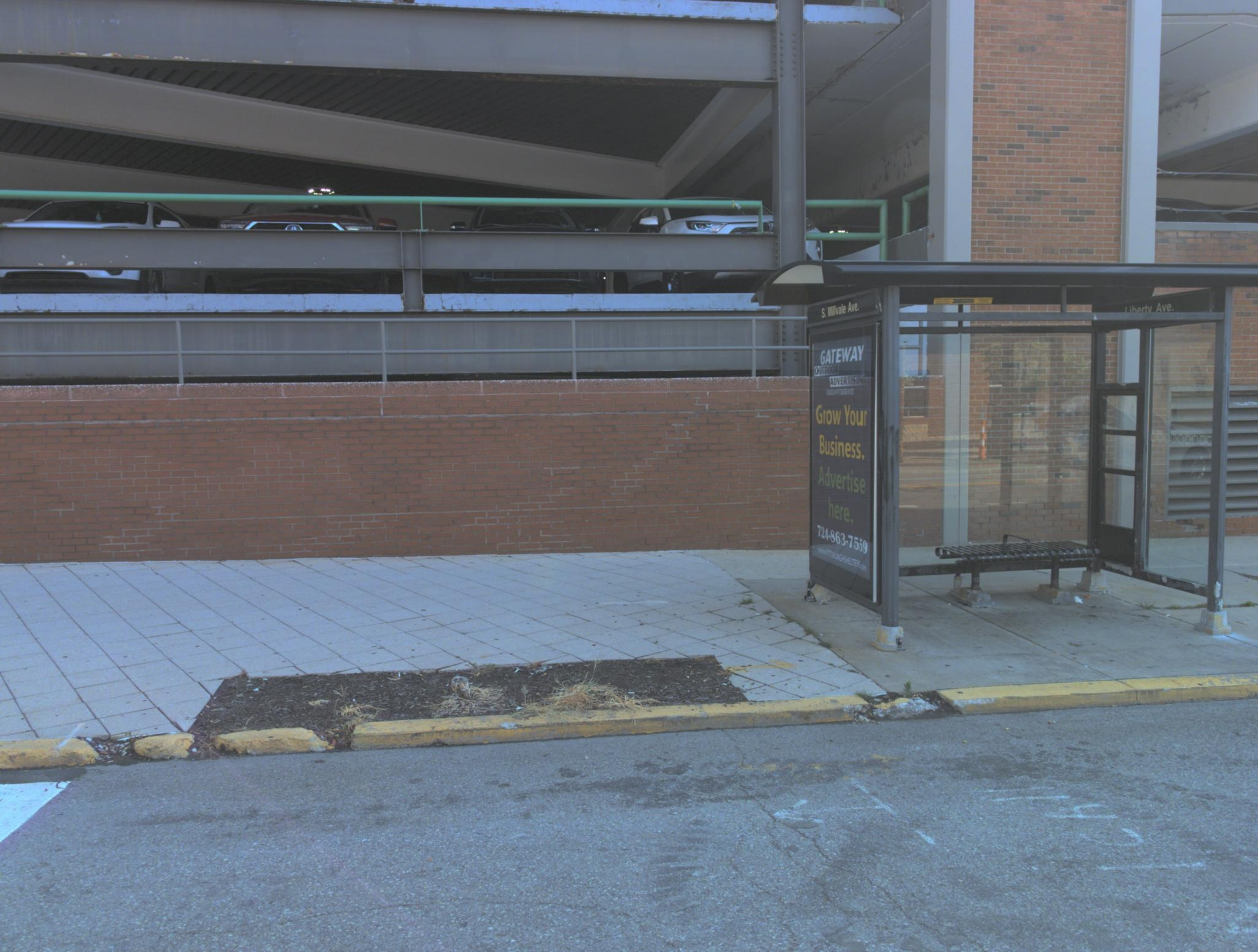}}
\put(552.028608,130){\includegraphics[trim=163.84bp 0bp 163.84bp 0bp,clip,width=270.434304bp,height=243.66bp,keepaspectratio]{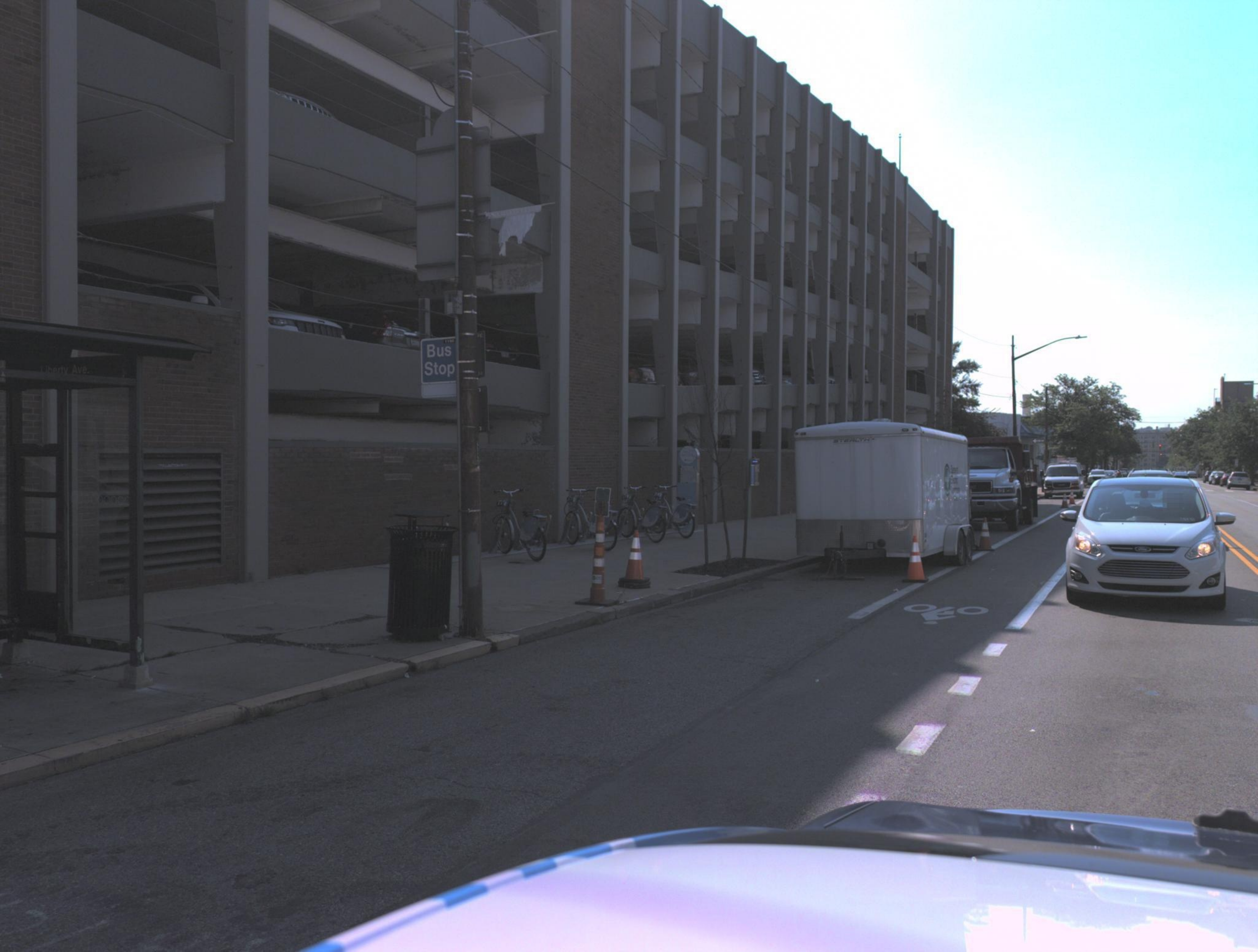}}
\put(276.014304,130){\includegraphics[trim=401.856486bp 0bp 401.856486bp 0bp,clip,width=270.434304bp,height=742.14bp,keepaspectratio]{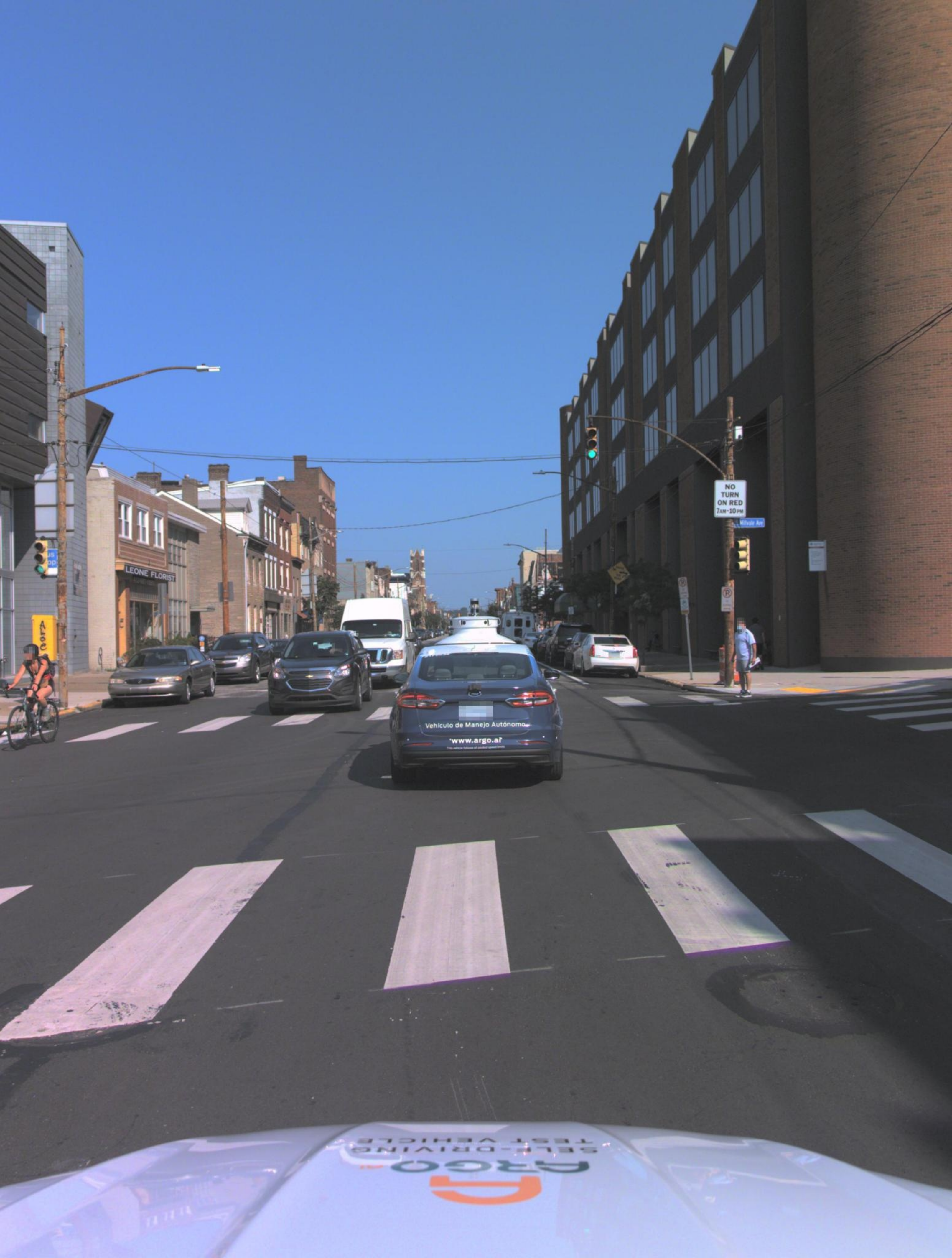}}
\put(284.014304,864.14){\makebox(0,0)[tl]{\setlength{\fboxsep}{5bp}\colorbox{black!70}{\color{white}\fontsize{32}{32}\selectfont\sffamily\bfseries\raisebox{-0.5\height}[16bp][16bp]{F}}}}
\put(8,365.66){\makebox(0,0)[tl]{\setlength{\fboxsep}{5bp}\colorbox{black!70}{\color{white}\fontsize{32}{32}\selectfont\sffamily\bfseries\raisebox{-0.5\height}[16bp][16bp]{RL}}}}
\put(8,614.9){\makebox(0,0)[tl]{\setlength{\fboxsep}{5bp}\colorbox{black!70}{\color{white}\fontsize{32}{32}\selectfont\sffamily\bfseries\raisebox{-0.5\height}[16bp][16bp]{SL}}}}
\put(8,864.14){\makebox(0,0)[tl]{\setlength{\fboxsep}{5bp}\colorbox{black!70}{\color{white}\fontsize{32}{32}\selectfont\sffamily\bfseries\raisebox{-0.5\height}[16bp][16bp]{FL}}}}
\put(560.028608,365.66){\makebox(0,0)[tl]{\setlength{\fboxsep}{5bp}\colorbox{black!70}{\color{white}\fontsize{32}{32}\selectfont\sffamily\bfseries\raisebox{-0.5\height}[16bp][16bp]{RR}}}}
\put(560.028608,614.9){\makebox(0,0)[tl]{\setlength{\fboxsep}{5bp}\colorbox{black!70}{\color{white}\fontsize{32}{32}\selectfont\sffamily\bfseries\raisebox{-0.5\height}[16bp][16bp]{SR}}}}
\put(560.028608,864.14){\makebox(0,0)[tl]{\setlength{\fboxsep}{5bp}\colorbox{black!70}{\color{white}\fontsize{32}{32}\selectfont\sffamily\bfseries\raisebox{-0.5\height}[16bp][16bp]{FR}}}}
\put(0.00,0){\includegraphics[trim=158.18bp 0bp 156.32bp 0bp,clip,width=3673.50bp,height=60bp]{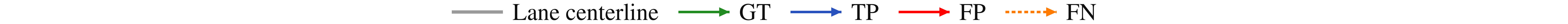}}
\put(1186.747548,885.16){\color{gray!40}\rule{.465bp}{742.14bp}}
\put(1542.662184,885.16){\color{gray!40}\rule{.465bp}{742.14bp}}
\put(1898.57682,885.16){\color{gray!40}\rule{.465bp}{742.14bp}}
\put(2254.491456,885.16){\color{gray!40}\rule{.465bp}{742.14bp}}
\put(2610.406092,885.16){\color{gray!40}\rule{.465bp}{742.14bp}}
\put(2966.320728,885.16){\color{gray!40}\rule{.465bp}{742.14bp}}
\put(3322.235364,885.16){\color{gray!40}\rule{.465bp}{742.14bp}}
\put(1186.747548,130){\color{gray!40}\rule{.465bp}{742.14bp}}
\put(1542.662184,130){\color{gray!40}\rule{.465bp}{742.14bp}}
\put(1898.57682,130){\color{gray!40}\rule{.465bp}{742.14bp}}
\put(2254.491456,130){\color{gray!40}\rule{.465bp}{742.14bp}}
\put(2610.406092,130){\color{gray!40}\rule{.465bp}{742.14bp}}
\put(2966.320728,130){\color{gray!40}\rule{.465bp}{742.14bp}}
\put(3322.235364,130){\color{gray!40}\rule{.465bp}{742.14bp}}
\end{picture}}
\caption{\textbf{Qualitative comparison of lane detection and lane--lane topology on OLR.} Each row shows multi-view images and BEV results for an intersection scene.}
\label{fig:qual}
\end{figure*}

%% file: tables/ablation.tex
\begin{table}[t]
  \centering
  \small
  \setlength{\tabcolsep}{3pt}
  \begin{adjustbox}{max width=\linewidth}
    \begin{tabular}{l|c|cc|gc}
      \toprule
      Method & \rols & DET$_l$ & DET$_t$ & \rtop{} & TOP$_{lt}$ \\
      \midrule
      Base detector & 43.5 & 29.9 & 53.7 & 12.7 & 30.1 \\
      \quad + End refinement & 45.0 & 33.2 & 53.4 & 13.9 & 31.7 \\
      \quad + Pair head & 49.0 & 32.9 & 53.8 & 28.0 & 32.0 \\
      \quad + Assignment & 50.6 & 33.8 & 54.1 & 32.6 & 33.1 \\
      \quad + Approach TE & 51.4 & 33.7 & 53.6 & 32.7 & 37.3 \\
      \bottomrule
    \end{tabular}
  \end{adjustbox}
  \caption{Cumulative ablation on OLR \textit{subset\_A} val. Each row adds one component to the row above. The shaded column is \rtop{}.}
  \label{tab:ablation}
\end{table}

%% file: sec/6_conclusion.tex
\section{Conclusion}
\label{sec:concl}
We uncovered a shortcut in OpenLane-V2: its proximity-based connection labels let a simple endpoint rule perform on par with learned topology heads on the connections at intersections, since connected lanes share endpoints.
We built OpenLane-V2 Reachability (OLR) to close this shortcut by replacing the centerlines inside intersections with direct connections, so that \rtop{} measures which exits an entry lane can reach.
On OLR, the endpoint rule recovers almost none of the connections across intersections, and a higher TOP$_{ll}$ on OpenLane-V2 does not carry over to \rtop{}.
Our proposed \ours, designed around reachability, ranks first in every metric on both subsets of OLR.
To evaluate the reachability decision free of lane detection, we give all methods the same ground-truth lanes.
Under this setting, \ours still predicts connections with the highest accuracy.

\paragraph{Limitations.}
\rtop{} is still bounded by lane detection, as a connection counts only between lanes matched to the ground truth.
OLR also leaves the paths through intersections unscored, which future work can predict once it knows where a lane leads.

%% file: sec/X_suppl.tex
\clearpage
\setcounter{page}{1}
\maketitlesupplementary
\setcounter{section}{0}\renewcommand{\thesection}{S\arabic{section}}\renewcommand{\theHsection}{S\arabic{section}}
\setcounter{table}{0}\renewcommand{\thetable}{S\arabic{table}}\renewcommand{\theHtable}{S\arabic{table}}
\setcounter{figure}{0}\renewcommand{\thefigure}{S\arabic{figure}}\renewcommand{\theHfigure}{S\arabic{figure}}
\setcounter{equation}{0}\renewcommand{\theequation}{S\arabic{equation}}\renewcommand{\theHequation}{S\arabic{equation}}
\newcommand{\supptodo}[1]{\textcolor{red}{[TODO: #1]}}

\suppressfloats[t]
\section*{Overview}
\begin{center}
  \small
  \setlength{\tabcolsep}{5pt}
  \begin{tabular}{@{}llc@{}}
    \toprule
    & Content & Main text \\
    \midrule
    \ref{sec:supp_rule} & Learned heads vs.\ endpoint rule & \S\ref{sec:olv2_limits} \\
    \ref{sec:supp_rules} & Constructing OLR & \S\ref{sec:olv2r} \\
    \ref{sec:supp_metrics} & Evaluation metrics & \S\ref{sec:setup} \\
    \ref{sec:supp_impl} & Implementation details & \S\ref{sec:method} \\
    \ref{sec:supp_olv2} & Results on OpenLane-V2 & \S\ref{sec:main} \\
    \ref{sec:supp_gtdet} & Reachability given ground-truth lanes & \S\ref{sec:gtdet} \\
    \ref{sec:supp_qual} & More qualitative results & \S\ref{sec:qual} \\
    \bottomrule
  \end{tabular}
\end{center}

\section{Learned Heads vs.\ Endpoint Rule}
\label{sec:supp_rule}

\begin{table*}[t]
  \centering
  \small
  \setlength{\tabcolsep}{4pt}
  \begin{tabularx}{\textwidth}{l|YYY|G|YYY|G}
    \toprule
    \multirow{2}{*}[-0.6ex]{Method} & \multicolumn{4}{c|}{OpenLane-V2 (TOP$_{ll}$)} & \multicolumn{4}{c}{OLR (\rtop{})} \\
    \hhline{~--------}
    & \rule{0pt}{2.7ex}Learned & Rule & UB & \ghead{Rule / UB} & Learned & Rule & UB & \ghead{Rule / UB} \\
    \midrule
    TopoNet~\cite{toponet}     & 14.6 & 31.4 & 33.8 & 92.8 & 11.0 & 0.1 & 41.9 & 0.3 \\
    TopoMLP~\cite{topomlp}     & 29.5 & 32.2 & 34.2 & 94.1 & 27.9 & 0.1 & 40.6 & 0.4 \\
    TopoLogic~\cite{topologic} & 33.3 & 33.7 & 35.9 & 93.7 & 11.9 & 0.1 & 43.6 & 0.3 \\
    TopoFG~\cite{topofg}       & 38.2 & 38.5 & 40.3 & 95.7 & 15.7 & 0.1 & 47.1 & 0.3 \\
    TopoPoint~\cite{topopoint} & 37.7 & 36.5 & 39.1 & 93.5 & 34.6 & 0.2 & 46.7 & 0.3 \\
    CoPo~\cite{copo}           & 38.9 & 42.0 & 43.5 & 96.5 & 36.6 & 0.2 & 48.0 & 0.3 \\
    \midrule
    \ours (Ours)               & 42.9 & 43.5 & 45.5 & 95.6 & 46.3 & 0.2 & 52.9 & 0.4 \\
    \bottomrule
  \end{tabularx}
  \caption{Learned topology heads and the endpoint rule ($\tau = 3$\,m) on \textit{subset\_A} val, scored only on the connections that OLR changes. Each method uses its own detected lanes. UB (upper bound): the ground-truth connections among the matched lanes. The shaded Rule / UB columns (\%) of the prior methods are plotted in \cref{fig:teaser}~(d).}
  \label{tab:supp_rule}
\end{table*}

\begin{table}[t]
  \centering
  \small
  \setlength{\tabcolsep}{9pt}
  \begin{tabular}{c|cc|cc}
    \toprule
    \multirow{2}{*}[-0.6ex]{$\tau$ (m)} & \multicolumn{2}{c|}{OpenLane-V2} & \multicolumn{2}{c}{OLR} \\
    \hhline{~----}
    & \rule{0pt}{2.7ex}Min & Max & Min & Max \\
    \midrule
    1   & 50.6 & 82.9 & 0.08 & 0.19 \\
    1.5 & 72.0 & 90.1 & 0.16 & 0.24 \\
    2   & 83.6 & 93.5 & 0.19 & 0.30 \\
    \rowcolor{olsbg}
    3   & 92.8 & 96.5 & 0.26 & 0.37 \\
    4   & 95.4 & 97.6 & 0.30 & 0.46 \\
    5   & 96.0 & 97.8 & 0.35 & 0.53 \\
    10  & 96.2 & 97.9 & 2.00 & 2.42 \\
    \bottomrule
  \end{tabular}
  \caption{Endpoint rule as a percentage of the upper bound (Rule / UB, \%) for each threshold $\tau$, with the minimum and maximum over the six prior methods of \cref{tab:supp_rule}. Shaded: $\tau = 3$\,m, used in the main paper.}
  \label{tab:supp_tau}
\end{table}
This section details the comparison of \cref{sec:olv2_limits} and the numbers behind \cref{fig:teaser}~(d).

\paragraph{Endpoint rule.}
The endpoint rule scores an ordered pair of predicted lanes $(u, v)$ as
\begin{equation}
  s_{uv} = \frac{1}{1 + (d_{uv}/\tau)^2},
  \label{eq:supp_rule}
\end{equation}
where $d_{uv}$ is the 3D distance from the end of $u$ to the start of $v$ and $\tau = 3$\,m.
The evaluation counts a connection when its score exceeds 0.5, which holds exactly when $d_{uv} < \tau$.
The rule replaces only the lane--lane scores of a method: its detected lanes and their matching to the ground truth stay as they are.

\paragraph{Connections counted.}
\cref{fig:teaser}~(d) counts only the connections that OLR changes.
On OpenLane-V2, these are the connections with an end on an intersection centerline, that is, on a lane that OLR removes.
On OLR, they are the connections across an intersection.
TOP$_{ll}$ is averaged over the rows and columns of the adjacency matrix that hold such a connection, with the other connections left out of them, so the learned scores differ from those of \cref{tab:main,tab:olv2}.

\paragraph{Upper bound.}
The upper bound of a method is the score of the ground-truth connections among the lanes it matched.
No connection head can score higher on those lanes, so dividing by it separates the connection decision from lane detection.

\paragraph{Results.}
\cref{tab:supp_rule} lists the scores plotted in \cref{fig:teaser}~(d).
On OpenLane-V2, the rule reaches 92.8--96.5\% of the upper bound and scores higher than the learned head of six of the seven methods.
On OLR, it reaches 0.3--0.4\%, as an entry lane and the exits it reaches no longer touch.
\paragraph{Threshold.}
\cref{tab:supp_tau} repeats the comparison for $\tau$ from 1 to 10\,m.
On OpenLane-V2, the rule reaches 83.6--93.5\% of the upper bound already at $\tau = 2$\,m and more than 95\% from $\tau = 4$\,m.
On OLR, it stays below 0.6\% up to $\tau = 5$\,m.
Even at $\tau = 10$\,m, where lanes 10\,m apart count as connected, it reaches only 2.0--2.4\%.
The shortcut on OpenLane-V2 thus holds for any $\tau$ from 2\,m on, and its absence on OLR for every $\tau$ tested.

\section{Constructing OLR}
\label{sec:supp_rules}

\input{tables/olr_layouts}

\input{tables/olv2r_stats}

\input{tables/olr_conn_stats}

OLR is built from every frame of the training and validation sets of OpenLane-V2.
The conversion uses the centerlines of the source annotations, together with the HD maps from which OpenLane-V2 was built: Argoverse~2~\cite{argoverse2} for \textit{subset\_A} and nuScenes~\cite{nuscenes} for \textit{subset\_B}.
It proceeds in six steps: intersection areas, stop lines, cutting, connections, the region of interest, and lane--TE links.
\cref{tab:olr_layouts} lists how particular layouts are handled, and the end of this section reports statistics of the labels.

\paragraph{Intersections and stop lines.}
Intersection areas are taken from the intersection labels of the source maps, given per lane segment in Argoverse~2 and per road segment in nuScenes.
Not every labeled area is an intersection, however: some lie on a straight road that only parallel lanes cross, and U-turn bays join a road to itself.
An area therefore counts as an intersection only when roads from two or more directions meet in it.
These areas are merged into one intersection when they overlap or nearly touch.
Each arm of an intersection then gets one stop line, perpendicular to its lanes, where the first of them enters the area or branches at its edge.
The lanes of an arm are thus all cut at the same place, and the inside of an intersection is the region enclosed by its stop lines.

\begin{figure*}[t]
  \centering
  \includegraphics[width=\linewidth]{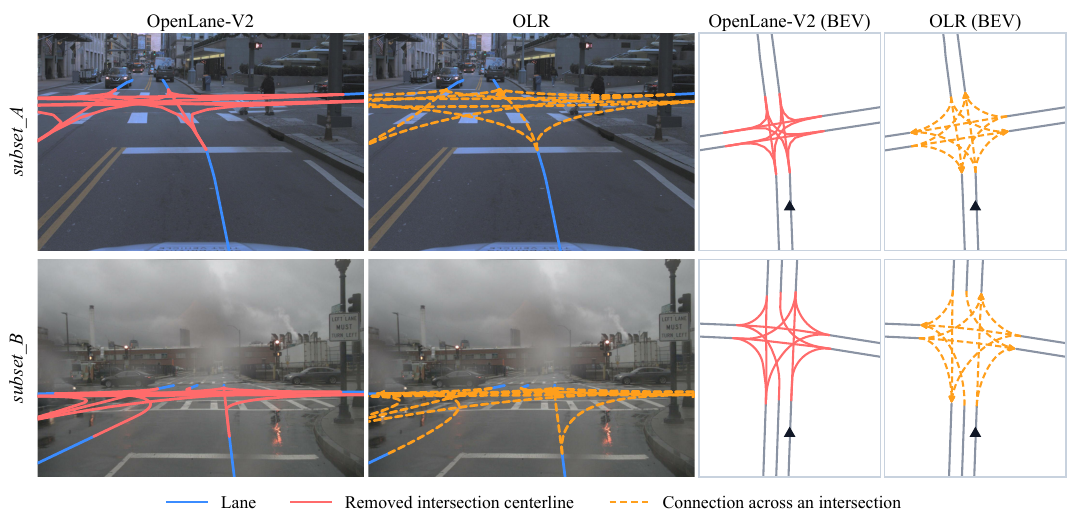}
  \caption{Lane--lane labels of OpenLane-V2 and OLR on \textit{subset\_A} (top) and \textit{subset\_B} (bottom).}
  \label{fig:olr_conv_lclc}
\end{figure*}

\paragraph{Lanes and connections.}
Each centerline is cut where it crosses a stop line, and the pieces inside are removed; the remaining pieces, joined where one simply continues into the next, become the lanes of OLR.
Two lanes of OLR are connected when removed pieces lead from the end of one to the start of the other (\cref{fig:olr_conv_lclc}).
These pieces are stored with the connection as its path, together with their traffic elements.
Neither is scored: the paths are not lanes of OLR, so DET$_l$ ignores them, and a connection is scored only through \rtop{}, by the lanes it joins.
OpenLane-V2 annotates lanes only within 50\,m ahead and behind and 25\,m to each side of the ego vehicle, so some connections leave this region.
Such a connection is stored apart from the connection matrix, with the point and direction at which it crosses the boundary.
\rtop{} does not score it, while its lane inside the region is still scored by DET$_l$.

\paragraph{Reachability.}
OLR adds no connection from traffic rules: every connection across an intersection follows OpenLane-V2 centerlines.
A turn is therefore reachable in OLR when OpenLane-V2 annotates a centerline for it, and a prohibited turn without one stays absent.
Errors in the OpenLane-V2 centerlines across an intersection, such as a missing or wrong one, thus carry over to OLR.
An entry lane with several exits has one connection to each, each with its own path and its own traffic elements.
OLR stores no turn class, since the turn of a connection follows from its path.

\begin{figure*}[t]
  \centering
  \includegraphics{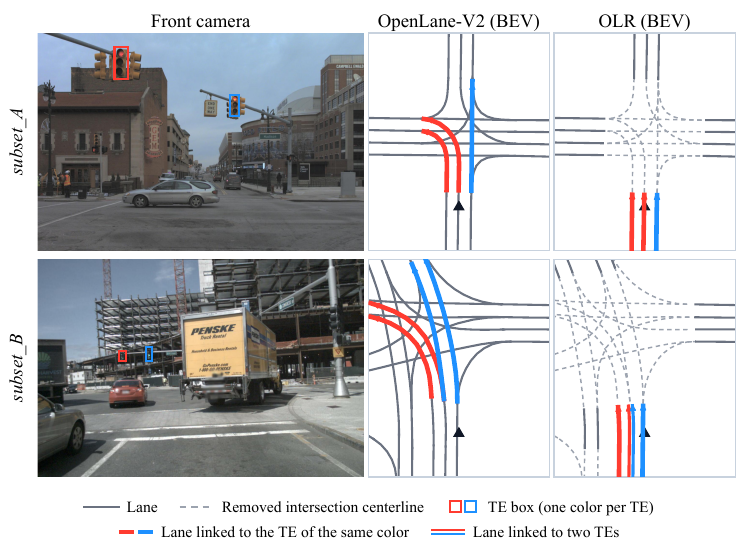}
  \caption{Lane--TE links of OpenLane-V2 and OLR on \textit{subset\_A} (top) and \textit{subset\_B} (bottom). Each lane is drawn in the color of the TE box it is linked to, and a lane linked to two TEs has two lines.}
  \label{fig:olr_conv_lcte}
\end{figure*}

\paragraph{Lane--TE links.}
OpenLane-V2 links traffic lights differently in its two subsets: 90\% of the linked lights in \textit{subset\_A} are linked only to lanes outside an intersection, and 61\% in \textit{subset\_B} only to lanes inside one.
A light governs whether a vehicle crosses the stop line, a decision made on the entry lane, so OLR links it to the entry lane in both subsets.
Each lane of OLR keeps the TE links of the OpenLane-V2 lanes it covers.
A TE linked to a removed piece moves to the entry lane before it, and a sign or road marking that lies past the start of the piece moves instead to the exit lane after it (\cref{fig:olr_conv_lcte}).
A traffic light moves only to an entry lane that faces it, as the lights seen by the front camera face the ego vehicle.
Lane--TE links are not split by turn, so a left-turn signal is linked to its entry lane.

\paragraph{Label statistics.}
\cref{tab:olv2r_stats} compares the labels of OpenLane-V2 and OLR.
OLR has fewer than half as many centerlines per frame.
Except at splits and merges on the road, connected lanes no longer touch, and the median gap from the end of a lane to the start of the next is about 20\,m.
The conversion keeps which lanes are connected: from the end of each OLR lane, the OpenLane-V2 centerlines reach the same lanes as the connections of OLR for all but 0.33\% of the 111{,}671 validation lanes, and 99.3\% (\textit{subset\_A}) and 98.8\% (\textit{subset\_B}) of the TEs linked to a lane stay linked to a lane in OLR.
\cref{tab:olr_conn_stats} describes the connections across an intersection, which 87\% of the validation frames of each subset contain.
51\% (\textit{subset\_A}) and 42\% (\textit{subset\_B}) of the entry lanes reach more than one exit, and 47\% and 39\% reach exits by more than one turn, so an entry lane alone does not tell where it leads.


\section{Evaluation Metrics}
\label{sec:supp_metrics}
OLR is scored with the official evaluation code of OpenLane-V2~\cite{openlanev2} (v2.1.0), unchanged; only the labels differ.
We restate its metrics here, since the matching they share decides which connections count.

\paragraph{Matching.}
For a predicted lane $\ell$ and a ground-truth lane $\bar{\ell}$, the relaxed Fr\'echet distance is
\begin{equation}
  \delta(\ell,\bar{\ell}) = \max\big(0.5,\, 1-0.005\,d(\bar{\ell})\big)\, d_{\mathrm{F}}(\ell,\bar{\ell}),
\end{equation}
where $d_{\mathrm{F}}$ is the discrete Fr\'echet distance and $d(\bar{\ell})$ is the distance in meters from the ego vehicle to the nearest point of $\bar{\ell}$.
At a threshold $\tau\in\mathcal{T}=\{1,2,3\}$\,m, predictions are taken in descending confidence, and each is matched to its nearest ground-truth lane under $\delta$ if $\delta<\tau$ and that lane is not yet matched.
TEs are matched in the same way, with an IoU above 0.25 in place of $\delta<\tau$.

\paragraph{Detection.}
DET$_l$ is the AP of lane detection averaged over $\mathcal{T}$, and DET$_t$ is the AP of TE detection averaged over the TE classes.

\paragraph{Topology.}
For a frame and a threshold, let $A$ be the adjacency matrix over the $N$ ground-truth lanes, and let $\pi(i)$ be the prediction matched to lane $i$.
The predicted scores are carried to the ground-truth lanes: $S_{ij}=\hat{s}_{\pi(i)\pi(j)}$ when $i$ and $j$ are both matched, and otherwise $S_{ij}=0$ if $A_{ij}=1$ and $S_{ij}=0.5+\epsilon$ if $A_{ij}=0$, which ranks these false positives below every predicted connection~\cite{topomlp}.
For lane $i$, let $\mathcal{N}_i=\{j : A_{ij}=1\}$, and rank the lanes $j$ with $S_{ij}>0.5$ by $S_{ij}$.
Then
\begin{equation}
  \mathrm{AP}_i = \frac{1}{|\mathcal{N}_i|}\sum_{k} P_i(k)\,\mathbf{1}\big[j_k\in\mathcal{N}_i\big],
\end{equation}
where $j_k$ is the $k$-th ranked lane and $P_i(k)$ is the precision of the top $k$; $\mathrm{AP}_i$ is 1 when $\mathcal{N}_i$ and the ranking are both empty and 0 when only one is.
The same is computed over the columns of $A$, and TOP$_{ll}$ is the mean of these $2N$ values over all frames and thresholds.
An unmatched ground-truth lane thus scores 0 in its row and its column, whatever its predicted connections.
On OLR, $A$ holds the connections of OLR, and TOP$_{ll}$ is denoted \rtop{}.
TOP$_{lt}$ is computed in the same way on the lane--TE matrix, over the rows of the lanes and the columns of the TEs, and a pair keeps its score only when both its lane and its TE are matched.

\section{Implementation Details}
\label{sec:supp_impl}

\cref{tab:supp_hparams} gives the values of the hyperparameters of \cref{sec:method}, and \cref{alg:train} outlines one training iteration.

\input{tables/supp_hparams}

\input{algs/supp_train}

\paragraph{Training.}
We use AdamW with a learning rate of $2\times10^{-4}$, scaled by 0.1 for the backbone, a weight decay of 0.01, and a cosine schedule after a linear warm-up of 500 optimizer updates.
Gradients are clipped at a norm of 35, separately for the detector and for the pair head and approach association.
The total batch size is 8, on four NVIDIA RTX 5090 GPUs.
The modules switched on at epoch 13 start from their own learning rate of $2\times10^{-4}$ with a warm-up of 500 updates, which decays with a cosine schedule to the end of epoch 24.
The gradient from the pair head into the detector is scaled by 0.1 (\cref{sec:train}).

\paragraph{Detector.}
The BEV grid has $100\times200$ cells and covers $\pm51.2$\,m by $\pm25.6$\,m.
The lane decoder has six layers, with 300 one-to-one lane queries and 100 TE queries.
Following hybrid matching~\cite{jia2023hdetr,topomlp}, 1{,}500 auxiliary queries are matched one-to-many to the ground-truth lanes repeated five times, with an auxiliary loss weight of 2.0.
They are trained in all 24 epochs, matched with the same cost as the one-to-one queries in each stage, and only the one-to-one queries are used at inference.
The L1 regression of lanes weights the start and end points twice, in both the decoder and the point refinement; this weighting comes with the refinement, and the base detector of \cref{tab:ablation} weights all points equally.
On \textit{subset\_B}, lanes are handled as 2D coordinates on the ground plane, and so are the offsets of the point refinement.

\paragraph{End-aware point refinement.}
The encoder $f$ has two layers with eight heads.
It reads the 15 points of $\mathcal{P}$, where the two points beyond each end lie 2\,m and 4\,m out.
The output layer is initialized to zero.
The refined lanes get their own Hungarian matching to the ground truth and an L1 loss.

\paragraph{Relative-pose pair head.}
The pair encoder has three layers and takes all 300 lane queries as candidates.
Every encoder layer outputs pair scores, and the loss is applied to each of them.
The input lane coordinates are detached from the gradient.
A prediction whose mean point distance to its nearest ground-truth lane is below 2\,m also takes the connections of that lane as targets.

\paragraph{Topology-preserving assignment.}
The relaxed Fr\'echet matching cost has a weight of 0.025, which puts it on the same scale as the L1 cost.

\input{tables/olv2}

\input{tables/supp_gtdet}

\paragraph{Approach-guided TE association.}
\cref{alg:approach} builds the approaches from the predicted connections.
The hidden dimension is 128.
The auxiliary loss has a weight of 1.0.
In the cues $\mathbf{c}_{ut}$ of \cref{eq:app_corr}, the role of $u$ is one of entry, upstream, exit, nearby, and none, and its lateral position is taken between the leftmost and rightmost lanes of the approach.
Lanes outside every approach also take the correction, with the cues from the approach set to zero.
Its output layer is initialized to zero, and its inputs are detached from the gradient.
On OpenLane-V2, connected lanes touch, so entry lanes cannot be found from connections across an intersection.
There, a lane is taken as an entry lane if its predicted successors split into two or more or turn by more than 30$^\circ$, together with the lanes in the same direction near its end.
Its connections form $\mathcal{X}$, and the turn of each is the change of heading from the end of the entry lane to the end of its successor.

\input{algs/supp_approach}

\paragraph{Baselines.}
All baselines use ResNet-50~\cite{he2016deep} and 24 epochs.
On OLR, each is retrained from its released code with its official settings.
On OpenLane-V2, TopoNet, TopoMLP, and TopoLogic are evaluated from their released checkpoints, and TopoFG, TopoPoint, and CoPo, which have none, are retrained from their released code.
All methods are scored with the official evaluation code in our environment.

\section{Additional Results}
\label{sec:supp_results}

\subsection{Results on OpenLane-V2}
\label{sec:supp_olv2}

\cref{sec:main} states that \ours also achieves state-of-the-art performance on OpenLane-V2 \textit{subset\_A}, and \cref{tab:olv2} gives all metrics against prior methods.
On OpenLane-V2, connections are defined by proximity.
TOP$_{ll}$ is therefore the original lane--lane metric, and OLS takes it in place of \rtop{} in the \rols{} of \cref{eq:ols}.
\ours on OpenLane-V2 uses the same architecture and training as on OLR, except for the entry-lane rule of the approach association (\cref{sec:supp_impl}).
\ours achieves the highest OLS, TOP$_{ll}$, and TOP$_{lt}$.
TopoHR and TopoCurve, whose code is not available, are listed with the numbers of their papers; TopoHR has the highest DET$_l$, and TopoCurve the highest DET$_t$.

\subsection{Reachability Given Ground-Truth Lanes}
\label{sec:supp_gtdet}

\cref{sec:gtdet} gives every method the ground-truth lanes of OLR and compares the precision--recall curves of their connections across an intersection.
As in \cref{sec:gtdet}, pairs joined by a touching connection are neither positives nor negatives.
\cref{tab:supp_gtdet} lists the numbers behind \cref{fig:gtdet}.
For TopoLogic, TopoFG, and TopoPoint, it also reports the scores with the distance term that their released code adds at test time, a fixed function of the end-to-start distance.
It also gives the results when touching connections, as at a split or a merge, are counted as positives.
The distance term then raises the AP$_{>0.5}$ of TopoLogic and TopoFG, since it scores touching pairs high, but still lowers the AP of all three methods, by 2.4--4.8 points.
\ours has the highest AP, AP$_{>0.5}$, recall, and F1 in both settings.

\subsection{More Qualitative Results}
\label{sec:supp_qual}

\paragraph{Lane--lane connections.}
\cref{fig:supp_qual_1,fig:supp_qual_2,fig:supp_qual_3,fig:supp_qual_4,fig:supp_qual_5,fig:supp_qual_6} show six additional intersections from OLR \textit{subset\_A}.
Each scene shows the seven camera images and, in BEV, the ground truth and the lane centerlines and lane--lane connections predicted by each method.
Green arrows indicate ground-truth connections, and blue, red, and dashed orange arrows indicate true positive, false positive, and false negative predictions, respectively.
In \cref{fig:supp_qual_1,fig:supp_qual_2,fig:supp_qual_3,fig:supp_qual_4,fig:supp_qual_5,fig:supp_qual_6}, as in \cref{fig:qual}, the methods predict largely similar centerlines and differ mainly in the connections across the intersection.
Prior methods miss many of these connections, and in \cref{fig:supp_qual_5}, TopoFG, TopoPoint, and TopoLogic detect every centerline of the junction but predict none of the connections across it.
\ours recovers every connection in five scenes and 23 of 25 in \cref{fig:supp_qual_6}, the largest intersection.
It adds one false connection in each of \cref{fig:supp_qual_4,fig:supp_qual_5,fig:supp_qual_6}; in \cref{fig:supp_qual_4}, it links a left turn to both of two adjacent exit lanes, as CoPo and TopoPoint also do.

\paragraph{Lane--TE links.}
\cref{fig:supp_lcte} shows the lane--TE predictions of all methods on two frames each of \textit{subset\_A} (a, b) and \textit{subset\_B} (c, d).
Above each BEV view, the TEs are cropped from the image and linked to their lanes, so the links can be read against the predicted lanes.
Most errors of prior methods are links left out between lanes and TEs that they detect.
In \cref{fig:supp_lcte} (d), TopoPoint and TopoLogic detect the lanes and TEs of all nine links but predict none of them.
\ours predicts every ground-truth link in all four frames, and TopoFG misses one, in \cref{fig:supp_lcte} (b).

\onecolumn
\input{figs/qualitative/supp_v71/scenes}

\clearpage
\input{figs/qualitative/lcte/scenes}

%% file: tables/olr_layouts.tex
\begin{table}[t]
  \centering
  \small
  \setlength{\tabcolsep}{7pt}
  \renewcommand{\arraystretch}{1.2}
  \begin{tabular}{l|l}
    \toprule
    Layout & Handling in OLR \\
    \midrule
    Roundabout & Ring areas are intersections \\
    Staggered junction & One if the areas nearly touch \\
    Adjacent junctions & Separated by their stop lines \\
    Slip lane, free right & Removed inside an area only \\
    Split / merge, inside & Part of a connection path \\
    Split / merge, outside & As in OpenLane-V2 \\
    U-turn bay & Not an intersection \\
    U-turn, inside & An ordinary connection \\
    Overlapping turn lane & From the through lane \\
    No stop line & The whole area is removed \\
    \bottomrule
  \end{tabular}
  \caption{Particular layouts and how the construction of OLR handles them. Area: an intersection area; inside and outside refer to it. A lane between two areas always stays a lane. Overlapping turn lane: a turn lane that starts with a through lane and overlaps it.}
  \label{tab:olr_layouts}
\end{table}

%% file: tables/olv2r_stats.tex
\begin{table}[t]
  \centering
  \small
  \setlength{\tabcolsep}{5pt}
  \begin{tabular}{l|r@{\;$\rightarrow$\;}l|r@{\;$\rightarrow$\;}l}
    \toprule
    & \multicolumn{4}{c}{OpenLane-V2 $\rightarrow$ OLR} \\
    Per frame & \multicolumn{2}{c|}{\textit{subset\_A}} & \multicolumn{2}{c}{\textit{subset\_B}} \\
    \midrule
    Centerlines & 26.4 & 11.4 & 25.0 & 9.4 \\
    Connections & 25.0 & 9.6 & 22.4 & 6.6 \\
    Intersection centerlines (\%) & 54.5 & 0.0 & 58.1 & 0.0 \\
    Endpoint gap (m) & 0.0 & 19.9 & 0.0 & 19.4 \\
    \bottomrule
  \end{tabular}
  \caption{Labels per frame before (OpenLane-V2) and after (OLR) the conversion, on the validation sets. Endpoint gap: the median distance from the end of a centerline to the start of the next one.}
  \label{tab:olv2r_stats}
\end{table}

%% file: tables/olr_conn_stats.tex
\begin{table}[t]
  \centering
  \small
  \setlength{\tabcolsep}{3.8pt}
  \begin{tabular}{p{1.45cm}l|cc}
    \toprule
    \multicolumn{2}{l|}{} & \textit{subset\_A} & \textit{subset\_B} \\
    \midrule
    \multirow{2}{=}{Per frame} & Across an intersection & 8.5 & 5.9 \\
    & Leaving the region & 4.7 & 4.3 \\
    \midrule
    \multirow{4}{=}{Turn (\%)} & Straight & 42.2 & 41.6 \\
    & Left & 26.3 & 28.7 \\
    & Right & 30.8 & 29.7 \\
    & U-turn & 0.7 & 0.1 \\
    \midrule
    \multirow{2}{=}{Entry lanes (\%)} & More than one exit & 51.3 & 42.1 \\
    & More than one turn & 46.9 & 39.1 \\
    \bottomrule
  \end{tabular}
  \caption{Connections across an intersection in OLR on the validation sets. Turn: the change of heading from entry to exit; straight below 30$^\circ$, U-turn above 150$^\circ$. Entry lane: a lane with a connection across an intersection.}
  \label{tab:olr_conn_stats}
\end{table}

%% file: tables/supp_hparams.tex
\begin{table}[t]
  \centering
  \small
  \setlength{\tabcolsep}{2.3pt}
  \begin{tabular}{p{1.7cm}l|c|c}
    \toprule
    \multicolumn{2}{l|}{Hyperparameter} & Notation & Value \\
    \midrule
    \multirow{2}{=}{End refinement} & Layers / heads & $f$ & 2 / 8 \\
    & Points past the ends & $\mathcal{P}$ & 2, 4\,m \\
    \midrule
    \multirow{3}{=}{Pair head} & Layers / heads & $f'$ & 3 / 8 \\
    & Wavelengths (16) & $\phi$, $g$ & 0.4--200\,m \\
    & Duplicate target dist. & -- & $\leq$ 2\,m \\
    \midrule
    \multirow{3}{=}{Assignment} & Matching cost & $\delta$ & 0.025 \\
    & Target thresholds & $\mathcal{T}$ & $\{1, 2, 3\}$\,m \\
    & Focal loss & $\alpha$, $\gamma$ & 0.25, 2 \\
    \midrule
    \multirow{6}{=}{Approach TE} & Connection score & $\sigma(z_{uv})$ & $\geq$ 0.5 \\
    & Lane score & $\hat{y}_u$ & $\geq$ 0.2 \\
    & Gap at intersection & $\lVert\mathbf{s}_v-\mathbf{e}_u\rVert$ & $\geq$ 1.5\,m \\
    & Straight / U-turn & $|\Delta_{uv}|$ & $30^\circ$ / $150^\circ$ \\
    & End-direction angle & $\mathbf{d}^{\mathrm{e}}_u$ & $\leq 20^\circ$ \\
    & Offset along / across & $\mathbf{e}_u$ & $\leq$ 8 / 12\,m \\
    \bottomrule
  \end{tabular}
  \caption{Hyperparameters of \ours and their notation in \cref{sec:method}, grouped by component (\cref{sec:refine,sec:pair,sec:tpa,sec:approach}). The values are the same on OLR and OpenLane-V2; on OpenLane-V2, entry lanes are found without the gap (\cref{sec:supp_impl}).}
  \label{tab:supp_hparams}
\end{table}

%% file: algs/supp_train.tex
\begin{algorithm}[t]
  \caption{One training iteration of \ours at epoch $e$. $\mathrm{sg}(\cdot)$ stops the gradient, and $\mathrm{gs}(\cdot)$ passes its input unchanged but scales its gradient by 0.1.}
  \label{alg:train}
  \small
  \begin{algorithmic}[1]
    \Require multi-view images, front view, and the labels of lanes, connections, TEs, and lane--TE links
    \State $\mathbf{F}_{\mathrm{bev}} \gets$ BEV encoder on the multi-view images
    \State $\{(\mathbf{q}_i, \ell_i, \hat{y}_i)\}_{i=1}^{300} \gets$ lane decoder on $\mathbf{F}_{\mathrm{bev}}$
    \State $\ell'_i \gets \ell_i + \Delta\ell_i$ \Comment{\cref{eq:refine,eq:refine_out}}
    \State TEs $t$ from the front view; lane--TE logits $z^{\mathrm{lt}}_{it}$
    \If{$e \le 12$}
      \State Hungarian matching with the L1 cost; $y_i \in \{0, 1\}$
      \State $\mathcal{L} \gets \mathcal{L}_{\mathrm{det}} + \mathcal{L}_{\mathrm{ref}}$
    \Else
      \State Hungarian matching with $\delta$; $y_i$ by \cref{eq:tpa}
      \State $z_{uv} \gets$ pair head on $\mathrm{gs}(\mathbf{q}_u)$, $\mathrm{sg}(\ell'_u)$ \Comment{\cref{eq:pair_token,eq:score}}
      \State approaches from $\mathrm{sg}(\ell', \hat{y}, \sigma(z))$ \Comment{\cref{alg:approach}}
      \State $a_{At}$, $\hat{z}^{\mathrm{lt}}_{ut}$ from detached inputs \Comment{\cref{eq:app_aux,eq:app_corr}}
      \State $\mathcal{L} \gets \mathcal{L}_{\mathrm{det}} + \mathcal{L}_{\mathrm{ref}} + \mathcal{L}_{\mathrm{pair}} + \mathcal{L}^{\mathrm{app}}_{\mathrm{lt}} + \mathcal{L}_{\mathrm{aux}}$
    \EndIf
    \State AdamW step, clipping the detector and the new modules separately
  \end{algorithmic}
\end{algorithm}

%% file: tables/olv2.tex
\begin{table}[t]
  \centering
  \small
  \setlength{\tabcolsep}{4pt}
  \begin{adjustbox}{max width=\linewidth}
    \begin{tabular}{l|c|cc|cc}
      \toprule
      Method & OLS & DET$_l$ & DET$_t$ & TOP$_{ll}$ & TOP$_{lt}$ \\
      \midrule
      TopoNet~\cite{toponet} & 39.8 & 28.6 & 48.6 & 10.9 & 23.9 \\
      TopoMLP~\cite{topomlp} & 44.1 & 28.5 & 49.5 & 21.7 & 26.9 \\
      TopoLogic~\cite{topologic} & 44.1 & 29.9 & 47.2 & 24.0 & 25.3 \\
      TopoFG~\cite{topofg} & 46.5 & 32.3 & 48.1 & 28.3 & 27.5 \\
      TopoPoint~\cite{topopoint} & 48.0 & 30.3 & \underline{54.9} & 27.6 & 29.3 \\
      CoPo~\cite{copo} & 47.8 & 32.1 & 48.6 & 29.2 & 32.2 \\
      TopoHR$^\dagger$~\cite{topohr} & 49.9 & \textbf{36.1} & 48.3 & \underline{31.8} & \underline{34.6} \\
      TopoCurve$^\dagger$~\cite{topocurve} & \underline{50.6} & \underline{33.9} & \textbf{55.6} & 30.7 & 33.0 \\
      \midrule
      \ours (Ours) & \textbf{50.9} & 32.4 & 53.6 & \textbf{32.3} & \textbf{36.9} \\
      \bottomrule
    \end{tabular}
  \end{adjustbox}
  \caption{Results on OpenLane-V2 \textit{subset\_A} val. Best in bold, second best underlined. $^\dagger$Numbers from their papers.}
  \label{tab:olv2}
\end{table}

%% file: tables/supp_gtdet.tex
\begin{table*}[t]
  \centering
  \small
  \setlength{\tabcolsep}{4pt}
  \begin{tabularx}{\textwidth}{l|YYYYY|YYYYY}
    \toprule
    \multirow{2}{*}[-0.6ex]{Method} & \multicolumn{5}{c|}{Across an intersection} & \multicolumn{5}{c}{Touching counted as positives} \\
    \hhline{~----------}
    & \rule{0pt}{2.7ex}AP & AP$_{>0.5}$ & P$_{0.5}$ & R$_{0.5}$ & F1$_{0.5}$ & AP & AP$_{>0.5}$ & P$_{0.5}$ & R$_{0.5}$ & F1$_{0.5}$ \\
    \midrule
    TopoNet~\cite{toponet}     & 69.6 & 26.9 & 87.7 & 28.9 & 43.5 & 68.3 & 24.9 & 88.1 & 26.7 & 41.0 \\
    TopoMLP~\cite{topomlp}     & 85.9 & 64.4 & 85.5 & 68.4 & 76.0 & 88.0 & 67.3 & 87.3 & 70.7 & 78.1 \\
    TopoLogic~\cite{topologic} & 69.7 & 28.4 & 87.7 & 30.7 & 45.4 & 68.5 & 26.3 & 88.2 & 28.4 & 43.0 \\
    \quad + distance term      & 54.6 & 21.5 & 78.1 & 30.9 & 44.3 & 66.1 & 34.3 & 83.5 & 38.9 & 53.0 \\
    TopoFG~\cite{topofg}       & 73.2 & 35.9 & \textbf{91.1} & 37.9 & 53.5 & 71.7 & 34.7 & \textbf{91.7} & 36.3 & 52.0 \\
    \quad + distance term      & 57.7 & 28.6 & 82.5 & 38.1 & 52.2 & 68.2 & 40.3 & 86.3 & 45.2 & 59.4 \\
    TopoPoint~\cite{topopoint} & 84.8 & 67.2 & 82.0 & 72.5 & 77.0 & 86.1 & 68.1 & 83.8 & 72.6 & 77.8 \\
    \quad + distance term      & 72.5 & 57.2 & 78.1 & 72.7 & 75.3 & 81.3 & 67.1 & 80.8 & 75.9 & 78.2 \\
    CoPo~\cite{copo}           & 86.8 & 69.7 & 82.9 & 74.5 & 78.5 & 88.2 & 70.9 & 84.6 & 74.9 & 79.5 \\
    \midrule
    \ours (Ours)               & \textbf{89.5} & \textbf{82.1} & 79.1 & \textbf{88.8} & \textbf{83.7} & \textbf{91.5} & \textbf{84.6} & 81.2 & \textbf{90.0} & \textbf{85.4} \\
    \bottomrule
  \end{tabularx}
  \caption{Connection prediction given ground-truth lanes on OLR \textit{subset\_A} val, the numbers behind \cref{fig:gtdet}. AP: area under the precision--recall curve; AP$_{>0.5}$: the part of it above a score of 0.5; P$_{0.5}$, R$_{0.5}$, and F1$_{0.5}$: at a score of 0.5. Left: connections across an intersection, as in \cref{fig:gtdet}. Right: touching connections, as at a split or a merge, also count as positives. + distance term: the fixed function of the end-to-start distance that the released code adds to the learned score at test time, dashed in \cref{fig:gtdet}. Best in bold.}
  \label{tab:supp_gtdet}
\end{table*}

%% file: algs/supp_approach.tex
\begin{algorithm}[t]
  \caption{Approaches from the predicted connections}
  \label{alg:approach}
  \small
  \begin{algorithmic}[1]
    \Statex \textbf{Input:} lanes $u$ with score $\hat{y}_u$, start $\mathbf{s}_u$, and end $\mathbf{e}_u$; connection probabilities $p_{uv} = \sigma(z_{uv})$
    \State $\mathcal{C} \gets \{(u, v) : p_{uv} > 0.5,\ \hat{y}_u, \hat{y}_v \ge 0.2\}$ \Comment{connections}
    \State $\mathcal{X} \gets \{(u, v) \in \mathcal{C} : \lVert \mathbf{s}_v - \mathbf{e}_u \rVert \ge 1.5\,\mathrm{m}\}$ \Comment{across}
    \State \textbf{Entry}: $\mathcal{E} \gets \{u : (u, v) \in \mathcal{X}\}$; \textbf{exit}: $v \notin \mathcal{E}$ with $(u, v) \in \mathcal{X}$
    \State Turn profile: $\pi_u[k] \gets$ the largest $p_{uv}$ among $(u, v) \in \mathcal{X}$ of turn $k$ (\cref{sec:pair}, \cref{tab:olr_conn_stats}), or 0
    \State Approaches: connected groups of entry lanes, linking two whose end directions differ by at most $20^\circ$ and whose ends lie within $\pm8$\,m along and $\pm12$\,m across each other's end direction
    \State \textbf{Upstream}: a lane that reaches $u \in \mathcal{E}$ through at most three connections in $\mathcal{C}$, each touching (${<}\,1.5$\,m) and straight (${<}\,30^\circ$), joins the approach of $u$ and takes $\pi_u$
    \State \textbf{Nearby}: a lane with no connection whose end lies within that window around some $u \in \mathcal{E}$ joins the approach of $u$
  \end{algorithmic}
\end{algorithm}

%% file: figs/qualitative/supp_v71/scenes.tex
\begingroup
\setlength{\tabcolsep}{1.5pt}
\newlength{\sqSideW}\setlength{\sqSideW}{.11\textwidth}
\newlength{\sqSideH}\setlength{\sqSideH}{\dimexpr\sqSideW*1550/2048\relax}
\newlength{\sqFrontH}\setlength{\sqFrontH}{\dimexpr3\sqSideH+2pt\relax}
\newlength{\sqBevW}\setlength{\sqBevW}{.1\textwidth}
\newcommand{\sqAsset}[1]{figs/qualitative/supp_v71/#1.pdf}
\newcommand{\sqBadge}[2]{\raisebox{\dimexpr#1-\height-1.5pt\relax}[0pt][0pt]{\hspace*{1.5pt}\setlength{\fboxsep}{1pt}\colorbox{black!70}{\color{white}\fontsize{5}{5}\selectfont\sffamily\bfseries #2}}}
\newcommand{\sqLab}[4]{\makebox[#2][l]{\rlap{#3}\rlap{\sqBadge{#1}{#4}}}}
\newcommand{\sqCam}[3]{\sqLab{\sqSideH}{\sqSideW}{\includegraphics[width=\sqSideW]{\sqAsset{#1_#2}}}{#3}}
\newcommand{\sqStack}[3]{\vtop{\offinterlineskip\hrule height0pt\hbox{\sqCam{#1}{front_#2}{F#3}}\kern1pt\hbox{\sqCam{#1}{side_#2}{S#3}}\kern1pt\hbox{\sqCam{#1}{rear_#2}{R#3}}}}
\newcommand{\sqFront}[1]{\vtop{\hrule height0pt\hbox{\sqLab{\sqFrontH}{\dimexpr\sqFrontH*1550/2048\relax}{\includegraphics[height=\sqFrontH]{\sqAsset{#1_front_center}}}{F}}}}
\newcommand{\sqCams}[1]{\centerline{\hbox{\sqStack{#1}{left}{L}\kern1pt\sqFront{#1}\kern1pt\sqStack{#1}{right}{R}}}}
\newcommand{\sqB}[3]{\includegraphics[trim=#3,clip,width=\sqBevW]{\sqAsset{#1_#2}}}
\newcommand{\sqBev}[2]{\centerline{\scriptsize\begin{tabular}{@{}c*{7}{!{\color{gray!50}\vrule width .3pt}c}@{}}
\sqB{#1}{gt}{#2}&\sqB{#1}{toporeach}{#2}&\sqB{#1}{copo}{#2}&\sqB{#1}{topofg}{#2}&\sqB{#1}{topopoint}{#2}&\sqB{#1}{topologic}{#2}&\sqB{#1}{topomlp}{#2}&\sqB{#1}{toponet}{#2}\\[1pt]
GT&\textbf{TopoReach}&CoPo~\cite{copo}&TopoFG~\cite{topofg}&TopoPoint~\cite{topopoint}&TopoLogic~\cite{topologic}&TopoMLP~\cite{topomlp}&TopoNet~\cite{toponet}
\end{tabular}}}
\newcommand{\sqScene}[2]{\sqCams{#1}\par\vspace{4pt}\sqBev{#1}{#2}\par}
\newcommand{\sqFigure}[3]{%
  \noindent\begin{minipage}{\textwidth}
    \captionsetup{type=figure}
    \sqScene{#1}{#2}
    \vspace{2pt}\noindent\includegraphics[trim=158.18bp 0bp 156.32bp 0bp,clip,width=\linewidth]{figs/qualitative/v63/legend_gt.pdf}\par
    \vspace{4pt}
    \caption{\textbf{Qualitative comparison on OLR \textit{subset\_A}.}}
    \label{#3}
  \end{minipage}}

\vspace*{\fill}
\sqFigure{10071_02}{0bp 150bp 0bp 150bp}{fig:supp_qual_1}

\vfill

\sqFigure{10023_09}{0bp 151bp 0bp 151bp}{fig:supp_qual_2}

\vspace*{\fill}\clearpage
\vspace*{\fill}
\sqFigure{10131_06}{0bp 150bp 0bp 150bp}{fig:supp_qual_3}

\vfill

\sqFigure{10008_16}{0bp 5bp 0bp 5bp}{fig:supp_qual_4}

\vspace*{\fill}\clearpage
\vspace*{\fill}
\sqFigure{10084_13}{0bp 151bp 0bp 151bp}{fig:supp_qual_5}

\vfill

\sqFigure{10061_01}{0bp 151bp 0bp 151bp}{fig:supp_qual_6}
\par\vspace*{\fill}
\endgroup

%% file: figs/qualitative/lcte/scenes.tex
\begingroup
\setlength{\tabcolsep}{1.5pt}
\newlength{\ltW}\setlength{\ltW}{.112\textwidth}
\newcommand{\ltF}[2]{\includegraphics[width=\ltW]{figs/qualitative/lcte/#1/front_#2.jpg}}
\newcommand{\ltB}[2]{\includegraphics[width=\ltW]{figs/qualitative/lcte/#1/bev_#2.pdf}}
\newcommand{\ltRow}[2]{%
\multirow{2}{*}{\footnotesize(#2)}&\ltF{#1}{gt}&\ltF{#1}{toporeach}&\ltF{#1}{copo}&\ltF{#1}{topofg}&\ltF{#1}{topopoint}&\ltF{#1}{topologic}&\ltF{#1}{topomlp}&\ltF{#1}{toponet}\\
&\ltB{#1}{gt}&\ltB{#1}{toporeach}&\ltB{#1}{copo}&\ltB{#1}{topofg}&\ltB{#1}{topopoint}&\ltB{#1}{topologic}&\ltB{#1}{topomlp}&\ltB{#1}{toponet}\\}
\noindent\begin{minipage}{\textwidth}
  \captionsetup{type=figure}
  \centerline{\scriptsize\begin{tabular}{@{}c@{\hspace{3pt}}c*{7}{!{\color{gray!50}\vrule width .3pt}c}@{}}
  &GT&\textbf{TopoReach}&CoPo~\cite{copo}&TopoFG~\cite{topofg}&TopoPoint~\cite{topopoint}&TopoLogic~\cite{topologic}&TopoMLP~\cite{topomlp}&TopoNet~\cite{toponet}\\[1pt]
  \ltRow{10073}{a}\noalign{\vspace{5pt}}
  \ltRow{10024}{b}\noalign{\vspace{5pt}}
  \ltRow{11051}{c}\noalign{\vspace{5pt}}
  \ltRow{11018}{d}
  \end{tabular}}
  \vspace{3pt}\centerline{\includegraphics[width=\textwidth]{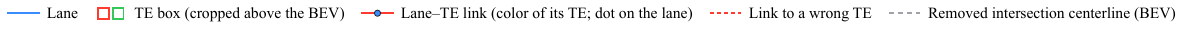}}
  \vspace{2pt}
  \caption{\textbf{Lane--TE predictions on OLR}, on two frames each of \textit{subset\_A} (a, b) and \textit{subset\_B} (c, d).}
  \label{fig:supp_lcte}
\end{minipage}
\endgroup